\documentclass{article}

\usepackage[dvipsnames]{xcolor}
\usepackage{microtype}
\usepackage{graphicx}
\usepackage{subfigure}
\usepackage{dblfloatfix}
\usepackage{booktabs} 

\usepackage{hyperref}

\usepackage[accepted]{icml2024-diffxyz}

\usepackage{amsmath}
\usepackage{amssymb}
\usepackage{mathtools}
\usepackage{amsthm}

\usepackage{float}
\usepackage{multicol}
\usepackage{array}
\usepackage{caption}

\usepackage[capitalize,noabbrev]{cleveref}

\theoremstyle{plain}

\theoremstyle{definition}

\theoremstyle{remark}

\usepackage[textsize=tiny]{todonotes}

\usepackage[
backend=biber,
style=ieee,
citestyle=ieee,
]{biblatex}
\AtBeginBibliography{\small}

\icmltitlerunning{BMapEst: Estimation of Brain Tissue Probability Maps using a Differentiable MRI Simulator}

\begin{document}

\twocolumn[
\icmltitle{BMapEst: Estimation of Brain Tissue Probability \\ Maps using a Differentiable MRI Simulator}



\icmlsetsymbol{equal}{*}

\begin{icmlauthorlist}
\icmlauthor{Utkarsh Gupta}{ucsccse}
\icmlauthor{Emmanouil Nikolakakis}{ucscece}
\icmlauthor{Moritz Zaiss}{fau,earl}
\icmlauthor{Razvan Marinescu}{ucsccse}
\end{icmlauthorlist}


\icmlaffiliation{ucsccse}{University of California, Santa Cruz, Computer Science and Engineering Department}
\icmlaffiliation{ucscece}{University of California, Santa Cruz, Electrical and Computer Engineering Department}
\icmlaffiliation{fau}{Department of Artificial Intelligence in Biomedical Engineering, Friedrich-Alexander-Universität Erlangen}
\icmlaffiliation{earl}{Department of Neuroradiology, Universitätsklinik Erlangen, Erlangen, Germany}

\icmlcorrespondingauthor{Utkarsh Gupta}{utgupta@ucsc.edu}

\icmlkeywords{Machine Learning, ICML}

\vskip 0.3in
]



\printAffiliationsAndNotice{} 

\begin{abstract}
Reconstructing digital brain phantoms in the form of voxel-based, multi-channeled tissue probability maps for individual subjects is essential for capturing brain anatomical variability, understanding neurological diseases, as well as for testing image processing methods. We demonstrate the first framework that estimates brain tissue probability maps (Grey Matter - GM, White Matter - WM, and Cerebrospinal fluid - CSF) with the help of a Physics-based differentiable MRI simulator that models the magnetization signal at each voxel in the volume. Given an observed $T_1$/$T_2$-weighted MRI scan, the corresponding clinical MRI sequence, and the MRI differentiable simulator, we estimate the simulator's input probability maps by back-propagating the L2 loss between the simulator's output and the $T_1$/$T_2$-weighted scan. This approach has the significant advantage of not relying on any training data and instead uses the strong inductive bias of the MRI simulator. We tested the model on 20 scans from the BrainWeb database and demonstrated a highly accurate reconstruction of GM, WM, and CSF. Our source code is available online: \url{https://github.com/BioMedAI-UCSC/BMapEst}.
\end{abstract}

\section{Introduction}

Quantitative $T_1$ (q$T_1$) and $T_2$ (q$T_2$) MRI plays a pivotal role in understanding various neurological conditions, including multiple sclerosis \cite{granziera2021quantitative}, brain tumors \cite{blystad2017quantitative}, and neurodegenerative diseases such as Alzheimer's \cite{leandrou2018quantitative}, by providing valuable insights into tissue characteristics and pathological changes. However, even more information than $T_1$/$T_2$-weighted MRI can be obtained by reconstructing the tissue probability maps of the digital brain phantoms. Such multi-channeled tissue probability maps typically include the percentage of a particular tissue (GM, WM, CSF, blood vessels, glia cells, etc.) that exists at each 3D voxel and thus give significantly more information than standard $T_1$/$T_2$-weighted MRIs. In addition, probability maps are critical as priors in accurate self-supervised semantic segmentation or classification of a 3D volume reconstruction. They also help regularize and eliminate artifacts such as partial volume effects, motion artifacts, or bias field artifacts, which typically require external atlas-based priors \cite{schramm2019metal, wang2022inconsistencies}. BrainWeb phantoms \cite{cocosco1997brainweb,aubert2006new} are popular examples that store probability maps for 11 tissues at each voxel, offering more precise anatomical detail than q$T_1$/q$T_2$ maps, customizable tissue properties, and the ability to simulate realistic imaging scenarios. 


A related problem to tissue probability map estimation is tissue segmentation, where a single label is assigned for each voxel. Current state-of-the-art methods for tissue segmentation use supervised deep-learning \cite{kumari2023residual,zhang2022multi} and clustering methods \cite{tavakoli2021segmentation,wang2008modified,singh2018dct,tavakoli2021segmentation,gong2012fuzzy}. However, tissue segmentation is a different and easier problem than probability map estimation since models don't need to estimate partial volume effects at all voxels. Despite their utility, we are unaware of any methods that fully infer tissue probability maps from $T_1$/$T_2$-weighted scans. Additionally, the problem is particularly challenging due to its ill-posedness at two different levels: 1) multiple probability map combinations can produce the same q$T_1$/q$T_2$/PD maps, and 2) multiple q$T_1$/q$T_2$/PD maps can produce the same $T_1$/$T_2$-weighted image.

MRI imaging simulators \cite{bittoun1984computer,summers1986computer,shkarin1996direct,shkarin1997time,kwan1999mri,yoder2004mri,benoit2005simri,jochimsen2006efficient,stocker2010high,baum2011simulation,xanthis2013mrisimul,xanthis2014high,cao2014bloch} offer a solution for inferring probability maps of digital brain phantoms. They act as a forward model to generate k-space measurements and subsequently reconstruct $T_1$/$T_2$-weighted scans. This represents a paradigm shift in medical imaging simulation and optimization. By leveraging simulators' inherent physics-based inductive biases, it becomes feasible to fine-tune the parameters of clinical sequences, leading to enhanced image quality. A recent innovation in the field is the building of MR-zero \cite{loktyushin2021mrzero}, a \emph{differentiable} MRI simulator implementing the Phase Distribution Graph \cite{Endres2023} (Appendix \ref{appendix:A}), thus alleviating the need for slow derivative-free optimizations.

In this paper, we introduce a novel framework leveraging the capabilities of MR-zero to estimate the tissue probability maps of digital brain phantoms (Fig. \ref{fig:subject42_best_results}) representing the CSF, GM, and WM. Unlike supervised learning, our framework does not need a-priori training pairs of inputs and outputs and can further estimate tissue probability maps for any arbitrary set of MRI sequences (e.g., T$_1$-only, T$_1$+T$_2$, T$_2$+T$_2^*$+GRE, etc.) at any arbitrary echo times. Our contributions are:
\begin{itemize}
    \item We demonstrate the first method of estimating brain tissue probability maps using a differentiable MRI simulator that conducts forward inference to generate a $T_1$/$T_2$-weighted image by backpropagating a loss function to the brain tissue probability maps. Our approach is versatile, applicable to many different MRI sequences, and does not require learnable parameters.
    \item We overcome the ill-posedness of probability maps estimation by using the inductive bias of the simulator and multiple $T_1$/$T_2$ contrasts.
    \item We validate our approach on BrainWeb's 20 subjects with the popular Fast Low Angle Shot (FLASH) sequence variants and obtain state-of-the-art results compared to supervised deep learning and clustering methods.
\end{itemize}

\section{Methodology}

\begin{figure*}[t]
    \centering
    \includegraphics[width=1\textwidth]{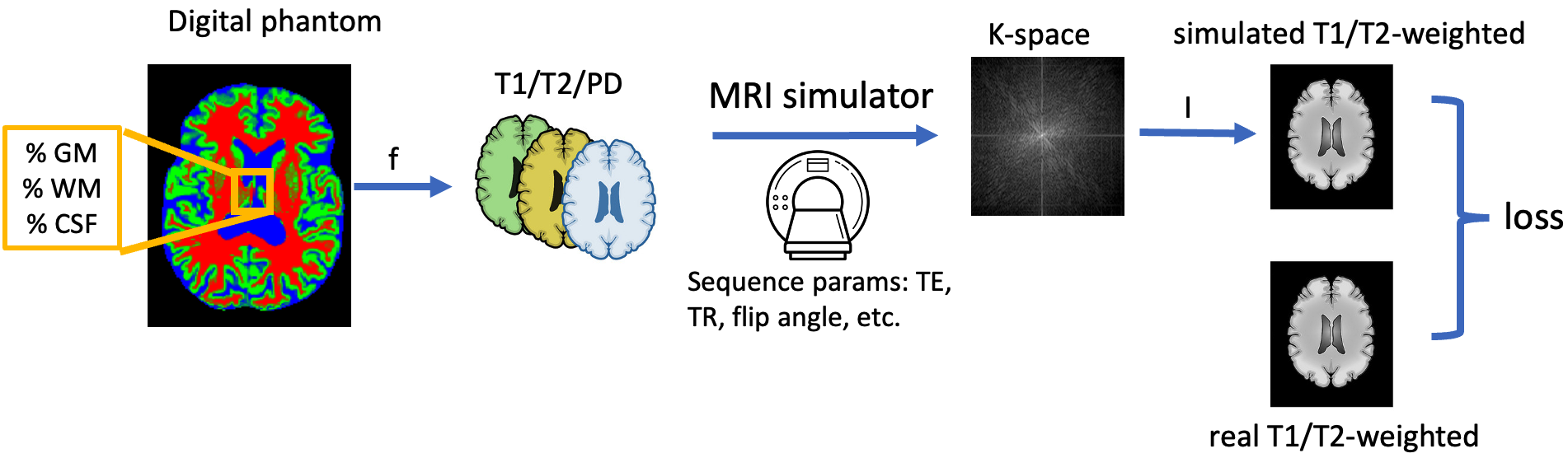}
    \caption{\small{Optimization pipeline: The CSF, GM, and WM probability maps are converted into q$T_1$, q$T_2$, and PD maps using known $T_1$ and $T_2$ relaxation times. These maps are input into the MR-zero simulator to generate K-space measurements, and the image is reconstructed via Inverse Fourier Transform $I$. The L2 loss is computed against a real MRI and backpropagated to the probability maps.}}
    \label{fig:optimization_pipeline}
\end{figure*}


Our framework optimizes brain tissue probability maps given one or more observed $T_1$/$T_2$-weighted MRI scans ($G$) and the corresponding MRI clinical sequence ($S$) used to generate each scan (Fig. \ref{fig:optimization_pipeline}). The optimization pipeline begins bwy transforming the brain tissue probability maps into q$T_1$, q$T_2$, and Proton Density (PD) maps with the help of known relaxation time values for a 1.5T scanner \cite{relaxationTime}. These maps serve as input for the MR-zero (Appendix \ref{appendix:B}) MRI differentiable simulator that produces K-space measurements as its output of the forward pass. By applying the Inverse Fourier Transform ($I$), the simulated $T_1$/$T_2$-weighted scan ($G^{\prime}$) is reconstructed from K-space. Therefore, the end-to-end forward pass is given as follows:

\begin{equation}
    G' = I(MRZero(f(GM, WM, CSF), S))
\end{equation}

where $f$ is a fixed linear transformation mapping tissue probability maps to q$T_1$/q$T_2$/PD maps using known values from the literature. The pixel-wise L2 loss $||G-G'||_2^2$ is calculated between the observed $T_1$/$T_2$-weighted scan $G'$ and the ground truth scan $G$ and backpropagated to the input probability maps (GM/WM/CSF).

\section{Experiments}

\subsection{Dataset} We demonstrate our method on the BrainWeb dataset, which comprises brain tissue probability maps from 20 subjects. We downscaled the 3D brain volumes to 64x64 (higher resolutions were not possible on our machine due to RAM requirements exceeding 64GB) and selected the middle slice, as it contains all three types of tissues, resulting in CSF, GM, and WM probability maps. Due to the simulator's compute-intensive nature, we only select a single slice, but our method is equally generalizable to other slices of the volume as well.

\subsection{Ill-posed problem} The estimation of tissue probability maps presents a challenge due to the inherent ambiguity stemming from the ill-posed nature of the problem, wherein multiple potential solutions (tissue probabilities) exist for a single observed $T_1$/$T_2$-weighted scan. This can be seen in Fig. \ref{fig:subject42_best_results}, where the optimization of all tissue probablitity maps of a subject with only a single observed contrast obtained from the $T_1$ inversion recovery sequence give a blurry estimation of tissue maps. Furthermore, the sensitivity of tissue delineation is influenced by the specific MRI clinical sequence utilized. For instance, the Fluid Attenuated Inversion Recovery (FLAIR) sequence effectively diminishes the signal from CSF, making CSF map estimation from this sequence highly ill-posed. Unlike supervised deep learning methods, the flexibility of our method allows us to augment the optimization process by incorporating additional output contrasts obtained from capturing images at varying echo times and by including a diverse array of sequences such as $T_2$-weighted imaging, $T_2^*$ imaging through echo time alterations, FLAIR, Double Inversion Recovery (DIR), and Diffusion Weighted Imaging (DWI) sequences. 

\subsection{Setup} Our optimization process spans 501 epochs with a learning rate of 0.01, and takes 5-6min for a single contrast and under 3 hours for 24 contrasts (6 sequences with 4 echo times each). We concurrently optimize all probability maps (CSF, GM, and WM), leveraging the output contrasts from up to six different MRI clinical sequences. Loss computation is conducted across reconstructed image spaces, representing the output $T_1$/$T_2$-weighted scans (each with 4 contrasts per sequence). This is the baseline configuration settings for comparing the results. We performed leave-one-out cross-validation. Our approach adds no extra learnable parameters beyond the probability maps themselves.

\section{Results}

As illustrated in Table \ref{tab:loss_table_3} and in Fig. \ref{fig:subject42_best_results}, our method achieves very good reconstructions when at least 4 contrasts (echoes) from a single sequence are observed. Only when one single contrast is used does our method perform poorly due to the strongly ill-posed problem. Best reconstructions are achieved using 5 sequences together: $T_1$, $T_2$, $T_2^*$, double inversion recovery and FLAIR.

In addition, we also compared against a supervised U-Net baseline \cite{ronneberger2015u}. We trained the U-Net by creating pixel-wise normal distributions of probability maps from the middle slices of ten (5, 18, 38, 42, 44, 46, 48, 50, 52, 54) BrainWeb subjects. The input to the U-Net is a 24-channeled MRI contrasts tensor, and the output is a 3-channeled CSF, GM, and WM probability maps. Since training a U-Net is difficult in this low-data regime, we augmented the dataset by building pixel-wise normal distributions and isotropically sampling 500 images from the distribution. We tested the model on the remaining subjects and the same brain slices of training data subjects on which our method's metrics are reported. Table \ref{tab:loss_table_3} shows that our method has better PSNR and SSIM than the U-Net across all tissues. Also, our method has a better DICE score than the U-Net on CSF and a similar DICE score on GM and WM. 

In Table \ref{tab:loss_table_3}, we also tested the BCEFCM \cite{feng2016image}, a fuzzy C-means clustering method. BCEFCM performed sub-optimally for the DICE score and poorly on PSNR and SSIM metrics, which is evident from Appendix Figure \ref{sup_fig:optimization_results},  as clustering-based methods cannot estimate the exact probability values. Moreover, we have taken Table \ref{sup_tab:other_baseline_comparison} from \cite{tavakoli2021segmentation} which shows the DICE metrics obtained for CSF, GM, and WM of the BrainWeb dataset for 4 other clustering methods: LNLFCM \cite{wang2008modified}, DCT-LNLFCM \cite{singh2018dct}, KWFLICM \cite{gong2012fuzzy}, and Double Estimation \cite{tavakoli2021segmentation} based fuzzy C-means. Though the slices used for the metric would differ, they are from the same BrainWeb dataset on which we have reported our results. Our method outperforms all the other techniques for GM and CSF and performs equally well for WM estimation.

For visual results, refer to Figure \ref{sup_fig:optimization_results} in the Appendix for inferred probability map results for our baseline, U-Net, and BCEFCM methods. Also, Appendix \ref{appendix:C} and \ref{appendix:D} show detailed experiments and other visual results respectively.

\section{Conclusion}

We have developed BMapEst, the first framework that estimates digital brain phantoms using an MRI differentiable simulator. In future work, we plan to estimate probability maps on large brain MRI datasets such as OASIS or ADNI. Furthermore, in ill-posed settings, our framework can also be extended using variational inference to capture a \emph{distribution} over the tissue probability maps that generate the same $T_1$/$T_2$-weighted MRI scan. This will allow us to capture the variability in tissue characteristics across different individuals having similar $T_1$/$T_2$-weighted scans. Another direction we plan to pursue is to optimize the MRI clinical sequence parameters towards a downstream goal, such as better image contrast, better segmentation of structures, or better disease detection. By leveraging the differentiable MRI simulators, we can efficiently and accurately model the impact of various sequence parameters on the resulting images, enabling us to fine-tune these parameters to enhance diagnostic utility and achieve optimal imaging sequence.

One of the limitations of differentiable simulators is their high computational requirements during optimization, in particular GPU memory. To address this, we plan to investigate the use of advanced optimization techniques and hardware acceleration, such as leveraging GPUs and CPU RAM in tandem to balance memory consumption, reduce computational load, and improve processing speed. To speed-up
\newcommand{\size}[0]{0.15}
\newcommand{\fcellsize}[0]{\small}
\newcommand{\unet}[0]{\color{red}}
\newcommand{\bce}[0]{\color{blue}}
\renewcommand{\arraystretch}{1}
\begin{figure*}
\centering
\captionof{table}{Dice score, PSNR, and SSIM comparisons for a combination of MRI clinical sequences. Except for the first row, all the experiments use 4 contrasts per sequence. \textdagger\ represents the baseline configuration. Our method is compared against two baselines: \textcolor{red}{U-Net} and  \textcolor{blue}{BCEFCM}. \textbf{Note:} \textcolor{blue}{BCEFCM} can only take a single contrast as input.}
\begin{tabular}{ >{\centering\arraybackslash}m{1.5cm} | >{\centering}m{1.2cm}| >{\centering}m{1.4cm}| >{\centering}m{1.2cm}| >{\centering}m{1.2cm}| >{\centering}m{1.35cm}| >{\centering}m{1.2cm}| >{\centering}m{1.2cm}| >{\centering}m{1.35cm}| >{\arraybackslash}m{1.2cm}}
    \hline
Contrasts    & \multicolumn{3}{c|}{CSF}
            & \multicolumn{3}{c|}{GM}
                    & \multicolumn{3}{c}{WM} \\
    \cline{2-10}
                                                        &  Dice$\uparrow$ & PSNR $\uparrow$  &   SSIM $\uparrow$  &  Dice$\uparrow$  & PSNR $\uparrow$  &   SSIM $\uparrow$  & Dice$\uparrow$ &   PSNR $\uparrow$  &   SSIM $\uparrow$  \\
                \hline
    \small{$T_1$ single contrast}                               & \fcellsize{0.38$\pm$0.07 \unet{0.31$\pm$0.05} \bce{0.04$\pm$0.01}} &   \fcellsize{17.84$\pm$1.18 \unet{18.29$\pm$0.86} \bce{11.47}$\pm$0.4}    &  \fcellsize{0.58$\pm$0.07 \unet{0.63$\pm$0.05} \bce{0.13$\pm$0.02}}   & \fcellsize{0.67$\pm$0.05 
    \unet{0.66$\pm$0.04} \bce{0.40$\pm$0.02}} & \fcellsize{14.22$\pm$0.97 \unet{14.94$\pm$0.74} \bce{7.53$\pm$0.27}}  & \fcellsize{0.57$\pm$0.08 \unet{0.64$\pm$0.05} \bce{0.16$\pm$0.02}}  & \fcellsize{0.62$\pm$0.05 
    \unet{0.65$\pm$0.04} \bce{0.38$\pm$0.04}} &  \fcellsize{12.83$\pm$1.1 \unet{13.5$\pm$0.83} \bce{7.68$\pm$0.27}}  &  \fcellsize{0.38$\pm$0.07 \unet{0.51$\pm$0.06} \bce{0.18$\pm$0.02}} \\
                \hline
    \small{$T_1$ four contrasts}                                & \fcellsize{0.55$\pm$0.04 \unet{0.43$\pm$0.04}} &  \fcellsize{27.57$\pm$2.11 \unet{23.57$\pm$0.85}}     &  \fcellsize{0.89$\pm$0.03 \unet{0.86$\pm$0.02}}   & \fcellsize{0.7$\pm$0.04 \unet{0.72$\pm$0.03}} & \fcellsize{16.16$\pm$0.97 \unet{16.52$\pm$0.53}}  &  \fcellsize{0.74$\pm$0.05 \unet{0.76$\pm$0.02}}  & \fcellsize{0.76$\pm$0.02 \unet{0.76$\pm$0.02}} & \fcellsize{18.15$\pm$0.98 \unet{17.38$\pm$0.48}} &  \fcellsize{0.78$\pm$0.04 \unet{0.75$\pm$0.03}} \\
                \hline
\small{$T_1$+$T_2$}                                             & \fcellsize{0.55$\pm$0.04 \unet{0.43$\pm$0.04}} &  \fcellsize{30.9$\pm$3.93 \unet{22.63$\pm$0.74}}      &  \fcellsize{0.97$\pm$0.02 \unet{0.85$\pm$0.02}}   & \fcellsize{0.76$\pm$0.03 \unet{0.76$\pm$0.02}} & \fcellsize{23.5$\pm$1.98 \unet{19.19$\pm$0.34}}   &  \fcellsize{0.93$\pm$0.03 \unet{0.87$\pm$0.01}}  & \fcellsize{0.79$\pm$0.01 \unet{0.79$\pm$0.02}} & \fcellsize{25.63$\pm$2.43 \unet{19.71$\pm$0.34}} &  \fcellsize{0.93$\pm$0.03 \unet{0.83$\pm$0.02}} \\
                \hline
\small{$T_1$+$T_2$+$T_2^*$}                                     & \fcellsize{\textbf{0.56}$\pm$\textbf{0.03} \unet{0.47$\pm$0.04}} &  \fcellsize{33.43$\pm$4.72 \unet{24.08$\pm$0.8}}     &  \fcellsize{0.97$\pm$0.02 \unet{0.88$\pm$0.02}}   & \fcellsize{0.76$\pm$0.03 \unet{0.76$\pm$0.02}} & \fcellsize{26.38$\pm$2.35 \unet{20.18$\pm$0.36}}  &  \fcellsize{0.96$\pm$0.03 \unet{0.9$\pm$0.01}}  & \fcellsize{0.79$\pm$0.01 \unet{0.79$\pm$0.02}} & \fcellsize{25.16$\pm$2.34 \unet{19.91$\pm$0.35}} &  \fcellsize{0.91$\pm$0.03 \unet{0.83$\pm$0.02}} \\
                \hline
\small{$T_1$ + $T_2$ + $T_2^*$ + DIR}                           & \fcellsize{\textbf{0.56}$\pm$\textbf{0.03} \unet{0.45$\pm$0.04}} &  \fcellsize{33.68$\pm$4.51 \unet{24.33$\pm$0.91}}     &  \fcellsize{0.97$\pm$0.02 \unet{0.89$\pm$0.02}}   & \fcellsize{0.76$\pm$0.03 \unet{0.76$\pm$0.02}} & \fcellsize{26.33$\pm$2.38 \unet{20.48$\pm$0.37}}  &  \fcellsize{0.96$\pm$0.03 \unet{0.91$\pm$0.01}}  & \fcellsize{0.79$\pm$0.01 \unet{0.79$\pm$0.02}} & \fcellsize{25.16$\pm$2.45 \unet{20.49$\pm$0.36}} &  \fcellsize{0.91$\pm$0.04 \unet{0.84$\pm$0.02}} \\
                \hline
\small{$T_1$ + $T_2$ + $T_2^*$ + DIR + FLAIR}           & \fcellsize{0.55$\pm$0.04 \unet{0.45$\pm$0.04}} &  \fcellsize{\textbf{34.64}$\pm$\textbf{5.29} \unet{24.08$\pm$0.78}}    &  \fcellsize{\textbf{0.98}$\pm$\textbf{0.02} \unet{0.89$\pm$0.02}}   & \fcellsize{0.76$\pm$0.02 \unet{\textbf{0.77}$\pm$\textbf{0.02}}} & \fcellsize{\textbf{27.71}$\pm$\textbf{2.48} \unet{20.56$\pm$0.38}}  &  \fcellsize{\textbf{0.97}$\pm$\textbf{0.02} \unet{0.9$\pm$0.01}}  & \fcellsize{\textbf{0.8}$\pm$\textbf{0.01} \unet{0.79$\pm$0.02}} & \fcellsize{\textbf{27.0}$\pm$\textbf{2.69} \unet{20.34$\pm$0.39}}  &  \fcellsize{\textbf{0.93}$\pm$\textbf{0.03} \unet{0.84$\pm$0.18}} \\
                \hline
\small{$T_1$ + $T_2$ + $T_2^*$ + DIR + FLAIR + DWI \textdagger} & \fcellsize{0.55$\pm$0.04 \unet{0.45$\pm$0.04}} &  \fcellsize{34.45$\pm$4.87 \unet{24.23$\pm$0.85}}     &  \fcellsize{\textbf{0.98}$\pm$\textbf{0.02} \unet{0.89$\pm$0.07}}   & \fcellsize{0.76$\pm$0.02 \unet{0.76$\pm$0.02}} & \fcellsize{26.8$\pm$2.54 \unet{20.38$\pm$0.37}}   &  \fcellsize{0.96$\pm$0.02 \unet{0.9$\pm$0.01}}  & \fcellsize{0.79$\pm$0.01 \unet{0.79$\pm$0.02}} & \fcellsize{25.99$\pm$2.69 \unet{20.42$\pm$0.39}} &  \fcellsize{0.92$\pm$0.03 \unet{0.85$\pm$0.02}} \\
    \hline
\end{tabular}
\label{tab:loss_table_3}
\end{figure*}

\begin{figure}
\textbf{\hspace{1.15cm}\small{CSF}\hspace{2.1cm}\small{GM}\hspace{2.05cm}\small{WM}}\par\medskip
    \centering
    \centering
    \includegraphics[width=\size\textwidth]{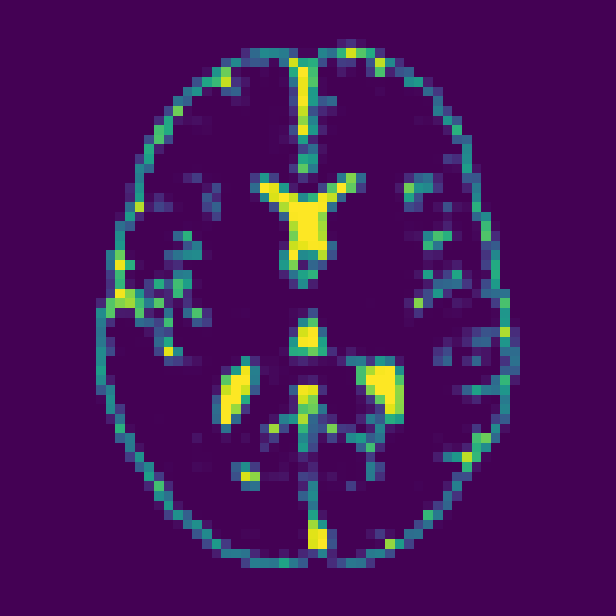}
    \includegraphics[width=\size\textwidth]{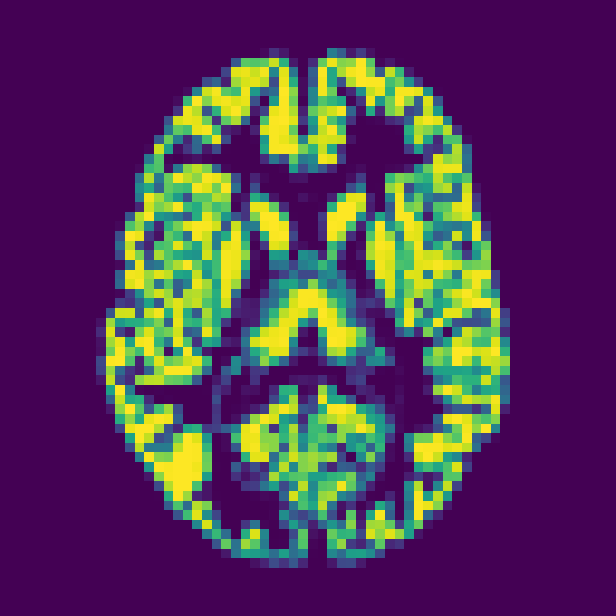}
    \includegraphics[width=\size\textwidth]{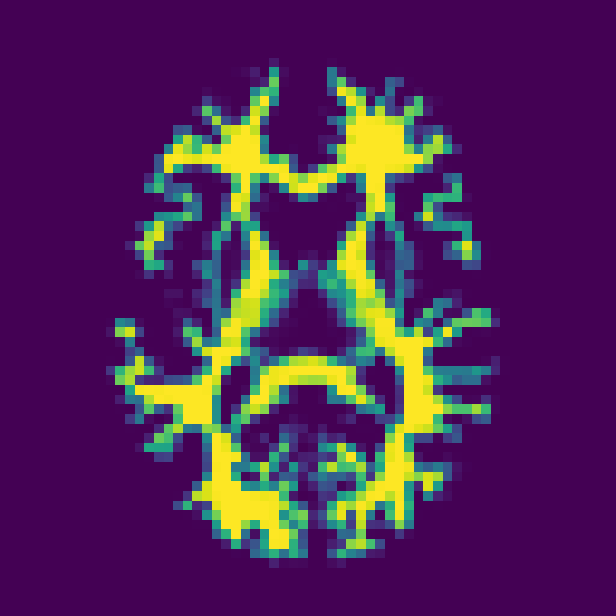}
    \\[\smallskipamount]
    \includegraphics[width=\size\textwidth]{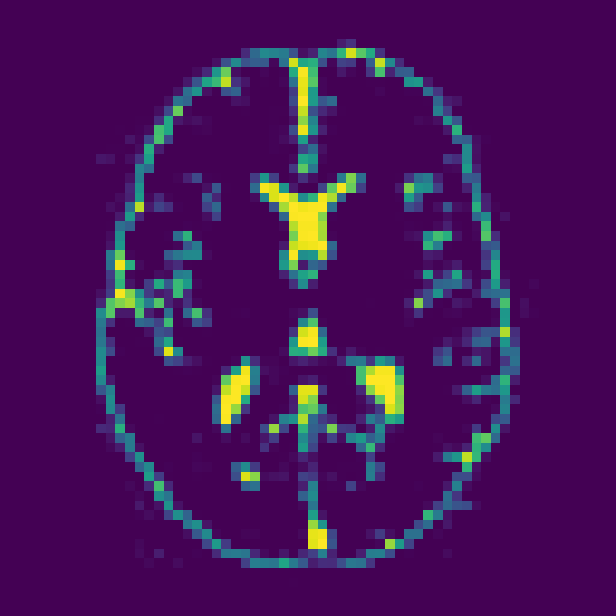}
    \includegraphics[width=\size\textwidth]{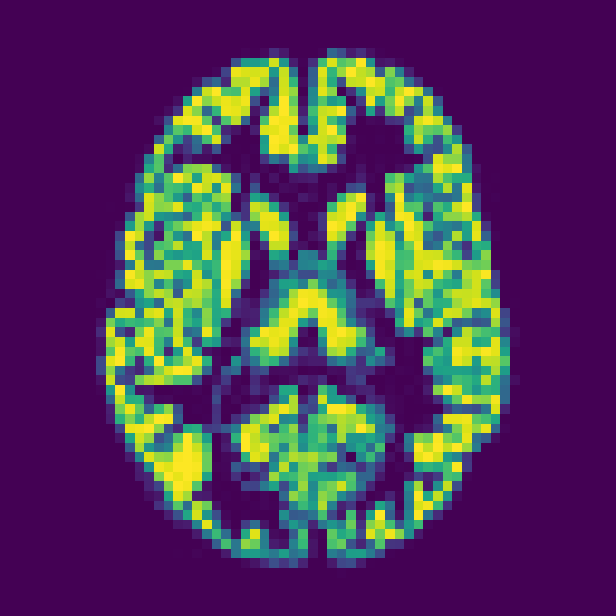}
    \includegraphics[width=\size\textwidth]{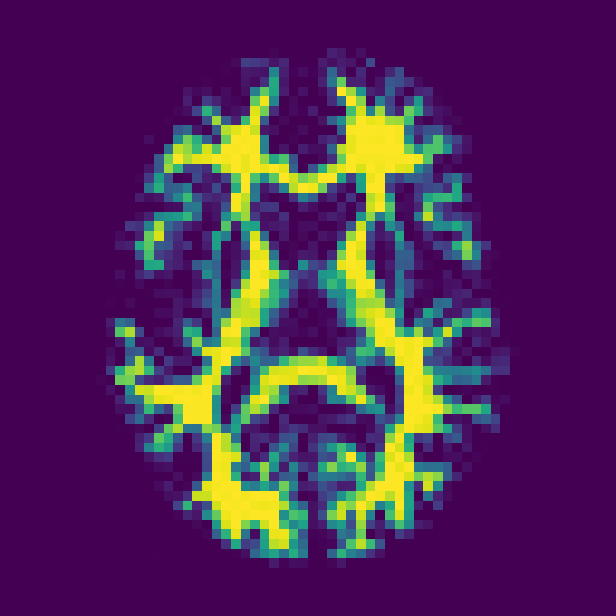}
    \\[\smallskipamount]
    \includegraphics[width=\size\textwidth]{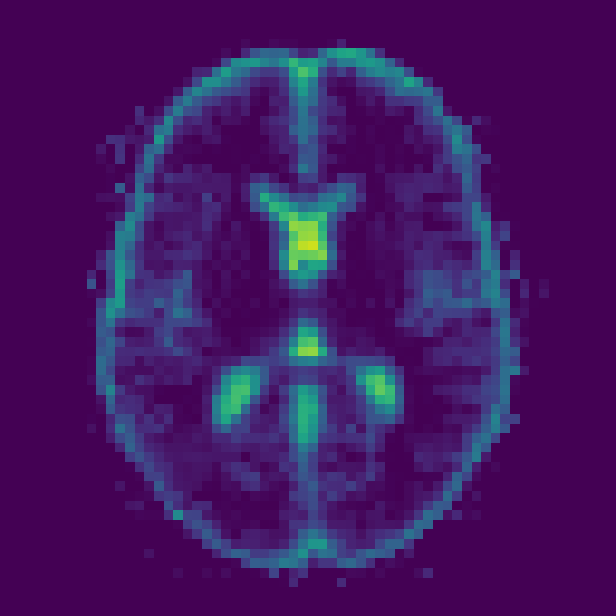}
    \includegraphics[width=\size\textwidth]{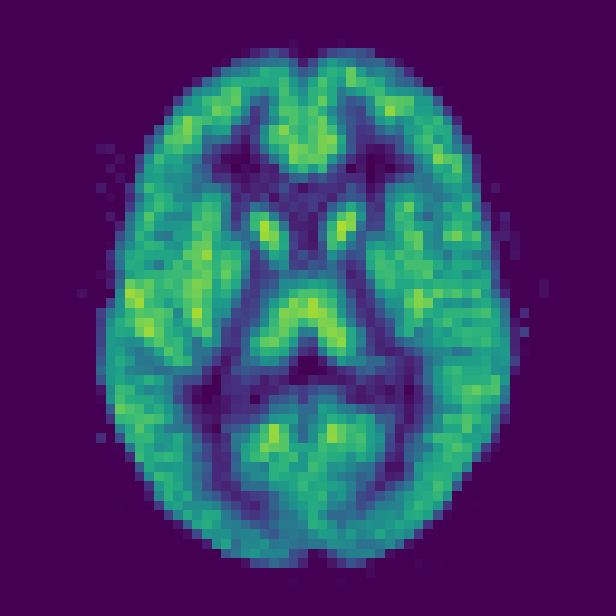}
    \includegraphics[width=\size\textwidth]{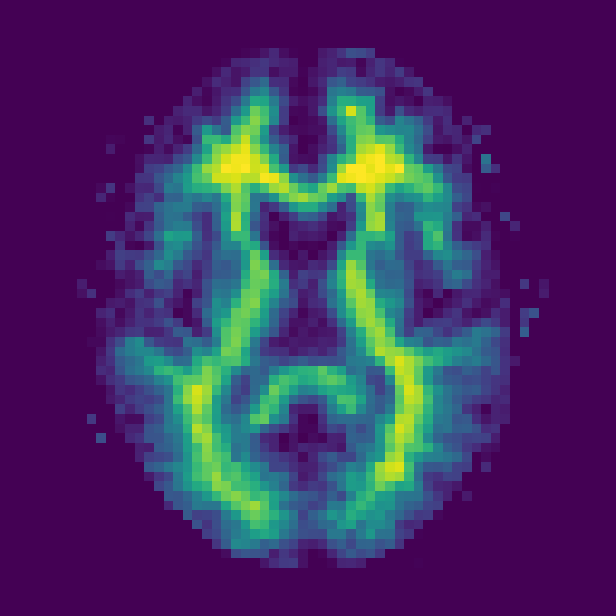}
    \caption{\small{From the left are CSF, GM, and WM. The first row shows the raw tissue probability maps for subject 42 in the BrainWeb dataset. The second row shows the estimated probability maps using our baseline configuration. The third row shows the ill-posed estimation using the single contrast from the $T_1$ inversion recovery sequence.}}
    \label{fig:subject42_best_results}
\end{figure}

the convergence, one can also run a pre-trained supervised deep-learning method to get an initial probability map estimate, which can then be further refined using our method.

\begin{table}
\vspace{0.75cm}
\centering
\caption{Dice score, PSNR, and SSIM comparisons with other clustering-based methods. \textdagger\ represents the baseline configuration.}
\begin{tabular}{||c c c c||}
 \hline
 Method & CSF & GM & WM \\ [0.5ex] 
 \hline\hline
\fcellsize{Our \textdagger} & \fcellsize{\textbf{0.55}$\pm$\textbf{0.04}} & \fcellsize{\textbf{0.76}$\pm$\textbf{0.02}} & \fcellsize{0.79$\pm$0.01} \\
\hline
\fcellsize{Double Estimation} & \fcellsize{0.41$\pm$0.31} & \fcellsize{0.60$\pm$0.37} & \fcellsize{0.80$\pm$0.23} \\
\hline
\fcellsize{LNLFCM} & \fcellsize{0.41$\pm$0.32} & \fcellsize{0.58$\pm$0.38} & \fcellsize{0.79$\pm$0.23} \\
\hline
\fcellsize{DCT-LNLFCM} & \fcellsize{0.41$\pm$0.31} & \fcellsize{0.60$\pm$0.36} & \fcellsize{\textbf{0.80}$\pm$\textbf{0.22}} \\
\hline
\fcellsize{KWFLICM} & \fcellsize{0.40$\pm$0.27} & \fcellsize{0.59$\pm$0.37} & \fcellsize{0.80$\pm$0.23} \\
\hline
\end{tabular}
\label{sup_tab:other_baseline_comparison}
\end{table}






\clearpage
\clearpage
\printbibliography

\newpage
\appendix
\onecolumn
\section{Extended Phase Graph}
\label{appendix:A}
Extended Phase Graphs (EPG) \cite{weigel2015extended} are a powerful way to simulate the magnetization response of various MR sequences. A key aspect of EPG involves the manipulation of magnetization vectors through changes in a Fourier basis. Initially, the longitudinal $M_z(z)$ and transverse $M_{xy}$ and $M^{*}_{xy}$ magnetization components are projected on a Fourier basis to obtain the Fourier components $F^{+}_{n}$, $F^{-}_{n}$, and $Z_{n}$: 

\[F^{+}_{n} = \int_{0}^{1}M_{xy}(z)e^{-2\pi\iota n z} dz\]
\[F^{-}_{n} = \int_{0}^{1}M^{*}_{xy}(z)e^{-2\pi\iota n z} dz\]
\[Z_n = \int_{0}^{1}M_z(z)e^{-2\pi\iota n z} dz\]
\[Q = \begin{bmatrix}
                F^{+}_{0} & F^{+}_{1} & F^{+}_{2} & ... & F^{+}_{N}\\
                F^{+*}_{0} & F^{-}_{1} & F^{-}_{2} & ... & F^{-}_{N}\\
                Z_{0} & Z_{1} & Z_{2} & ... & Z_{N}
            \end{bmatrix}\]

The $F$ states are then combined across $n$ harmonics to form the EPG basis $Q$ (shown above), which describes the magnetization state in one isochromat. Using the above EPG formalization, we can model the spin precession as a matrix operation:
\begin{equation}
    \begin{bmatrix}
        F^{+}_{n}\\
        F^{-}_{n}\\
        Z_{n}
    \end{bmatrix}^{'} = 
    \begin{bmatrix}
        e^{\iota\theta} & 0 & 0\\
        0 & e^{-\iota\theta} & 0\\
        0 & 0 & 1
    \end{bmatrix}
    \begin{bmatrix}
        F^{+}_{n}\\
        F^{-}_{n}\\
        Z_{n}
    \end{bmatrix}
    \label{precession}
\end{equation}

where $\theta$ is the accumulated phase. Similarly, an RF pulse that flips the magnetization longitudinally by $\alpha$ and transversally by $\phi$ as: 
\[
    \begin{bmatrix}
        F^{+}_{n}\\
        F^{-}_{n}\\
        Z_{n}
    \end{bmatrix}^{'} = 
    \begin{bmatrix}
        cos^2(\alpha/2) & e^{2\iota\phi} sin^2(\alpha/2) & -\iota e^{\iota\phi}sin(\alpha) \\
        e^{-2\iota\phi} sin^2(\alpha/2) & cos^2(\alpha/2) & \iota e^{-\iota\phi}sin(\alpha)\\
        -\frac{\iota}{2e^{-\iota\phi}sin(\alpha)} & \frac{\iota}{2e^{\iota\phi}sin(\alpha)} & cos(\alpha)
    \end{bmatrix}
    \begin{bmatrix}
        F^{+}_{n}\\
        F^{-}_{n}\\
        Z_{n}
    \end{bmatrix}
    \label{RFeqn}
\]

Also, the evolution of relaxation for Transverse states $F_n'(t)$ and Longitudinal states $Z_n(t)$ and $Z_0(t)$ follows the equations:
\[F_n'(t) = F_n e^{-t/T_2}\]
\[Z_n'(t) = Z_n e^{-t/T_1}\]
\[Z_0(t) = M_0 (1-e^{-t/T_1}) + Z_0 e^{{-t/T_1}}\]
Equivalent linear operations can also be used to model gradients. Therefore, an MRI clinical sequence of events (RF pulses, gradients, precession, relaxation, etc.) is simulated as a multiplication of $Q$ (representing the voxel's state) through several matrices corresponding to precession, RF rotations, relaxation operators, and gradients. Until recently, EPG could only describe echo amplitudes, but it was extended by Phase Distribution Graphs (PDG) \cite{Endres2023} to describe full echo shapes, including gradient echoes that provide spatial encoding.

\section{MRI Differentiable Simulator}
\label{appendix:B}
MR-zero \cite{loktyushin2021mrzero} is an MRI simulation tool that uses PDG to simulate the magnetization changes based on the MRI sequence events and input $T_1$, $T_2$, and Proton Density (PD) maps. MR-zero can simulate arbitrary MRI sequences, including clinical MRI sequences, such as Spin-Echo \cite{jung2013spin}, Gradient Recalled Echo \cite{markl2012gradient}, Rapid Acquisition with Relaxation Enhancement \cite{hennig1986rare}, and Fast Low Angle Shot \cite{haase1986flash}, etc. The simulation process in MR-zero comprises two passes: The first pass swiftly simulates numerous states to provide an initial signal estimate. This step identifies crucial states contributing to the signal and discards irrelevant ones. Secondly, the main pass utilizes the precise signal equation, considering all voxels but minimizing the number of simulated states based on information gleaned from the pre-pass. Doing so generates a meaningful signal suitable for reconstructing $T_1$/$T_2$-weighted scans as output. As opposed to previous simulators such as \cite{bittoun1984computer,summers1986computer,shkarin1996direct,shkarin1997time,kwan1999mri,yoder2004mri,benoit2005simri}, MR-zero is implemented in PyTorch and is differentiable by design. This allows us to incorporate MR-zero as part of the forward inference pipeline through which backpropagation can be performed, allowing one to efficiently optimize the input $T_1$/$T_2$/PD maps or the MRI sequence given a $T_1$/$T_2$-weighted image. 

\newcommand{\figsize}[0]{0.3}
\newcommand{\csfoffset}[0]{3.0cm}
\newcommand{\gmoffset}[0]{4.55cm}
\newcommand{\wmoffset}[0]{4.65cm}

\begin{figure}
    \textbf{\hspace{\csfoffset}CSF\hspace{\gmoffset}GM\hspace{\wmoffset}WM}\par\medskip
    \centering
    \includegraphics[width=\figsize\textwidth]{Figures/separate_plots/csf_gt.png}
    \includegraphics[width=\figsize\textwidth]{Figures/separate_plots/gm_gt.png}
    \includegraphics[width=\figsize\textwidth]{Figures/separate_plots/wm_gt.png}
    \\[\smallskipamount]
    \includegraphics[width=\figsize\textwidth]{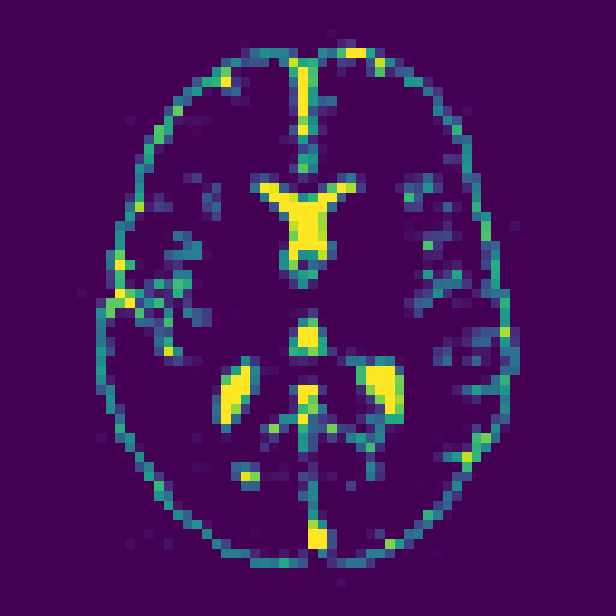}
    \includegraphics[width=\figsize\textwidth]{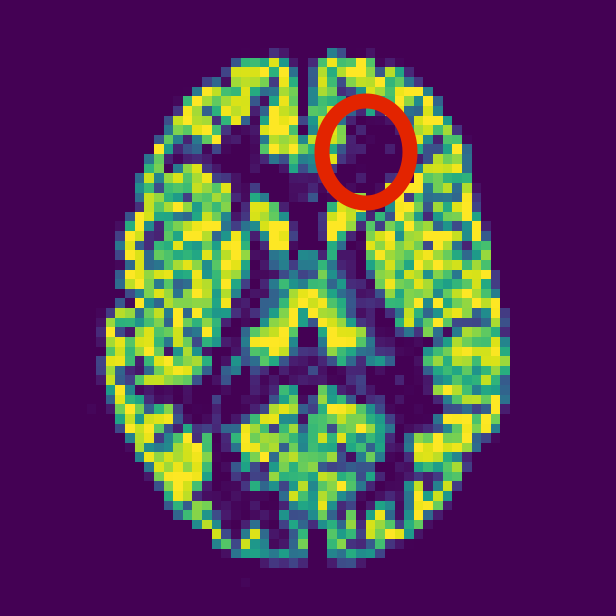}
    \includegraphics[width=\figsize\textwidth]{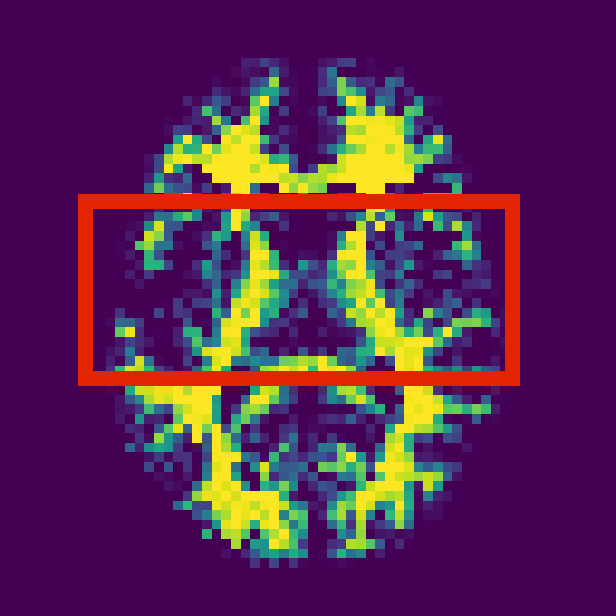}
    \caption{The first row shows the raw tissue probability maps for subject 42 in the BrainWeb dataset. The second row shows the checkerboard pattern enclosed in a red-marked region while optimizing 19 linear coefficients instead of using probability maps directly.}
    \label{sup_fig:subject_42_linear_coeff_best_results}
\end{figure}

\begin{table*}
\centering
\caption{Dice score, PSNR, and SSIM comparisons of different types of optimization procedures. \textdagger\ represents the baseline configuration.}
\begin{tabular}{ >{\centering\arraybackslash}m{1.73cm} | >{\centering}m{1.2cm}| >{\centering}m{1.4cm}| >{\centering}m{1.2cm}| >{\centering}m{1.2cm}| >{\centering}m{1.3cm}| >{\centering}m{1.2cm}| >{\centering}m{1.2cm}| >{\centering}m{1.35cm}| >{\centering\arraybackslash}m{1.1cm}}
    \hline
Method    & \multicolumn{3}{c|}{CSF}
            & \multicolumn{3}{c|}{GM}
                    & \multicolumn{3}{c}{WM} \\
    \cline{2-10}
                            & Dice$\uparrow$ & PSNR $\uparrow$  &   SSIM $\uparrow$  & Dice$\uparrow$ &  PSNR $\uparrow$  &   SSIM $\uparrow$  & Dice$\uparrow$ &  PSNR $\uparrow$  &   SSIM $\uparrow$  \\
                \hline
Direct pixel estimation \textdagger & \fcellsize{\textbf{0.55}$\pm$\textbf{0.04}} & \fcellsize{\textbf{34.45}$\pm$\textbf{4.87}} & \fcellsize{\textbf{0.98}$\pm$\textbf{0.02}} & \fcellsize{\textbf{0.76}$\pm$\textbf{0.02}} & \fcellsize{\textbf{26.8}$\pm$\textbf{2.54}} & \fcellsize{\textbf{0.96}$\pm$\textbf{0.02}} & \fcellsize{\textbf{0.79}$\pm$\textbf{0.01}} & \fcellsize{\textbf{25.99}$\pm$\textbf{2.69}} & \fcellsize{\textbf{0.92}$\pm$\textbf{0.03}} \\
\hline
Linear coefficients         & \fcellsize{0.54$\pm$0.05} &  \fcellsize{28.26$\pm$4.26} & \fcellsize{0.96$\pm$0.03} & \fcellsize{0.75$\pm$0.03} & \fcellsize{22.55$\pm$3.06} & \fcellsize{0.94$\pm$0.04} & \fcellsize{\textbf{0.79}$\pm$\textbf{0.01}} & \fcellsize{21.19$\pm$2.94} & \fcellsize{0.9$\pm$0.05} \\
    \hline
Scalar coefficient          & \fcellsize{0.32$\pm$0.09} & \fcellsize{16.46$\pm$1.1} & \fcellsize{0.42$\pm$0.09} & \fcellsize{0.66$\pm$0.05} & \fcellsize{13.69$\pm$1.01} & \fcellsize{0.51$\pm$0.1} & \fcellsize{0.63$\pm$0.04} & \fcellsize{13.92$\pm$1.09} & \fcellsize{0.53$\pm$0.07} \\
\hline
\end{tabular}
\label{sup_tab:loss_table_2}
\end{table*}

\begin{table*}
\centering
\caption{Dice score, PSNR, and SSIM comparisons on optimizing either all maps concurrently or only a single probability map given the other two fixed. \textdagger\ represents the baseline configuration.}
\begin{tabular}{>{\centering\arraybackslash}m{.95cm} | >{\centering}m{1.2cm}| >{\centering}m{1.5cm}| >{\centering}m{1.22cm}| >{\centering}m{1.2cm}| >{\centering}m{1.5cm}| >{\centering}m{1.22cm}| >{\centering}m{1.2cm}| >{\centering}m{1.5cm}| >{\centering\arraybackslash}m{1.3cm}}
    \hline
Maps    & \multicolumn{3}{c|}{CSF}
            & \multicolumn{3}{c|}{GM}
                    & \multicolumn{3}{c}{WM} \\
    \cline{2-10}
                            &  Dice$\uparrow$ & PSNR $\uparrow$  &   SSIM $\uparrow$  & Dice$\uparrow$ &  PSNR $\uparrow$  &   SSIM $\uparrow$  & Dice$\uparrow$ &  PSNR $\uparrow$  &   SSIM $\uparrow$  \\
                \hline
All 3 \textdagger          & \fcellsize{0.55$\pm$0.04} & \fcellsize{34.45$\pm$4.87} & \fcellsize{0.98$\pm$0.02} & \fcellsize{0.76$\pm$0.02} & \fcellsize{26.8$\pm$2.54} & \fcellsize{0.96$\pm$0.02} & \fcellsize{0.79$\pm$0.01} & \fcellsize{25.99$\pm$2.69} & \fcellsize{0.92$\pm$0.03} \\
\hline
CSF         & \fcellsize{\textbf{0.56}$\pm$\textbf{0.03}} & \fcellsize{\textbf{40.34}$\pm$\textbf{10.73}} & \fcellsize{\textbf{0.99}$\pm$\textbf{0.01}} & \fcellsize{-} & \fcellsize{-} & \fcellsize{-} & \fcellsize{-} & \fcellsize{-} & \fcellsize{-} \\
    \hline
GM          & \fcellsize{-} & \fcellsize{-} & \fcellsize{-} & \fcellsize{\textbf{0.78}$\pm$\textbf{0.02}} & \fcellsize{\textbf{47.23}$\pm$\textbf{10.43}} & \fcellsize{\textbf{0.99}$\pm$\textbf{0.01}} & \fcellsize{-} & \fcellsize{-} & \fcellsize{-} \\
\hline
WM          & \fcellsize{-} & \fcellsize{-} & \fcellsize{-} & \fcellsize{-} & \fcellsize{-} & \fcellsize{-} & \fcellsize{\textbf{0.8}$\pm$\textbf{0.01}} & \fcellsize{\textbf{43.33}$\pm$\textbf{10.65}} & \fcellsize{\textbf{0.99}$\pm$\textbf{0.01}} \\
\hline
\end{tabular}
\label{tab:loss_table_4}
\end{table*}

\begin{figure}
    \textbf{\hspace{\csfoffset}CSF\hspace{\gmoffset}GM\hspace{\wmoffset}WM}\par\medskip
    \centering
    \includegraphics[width=\figsize\textwidth]{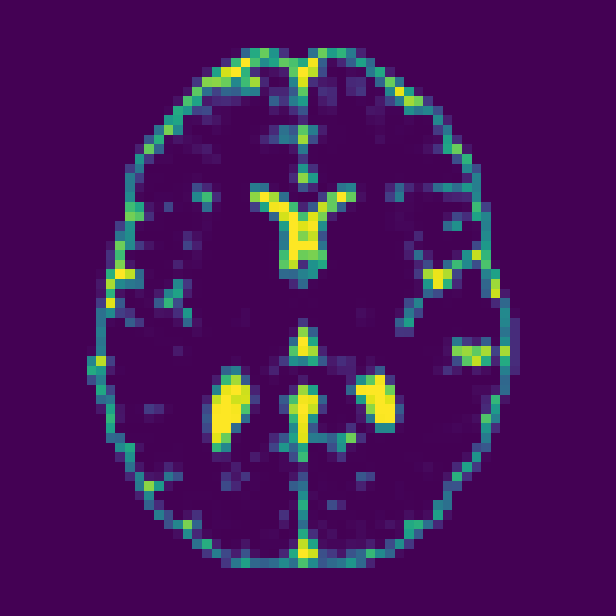}
    \includegraphics[width=\figsize\textwidth]{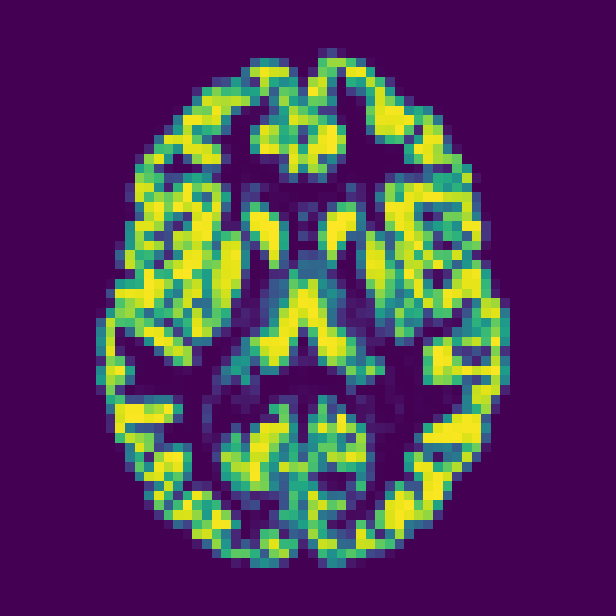}
    \includegraphics[width=\figsize\textwidth]{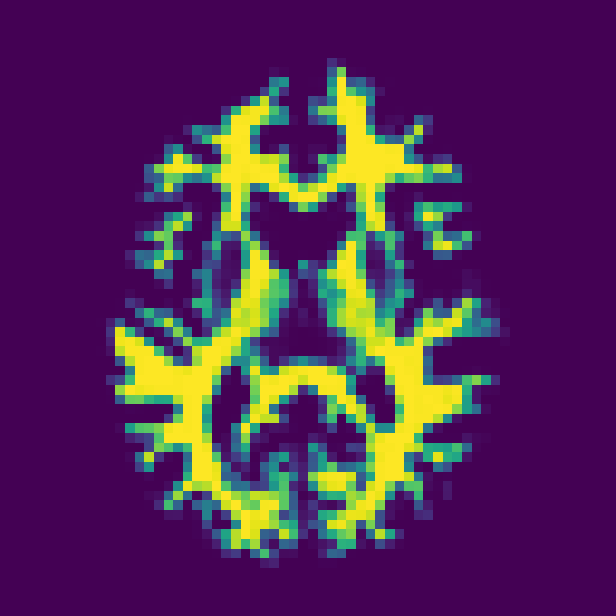}
    \\[\smallskipamount]
    \includegraphics[width=\figsize\textwidth]{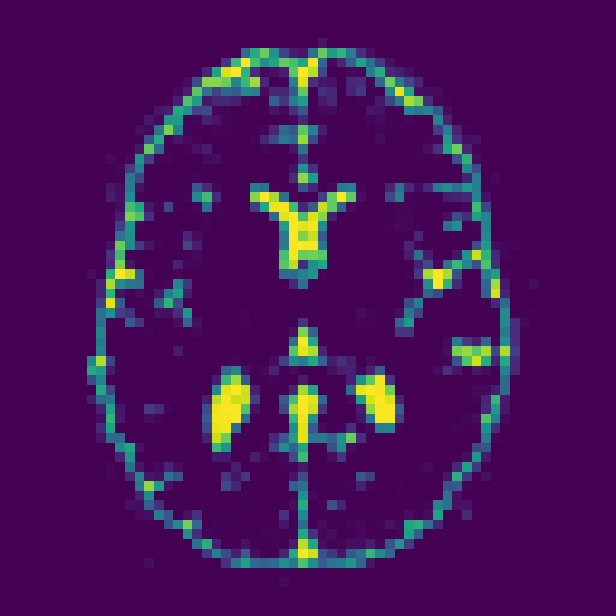}
    \includegraphics[width=\figsize\textwidth]{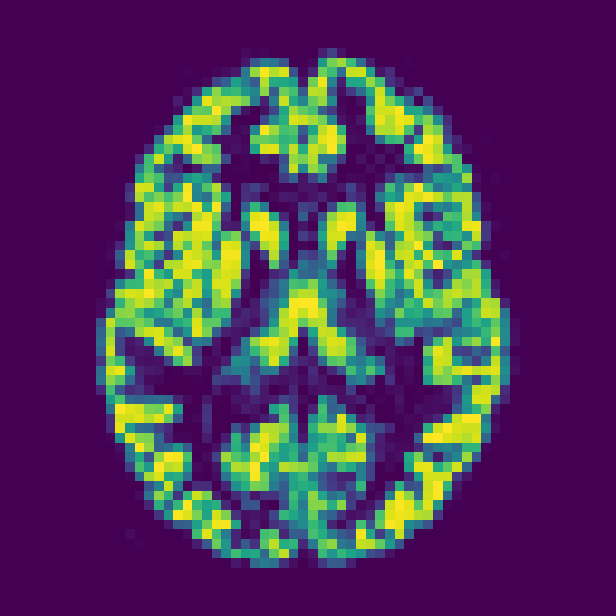}
    \includegraphics[width=\figsize\textwidth]{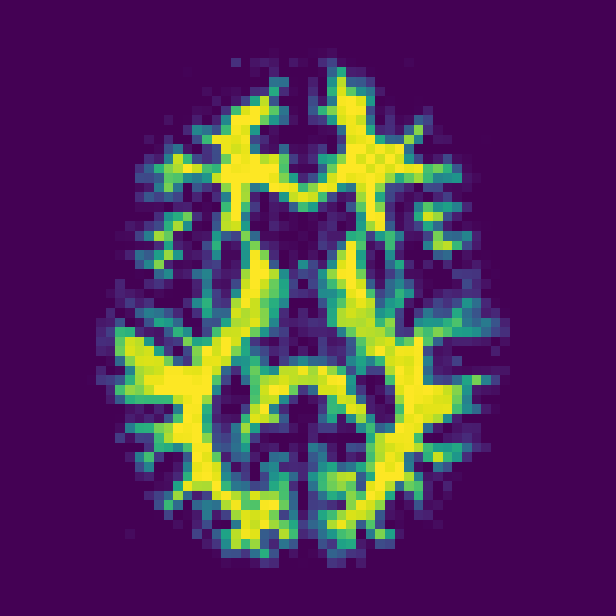}
    \\[\smallskipamount]
    \includegraphics[width=\figsize\textwidth]{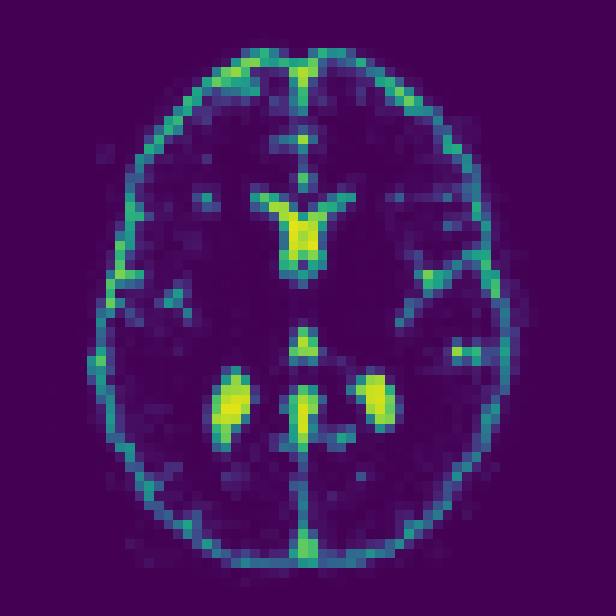}
    \includegraphics[width=\figsize\textwidth]{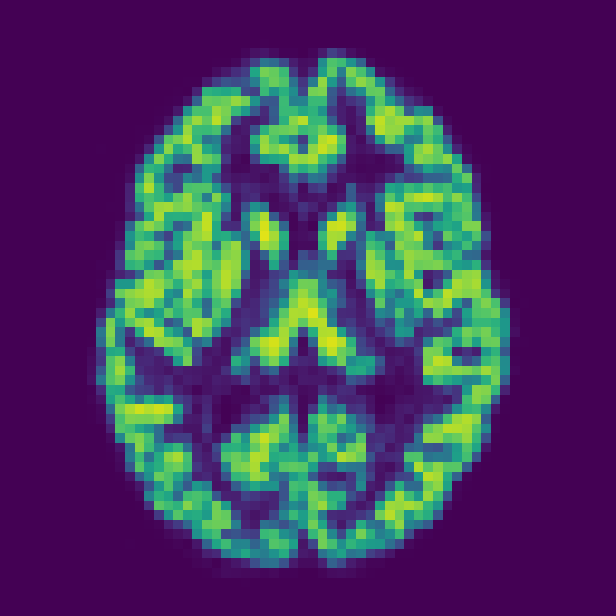}
    \includegraphics[width=\figsize\textwidth]{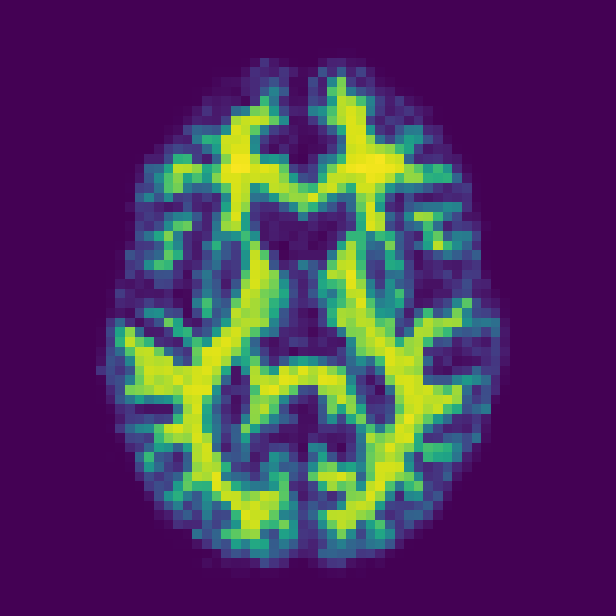}
    \\[\smallskipamount]
    \includegraphics[width=\figsize\textwidth]{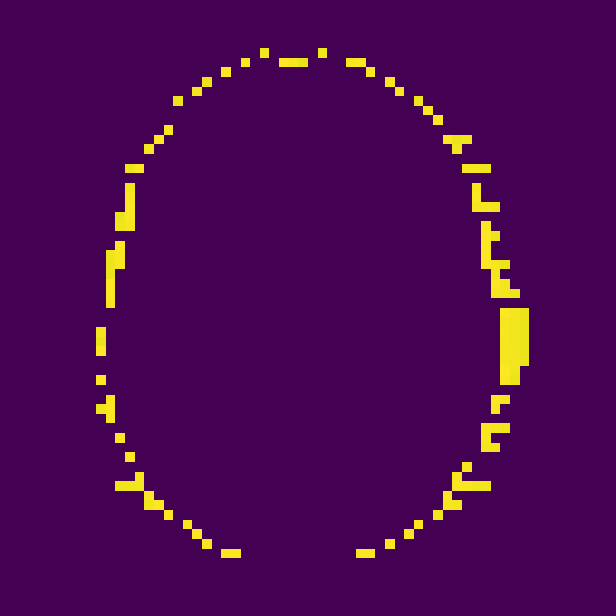}
    \includegraphics[width=\figsize\textwidth]{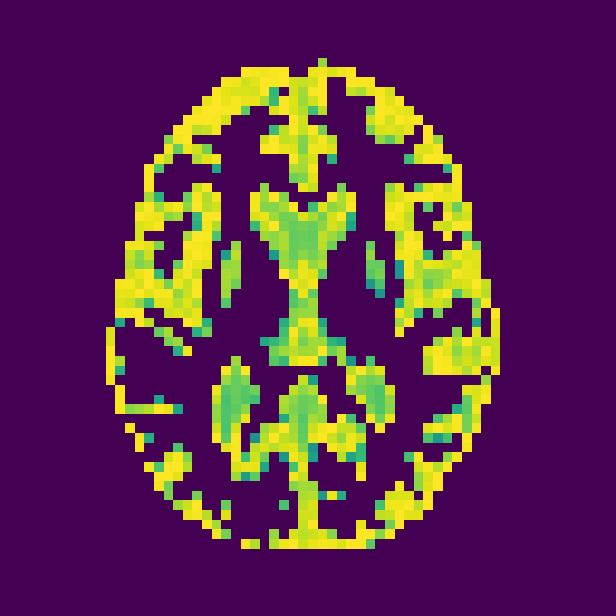}
    \includegraphics[width=\figsize\textwidth]{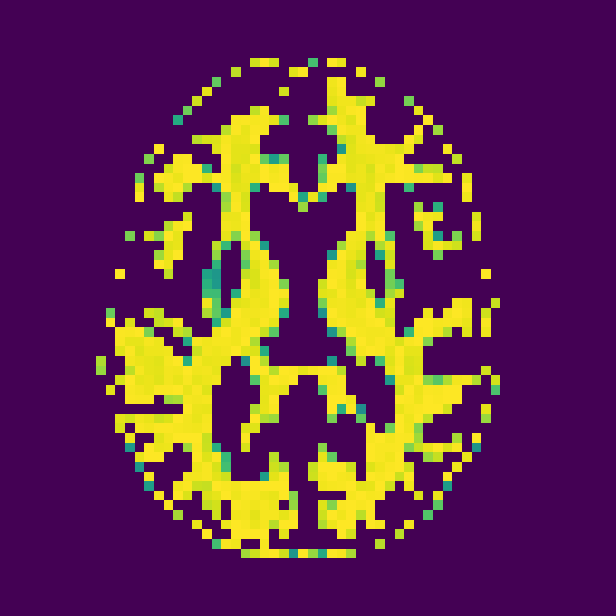}
    \caption{From the left are CSF, GM, and WM. The first row shows the ground truth probability maps for subject 20 of BrainWeb. The second row shows the estimated map using our baseline method. The third and fourth row shows the inferred probability maps using U-Net and BCEFCM.}
    \label{sup_fig:optimization_results}
\end{figure}

\begin{table*}
\centering
\caption{Dice score, PSNR, and SSIM comparisons on tissue probability maps while computing loss in Image-space and K-space domain. \textdagger\ represents the baseline configuration.}
\begin{tabular}{>{\centering\arraybackslash}m{1.25cm} | >{\centering}m{1.2cm}| >{\centering}m{1.4cm}| >{\centering}m{1.22cm}| >{\centering}m{1.2cm}| >{\centering}m{1.4cm}| >{\centering}m{1.22cm}| >{\centering}m{1.2cm}| >{\centering}m{1.4cm}| >{\centering\arraybackslash}m{1.3cm}}
    \hline
Maps    & \multicolumn{3}{c|}{CSF}
            & \multicolumn{3}{c|}{GM}
                    & \multicolumn{3}{c}{WM} \\
    \cline{2-10}
                            &  Dice$\uparrow$ & PSNR $\uparrow$  &   SSIM $\uparrow$  & Dice$\uparrow$ &  PSNR $\uparrow$  &   SSIM $\uparrow$  & Dice$\uparrow$ &   PSNR $\uparrow$  &   SSIM $\uparrow$  \\
                \hline
Image-space \textdagger          & \fcellsize{\textbf{0.55}$\pm$\textbf{0.04}} & \fcellsize{\textbf{34.45}$\pm$\textbf{4.87}} & \fcellsize{\textbf{0.98}$\pm$\textbf{0.02}} & \fcellsize{\textbf{0.76}$\pm$\textbf{0.02}} & \fcellsize{\textbf{26.8}$\pm$\textbf{2.54}} & \fcellsize{\textbf{0.96}$\pm$\textbf{0.02}} & \fcellsize{\textbf{0.79}$\pm$\textbf{0.01}} & \fcellsize{\textbf{25.99}$\pm$\textbf{2.69}} & \fcellsize{\textbf{0.92}$\pm$\textbf{0.03}} \\
\hline
K-space                          & \fcellsize{0.48$\pm$0.11} & \fcellsize{23.16$\pm$8.15} & \fcellsize{0.71$\pm$0.27} & \fcellsize{0.72$\pm$0.07} & \fcellsize{18.21$\pm$5.02} & \fcellsize{0.75$\pm$0.21} & \fcellsize{0.74$\pm$0.06} & \fcellsize{18.11$\pm$4.4} & \fcellsize{0.72$\pm$0.17} \\
    \hline
\end{tabular}
\label{tab:loss_table_1}
\end{table*}

\section{Results}
\label{appendix:C}
\subsection{Direct pixel optimization} We optimized probability values directly and got the PSNRs and SSIMs for CSF, GM, and WM, respectively (Table \ref{sup_tab:loss_table_2}). The direct pixel optimization outperforms all the other methods. One of those methods includes optimizing either 1 coefficient (\emph{scalar optimization}) or 19 coefficients (\emph{linear coefficients}) per pixel. The PSNR and SSIM values for coefficient optimization failed to surpass those of the baseline configuration and are off by a margin of $5.09$ and $0.02$ points, respectively, on average for all the maps. Additionally, examination of the results depicted in Fig \ref{sup_fig:subject_42_linear_coeff_best_results} revealed a checkerboard pattern present in the optimized probability maps generated using this technique. Also, we tried optimizing a scalar coefficient per probability map, with all the contrasts of six sequences, which performed poorly compared to all the methods by being $14.39$ and $0.47$ points lower on average, respectively.

\subsection{All 3 maps vs. single map} In Table \ref{tab:loss_table_4}, we report DICE, PSNR, and SSIM metrics for single map optimization (e.g., only optimize CSF while GM and WM are assumed to be known). For all tissues, single map optimization has outperformed the optimization of all three maps concurrently. Single map optimization is significantly easier because the two other maps are assumed to be known and fixed, thus acting as a strong prior for estimating the third map.

\subsection{Image-space vs. K-space loss} In Table \ref{tab:loss_table_1}, we report comparisons between an image-space loss and a k-space loss, i.e. computing the loss directly at k-space values, without applying the inverse Fourier transform.  The PSNR and SSIM values of estimated probability maps while computing the loss in K-space domain are $(0.48\pm0.11, 23.16\pm8.15, 0.71\pm0.27)$, $(0.72\pm0.07, 18.21\pm5.02, 0.75\pm0.21)$, and $(0.74\pm0.06, 18.11\pm4.4, 0.72\pm0.17)$ for CSF, GM, and WM respectively. These values, on average, are off by a huge margin of $9.25$ and $0.23$ points for PSNR and SSIM, respectively, from the baseline configuration metrics of the experiment done in Image-space. We found these results surprising since the pipeline with the k-space loss has one step less than the image-space pipeline. We plan to explore this more in our future work.

\section{More probability maps estimation results}
\label{appendix:D}
This section of the Appendix contains the brain probability maps optimization for all other subjects of the BrainWeb dataset.

\begin{figure}
    \textbf{\hspace{\csfoffset}CSF\hspace{\gmoffset}GM\hspace{\wmoffset}WM}\par\medskip
    \centering
    \includegraphics[width=\figsize\textwidth]{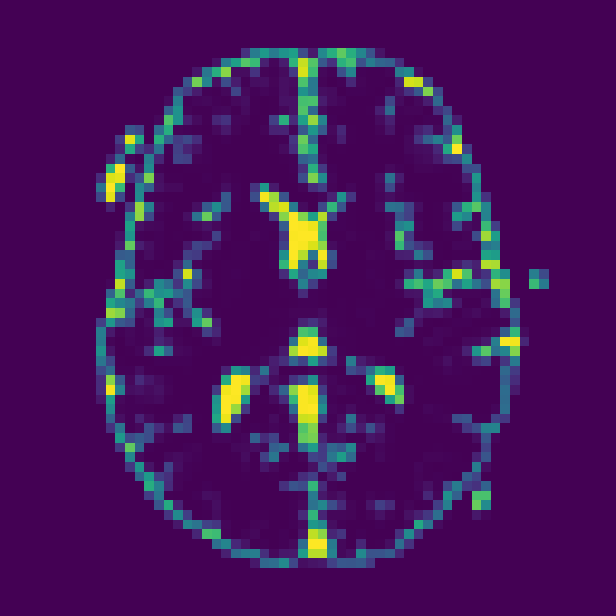}
    \includegraphics[width=\figsize\textwidth]{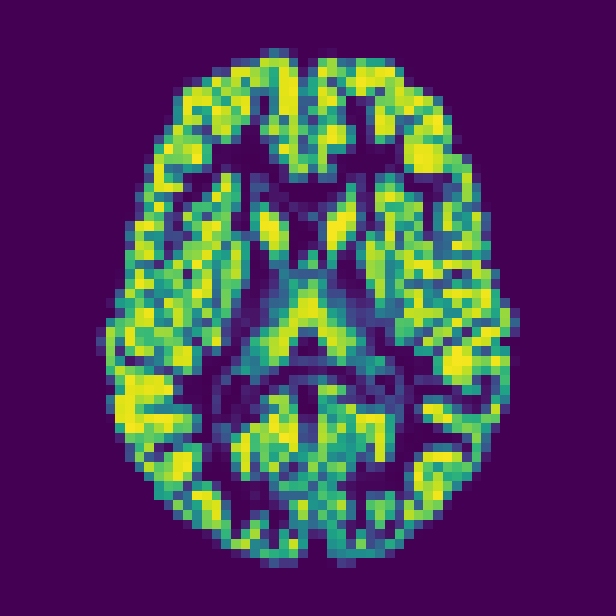}
    \includegraphics[width=\figsize\textwidth]{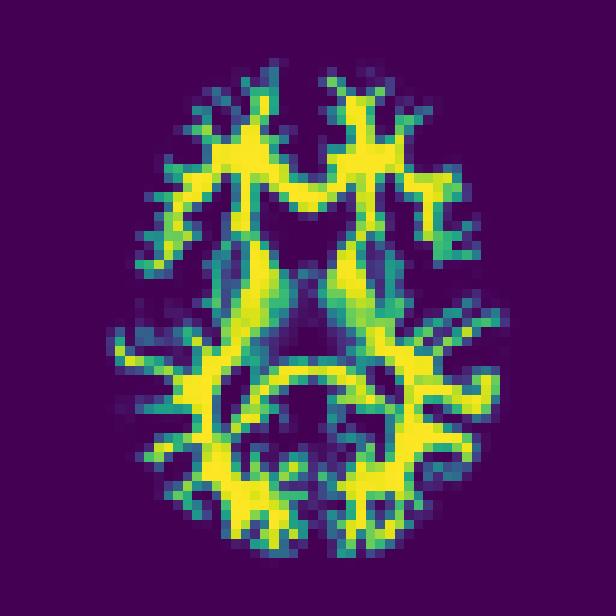}
    \\[\smallskipamount]
    \includegraphics[width=\figsize\textwidth]{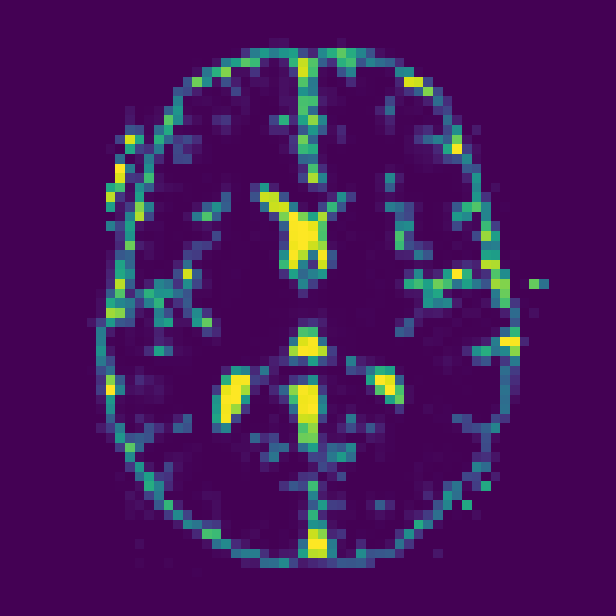}
    \includegraphics[width=\figsize\textwidth]{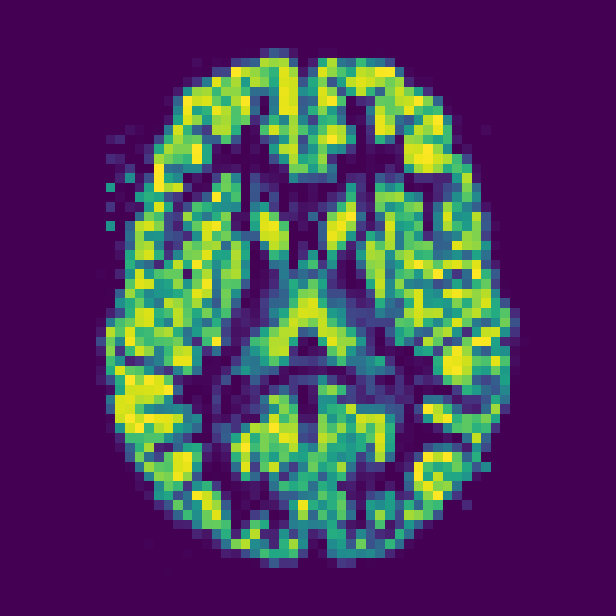}
    \includegraphics[width=\figsize\textwidth]{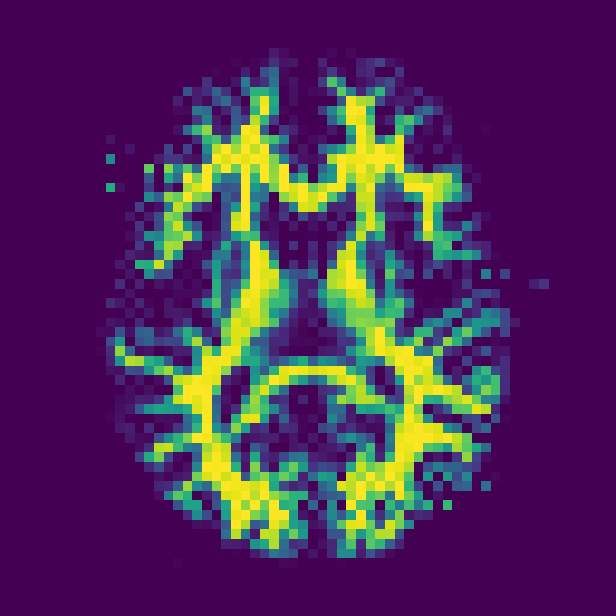}
    \\[\smallskipamount]
    \includegraphics[width=\figsize\textwidth]{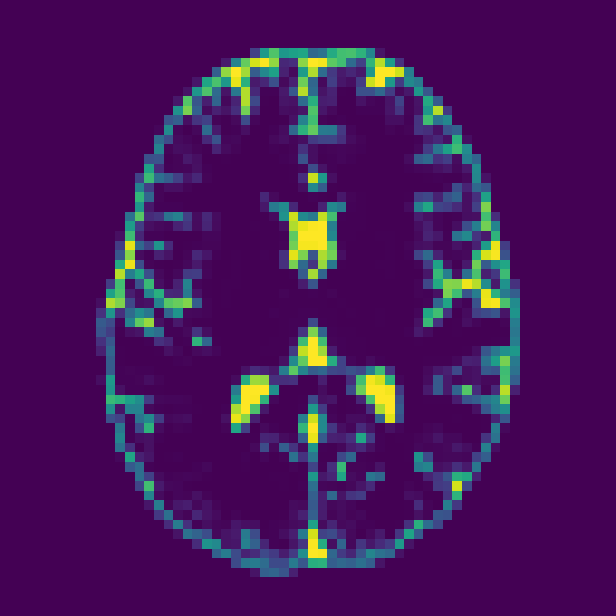}
    \includegraphics[width=\figsize\textwidth]{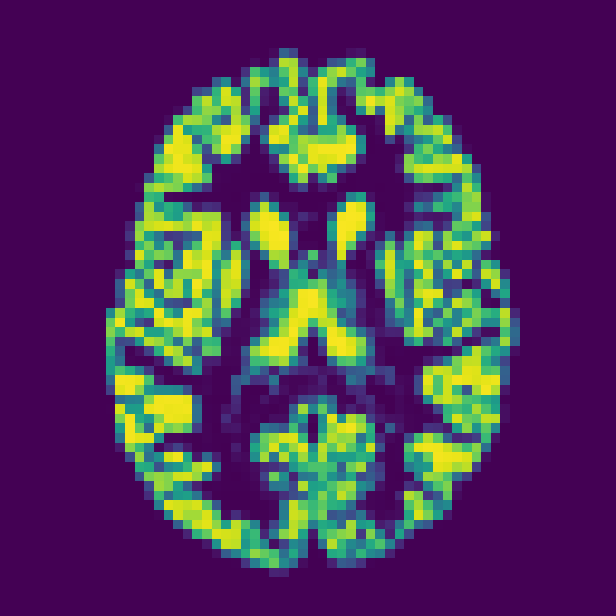}
    \includegraphics[width=\figsize\textwidth]{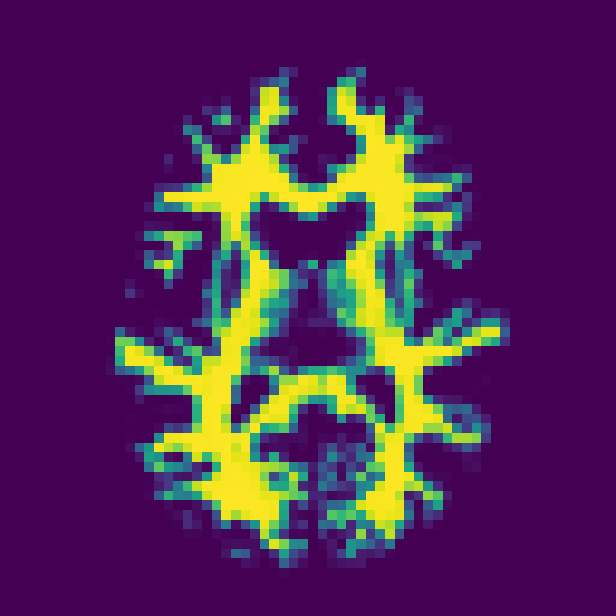}
    \\[\smallskipamount]
    \includegraphics[width=\figsize\textwidth]{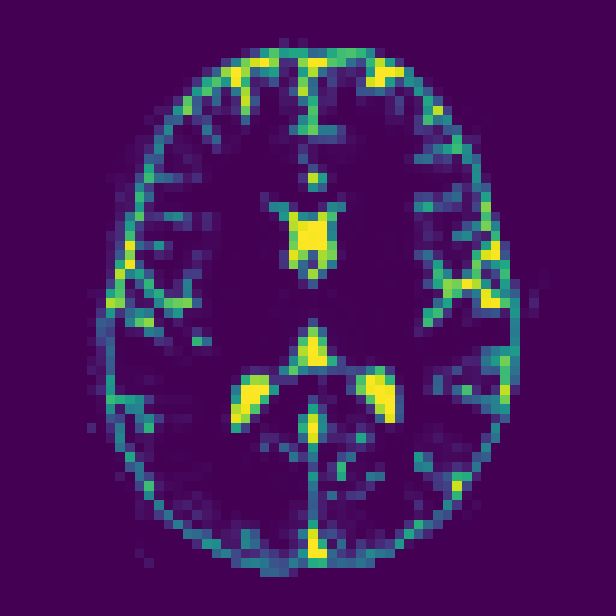}
    \includegraphics[width=\figsize\textwidth]{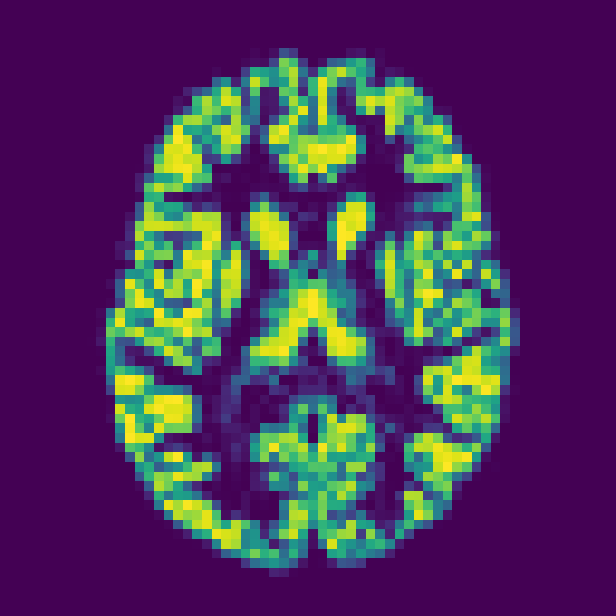}
    \includegraphics[width=\figsize\textwidth]{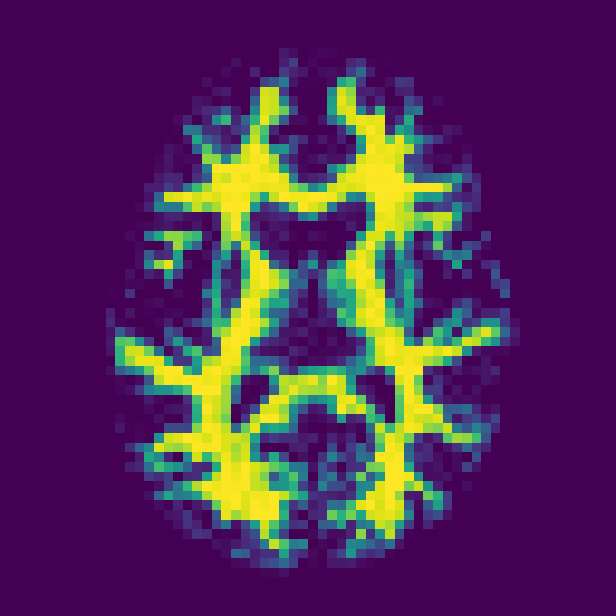}
    \caption{From the left are CSF, GM, and WM. The first and third rows show the raw tissue probability maps for subjects 04 and 05, respectively, of the BrainWeb dataset. The second and fourth rows show the optimized probability maps using the baseline configuration for subjects 04 and 05, respectively.}
    \label{sup_fig:subject_04_05_best_results}
\end{figure}

\begin{figure}
    \textbf{\hspace{\csfoffset}CSF\hspace{\gmoffset}GM\hspace{\wmoffset}WM}\par\medskip
    \centering
    \includegraphics[width=\figsize\textwidth]{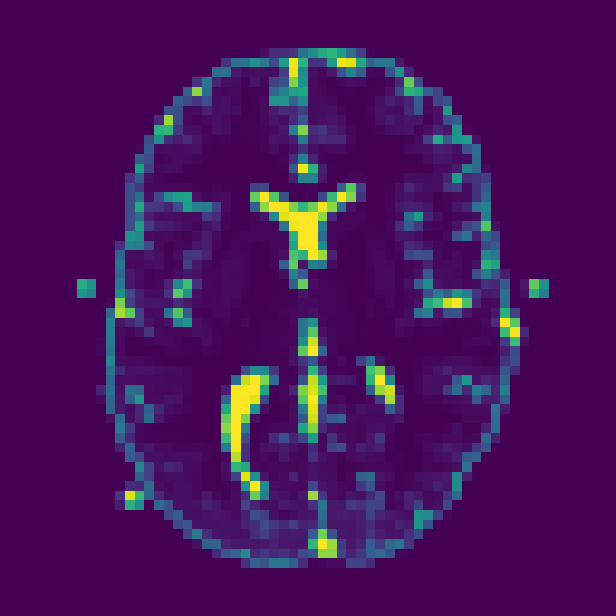}
    \includegraphics[width=\figsize\textwidth]{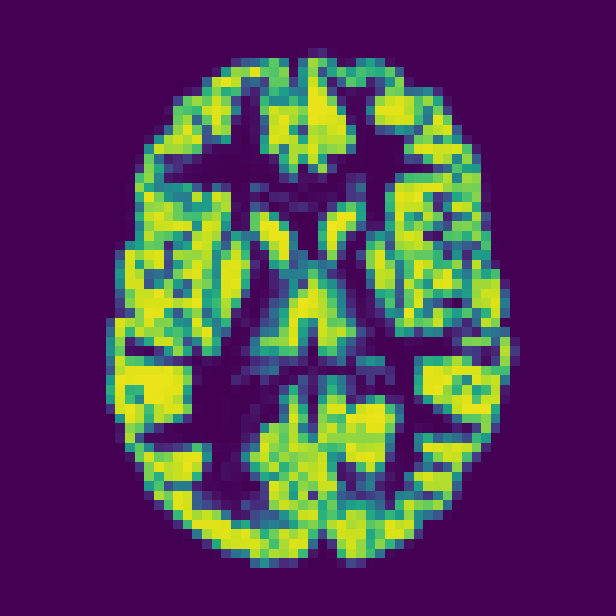}
    \includegraphics[width=\figsize\textwidth]{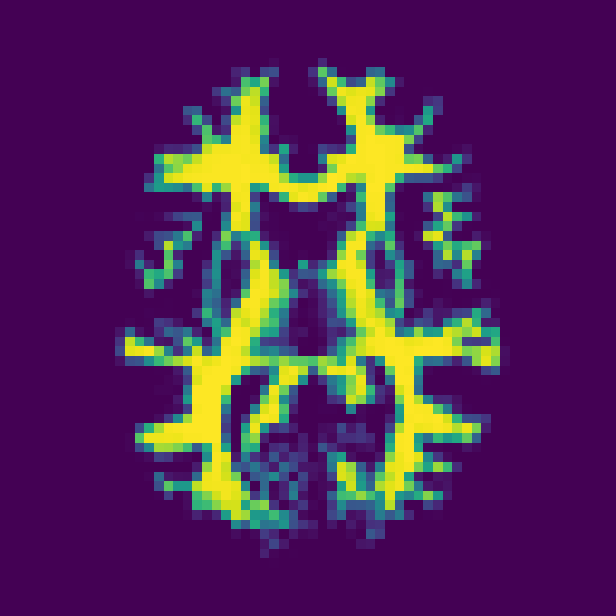}
    \\[\smallskipamount]
    \includegraphics[width=\figsize\textwidth]{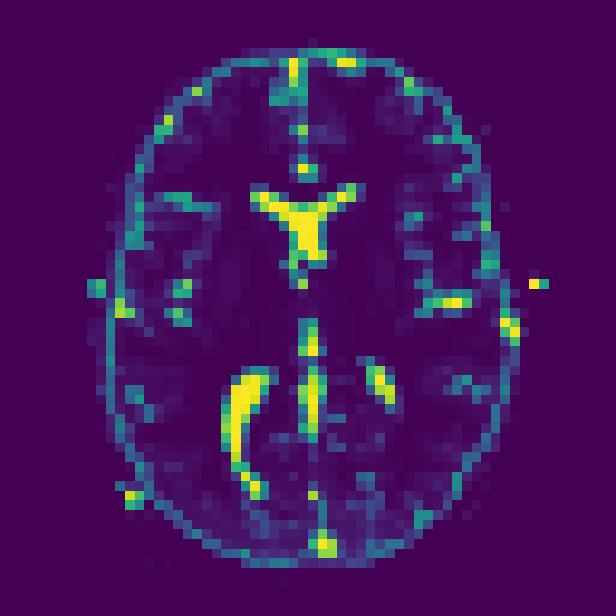}
    \includegraphics[width=\figsize\textwidth]{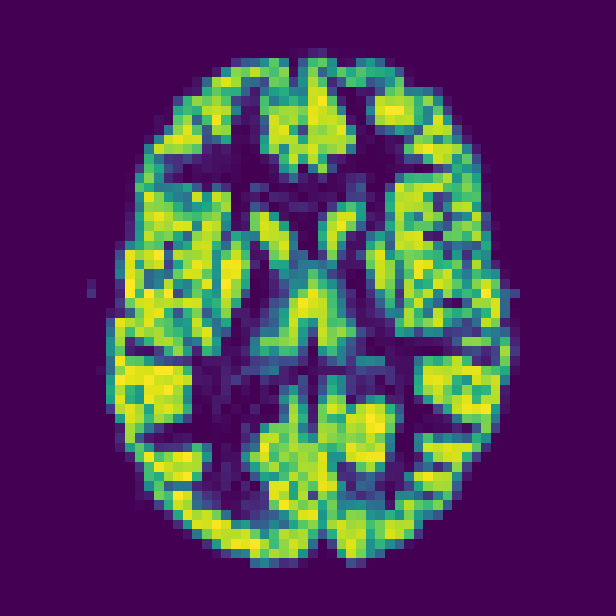}
    \includegraphics[width=\figsize\textwidth]{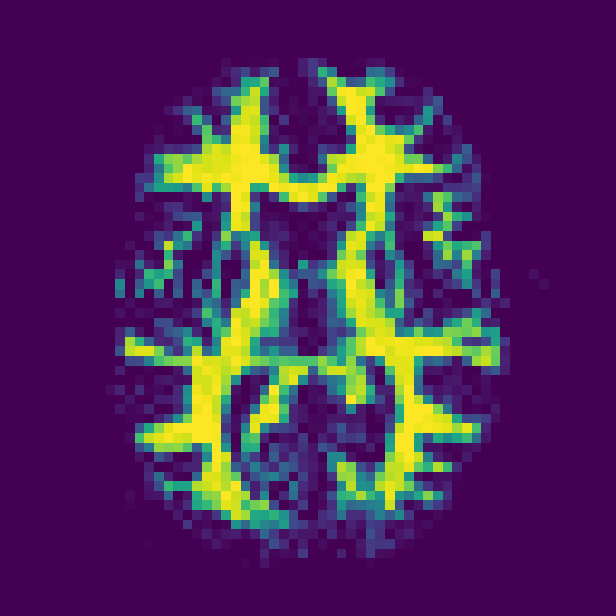}
    \\[\smallskipamount]
    \includegraphics[width=\figsize\textwidth]{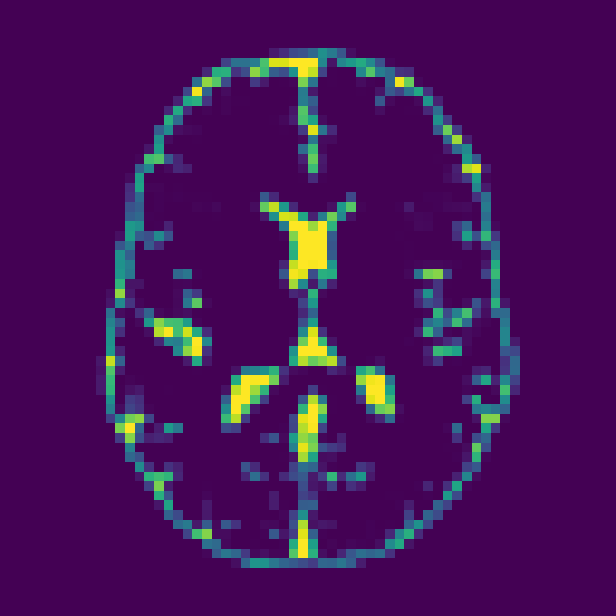}
    \includegraphics[width=\figsize\textwidth]{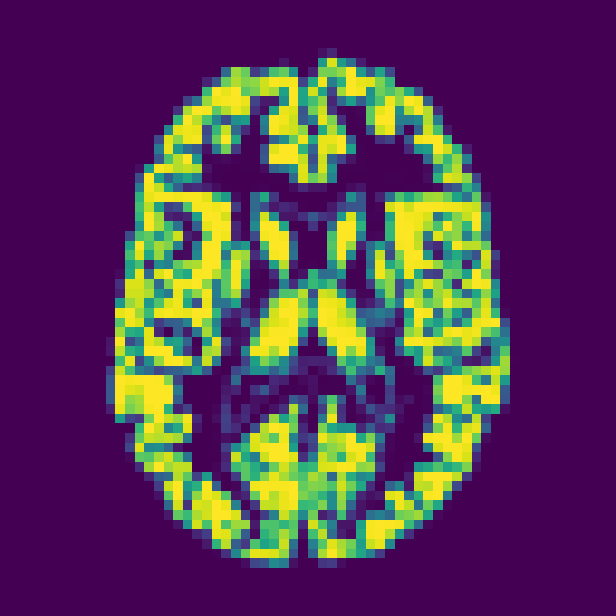}
    \includegraphics[width=\figsize\textwidth]{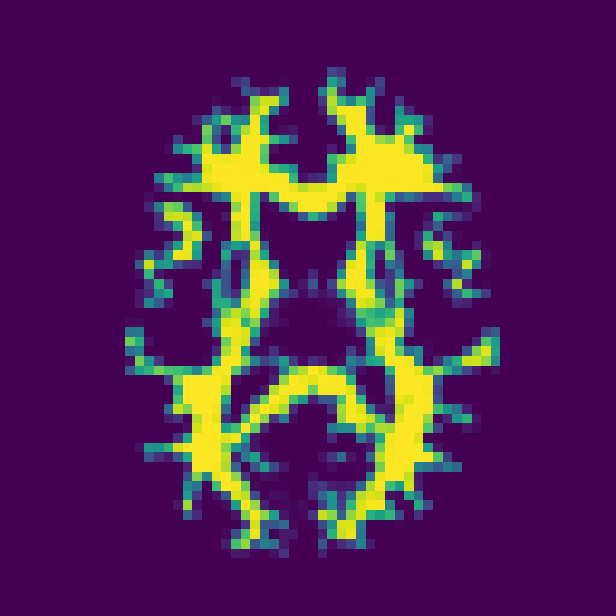}
    \\[\smallskipamount]
    \includegraphics[width=\figsize\textwidth]{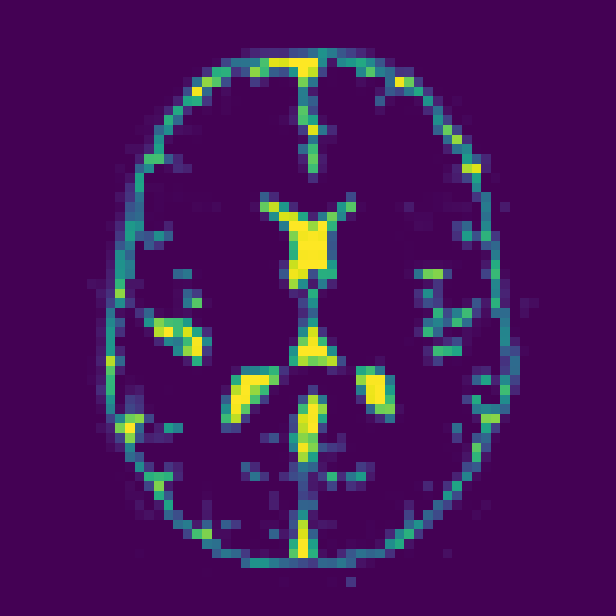}
    \includegraphics[width=\figsize\textwidth]{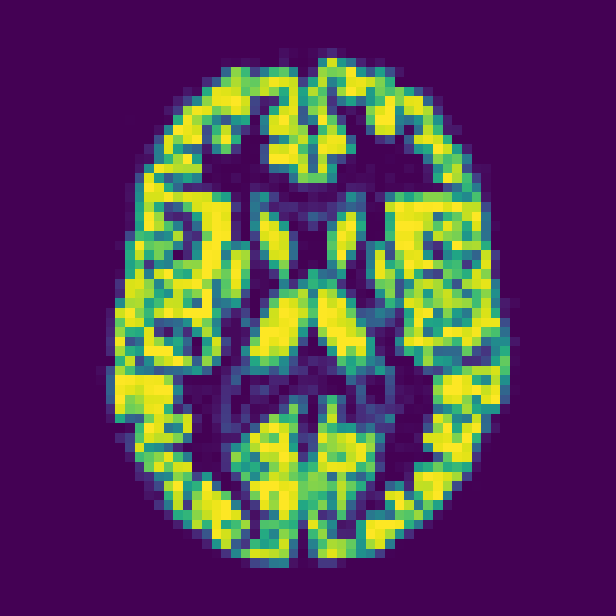}
    \includegraphics[width=\figsize\textwidth]{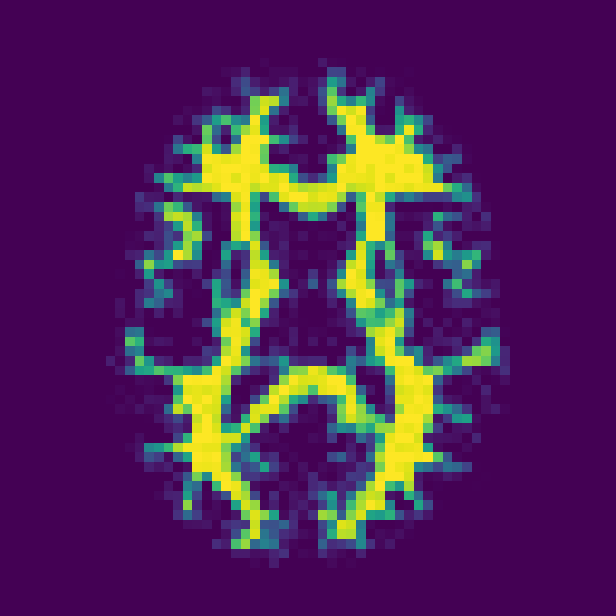}
    \caption{From the left are CSF, GM, and WM. The first and third rows show the raw tissue probability maps for subjects 06 and 18, respectively, of the BrainWeb dataset. The second and fourth rows show the optimized probability maps using the baseline configuration for subjects 06 and 18, respectively.}
    \label{sup_fig:subject_06_18_best_results}
\end{figure}

\begin{figure}
    \textbf{\hspace{\csfoffset}CSF\hspace{\gmoffset}GM\hspace{\wmoffset}WM}\par\medskip
    \centering
    \includegraphics[width=\figsize\textwidth]{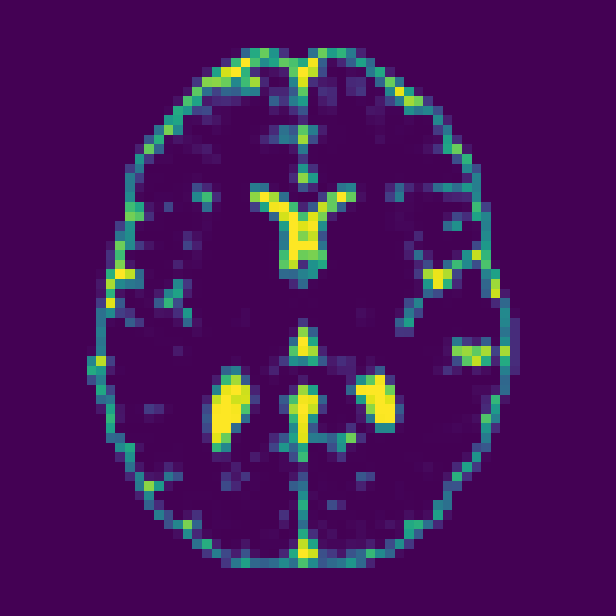}
    \includegraphics[width=\figsize\textwidth]{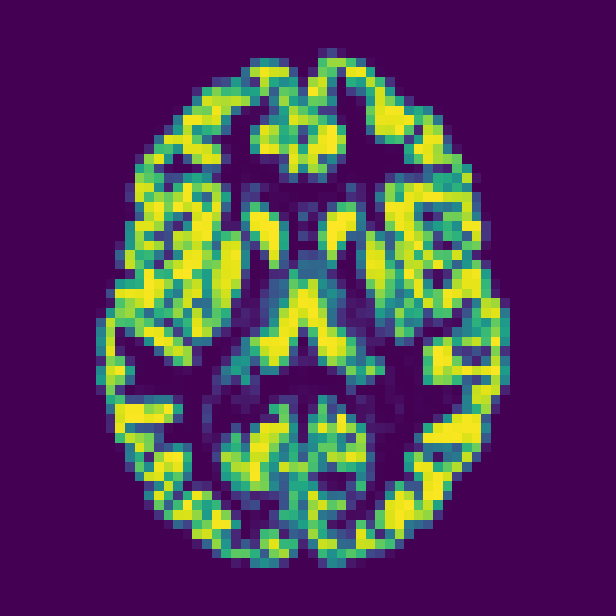}
    \includegraphics[width=\figsize\textwidth]{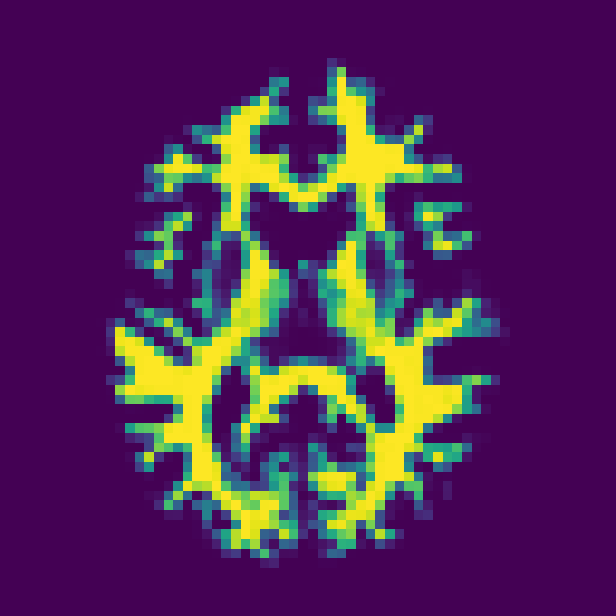}
    \\[\smallskipamount]
    \includegraphics[width=\figsize\textwidth]{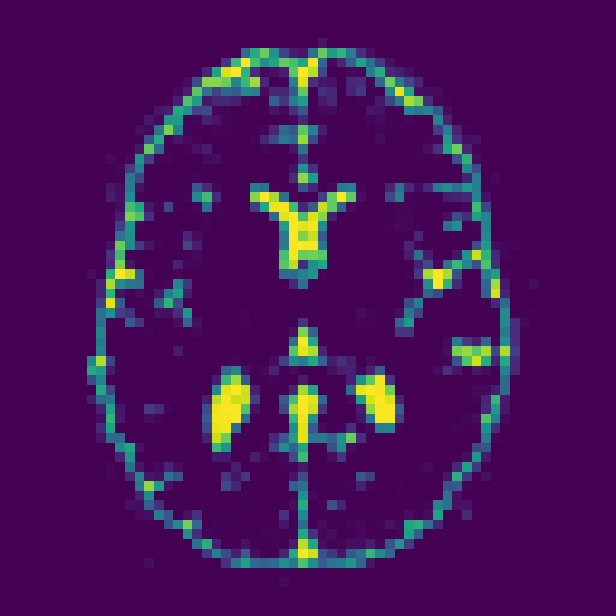}
    \includegraphics[width=\figsize\textwidth]{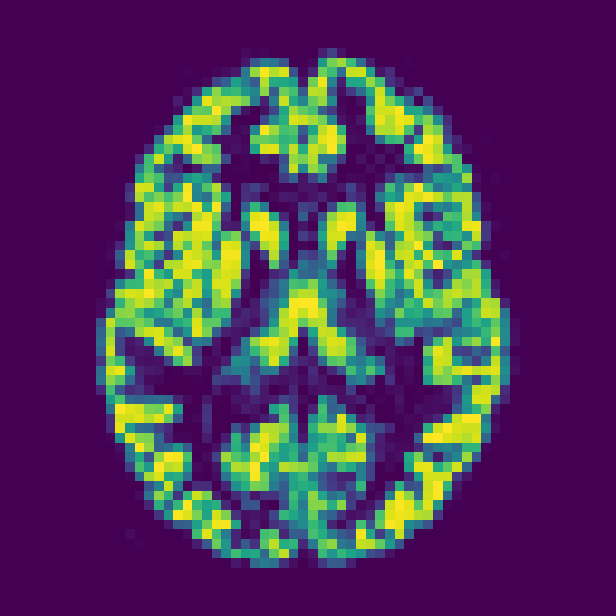}
    \includegraphics[width=\figsize\textwidth]{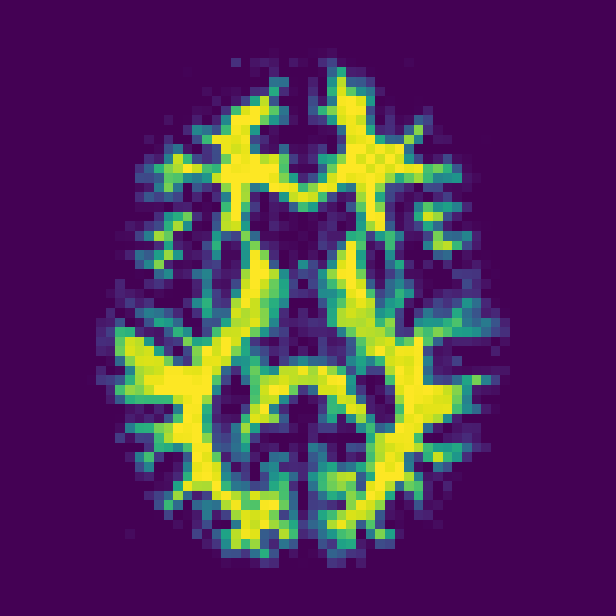}
    \\[\smallskipamount]
    \includegraphics[width=\figsize\textwidth]{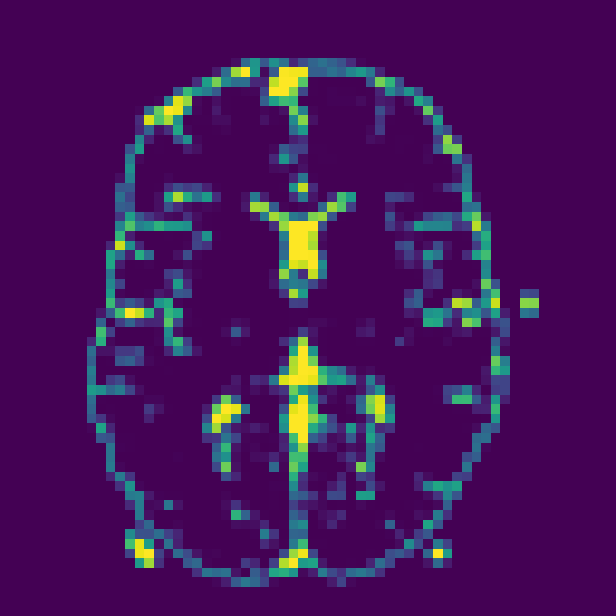}
    \includegraphics[width=\figsize\textwidth]{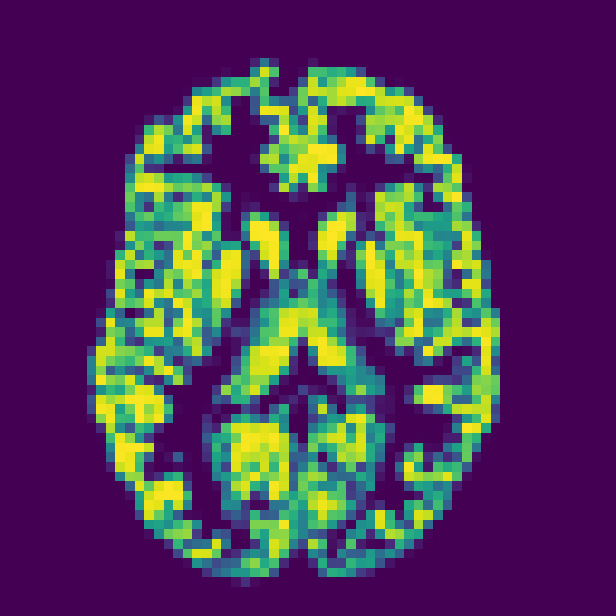}
    \includegraphics[width=\figsize\textwidth]{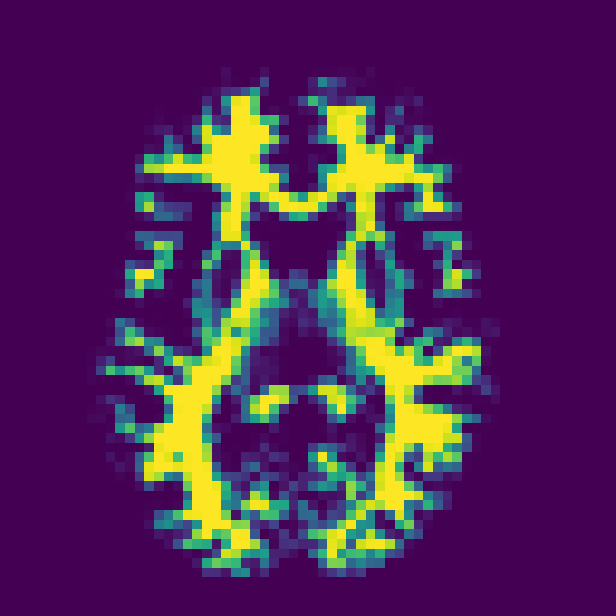}
    \\[\smallskipamount]
    \includegraphics[width=\figsize\textwidth]{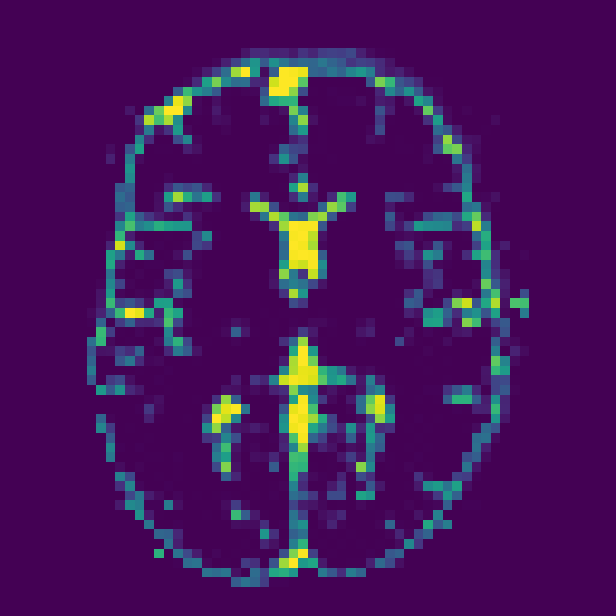}
    \includegraphics[width=\figsize\textwidth]{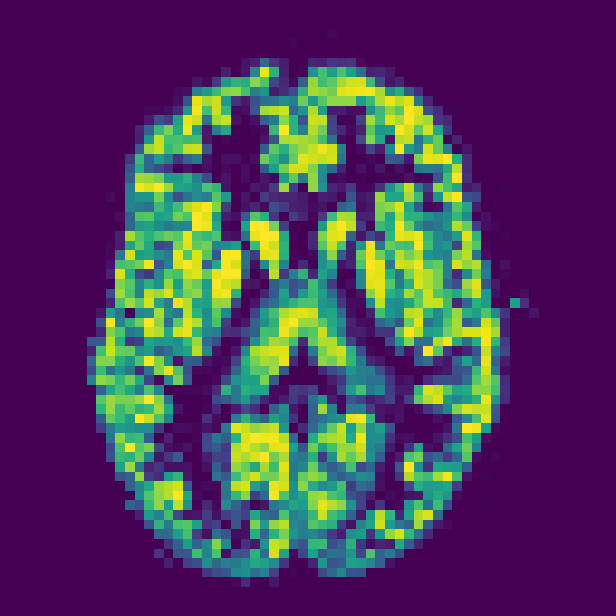}
    \includegraphics[width=\figsize\textwidth]{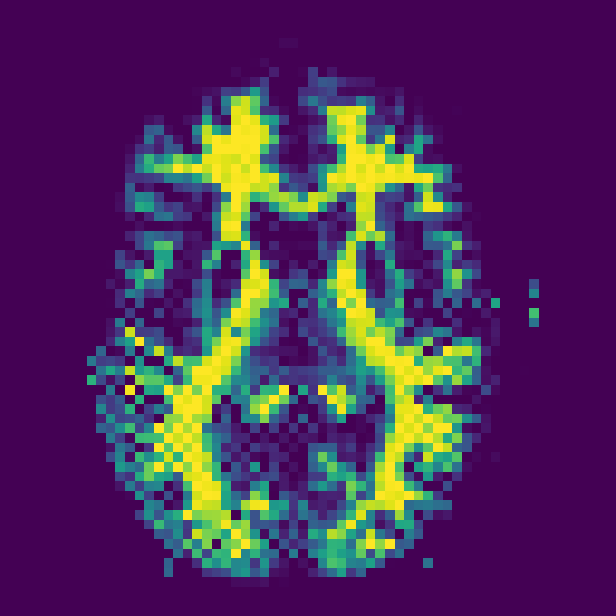}
    \caption{From the left are CSF, GM, and WM. The first and third rows show the raw tissue probability maps for subjects 20 and 38, respectively, of the BrainWeb dataset. The second and fourth rows show the optimized probability maps using the baseline configuration for subjects 20 and 38, respectively.}
    \label{sup_fig:subject_20_38_best_results}
\end{figure}

\begin{figure}
    \textbf{\hspace{\csfoffset}CSF\hspace{\gmoffset}GM\hspace{\wmoffset}WM}\par\medskip
    \centering
    \includegraphics[width=\figsize\textwidth]{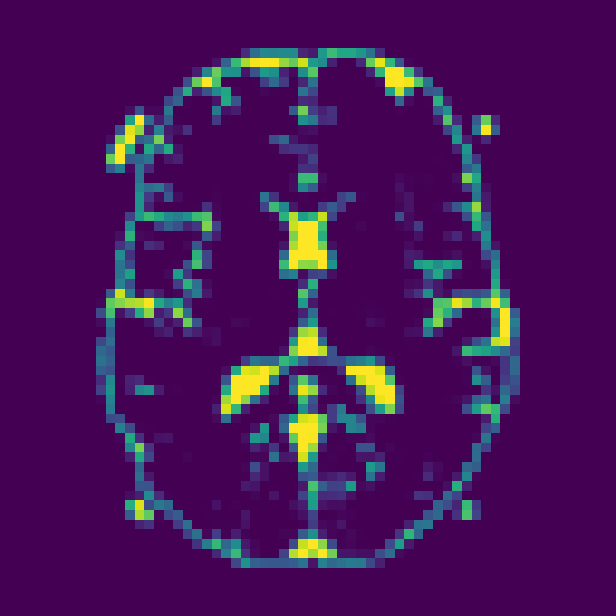}
    \includegraphics[width=\figsize\textwidth]{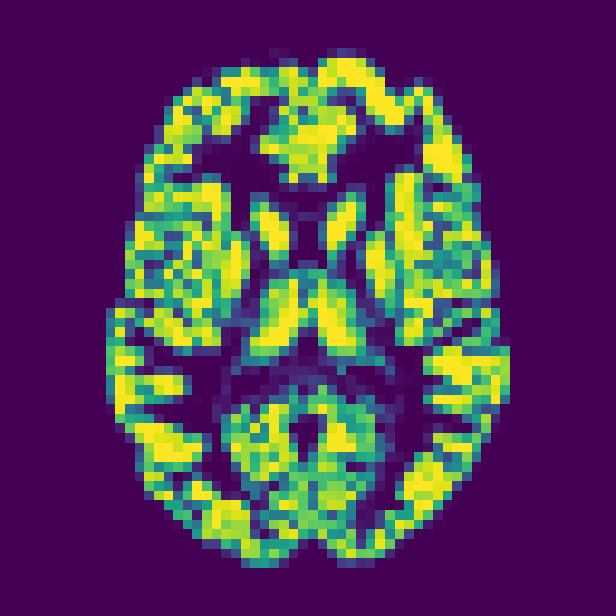}
    \includegraphics[width=\figsize\textwidth]{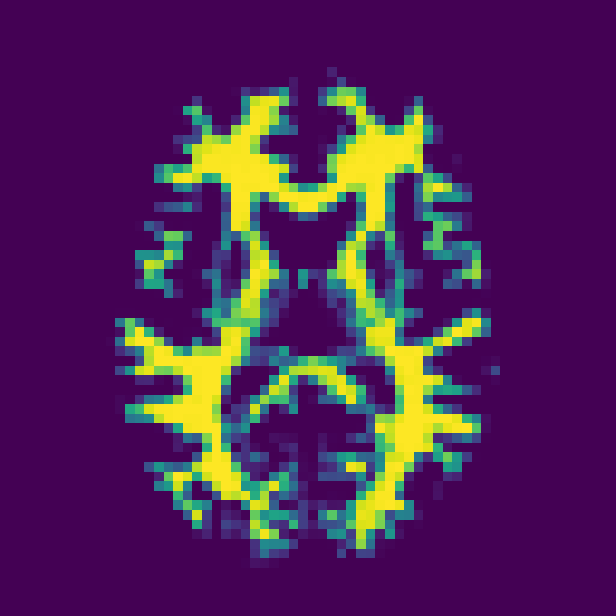}
    \\[\smallskipamount]
    \includegraphics[width=\figsize\textwidth]{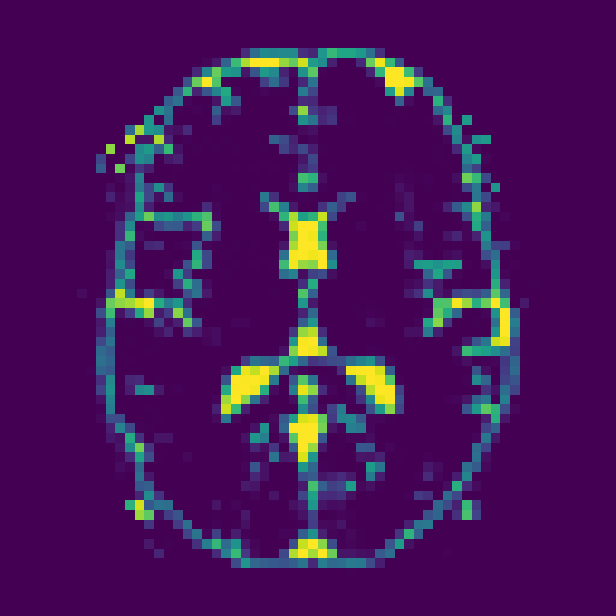}
    \includegraphics[width=\figsize\textwidth]{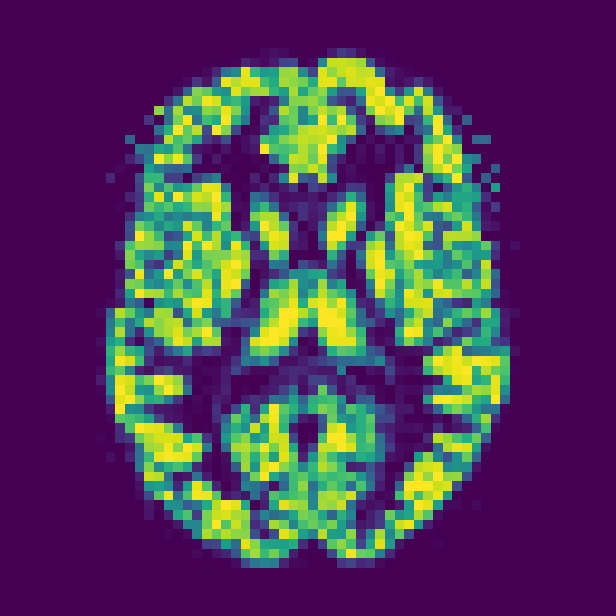}
    \includegraphics[width=\figsize\textwidth]{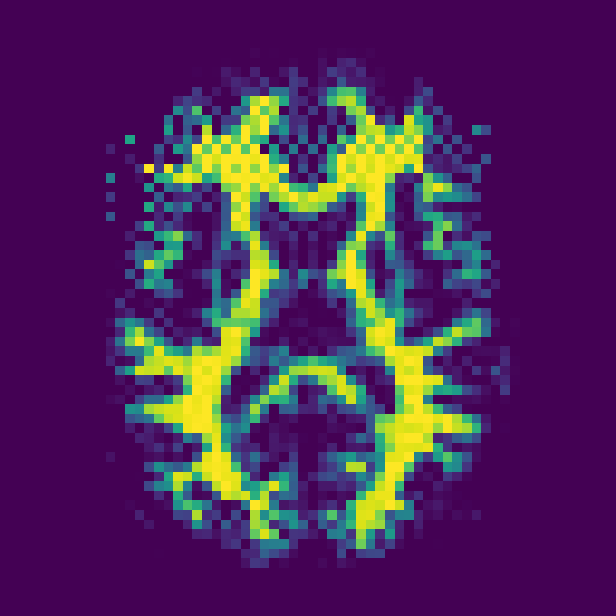}
    \\[\smallskipamount]
    \includegraphics[width=\figsize\textwidth]{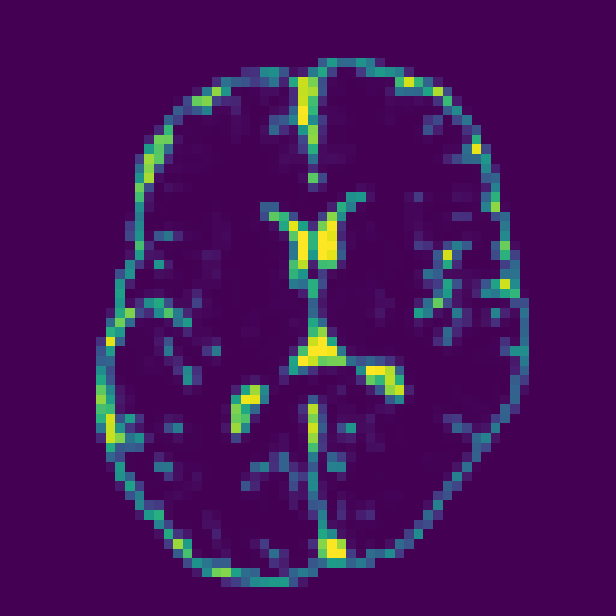}
    \includegraphics[width=\figsize\textwidth]{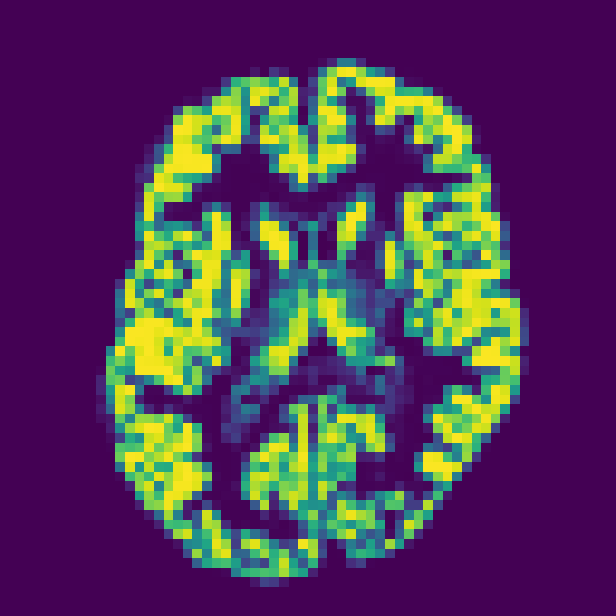}
    \includegraphics[width=\figsize\textwidth]{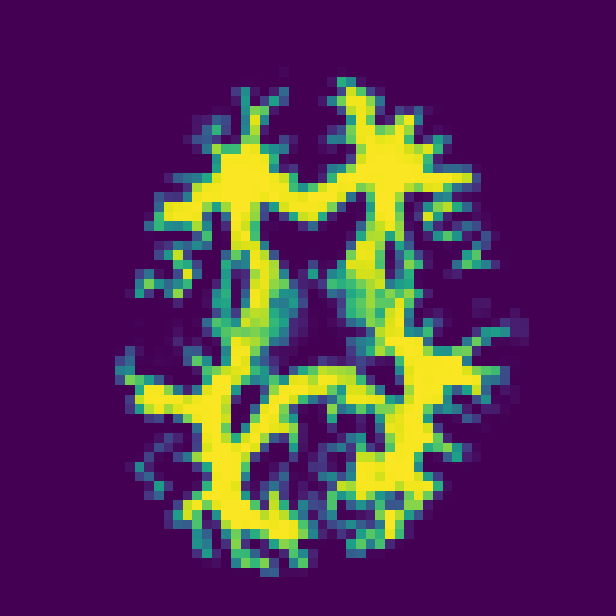}
    \\[\smallskipamount]
    \includegraphics[width=\figsize\textwidth]{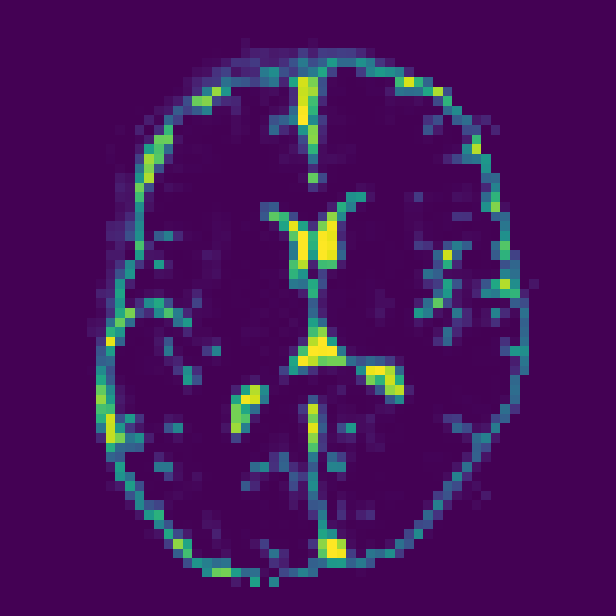}
    \includegraphics[width=\figsize\textwidth]{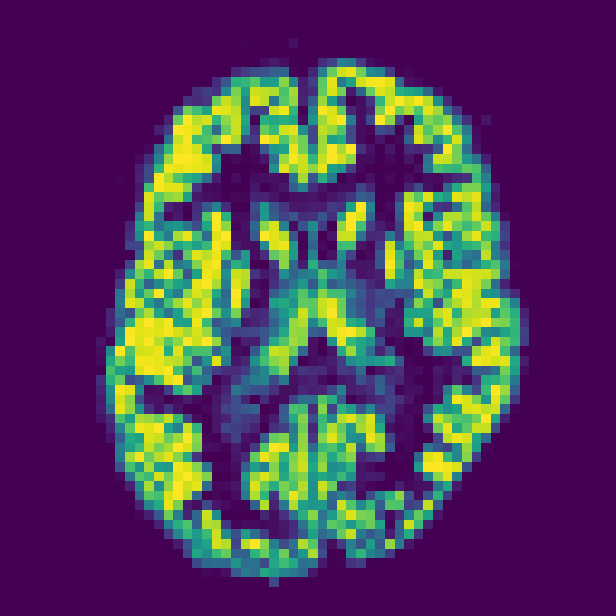}
    \includegraphics[width=\figsize\textwidth]{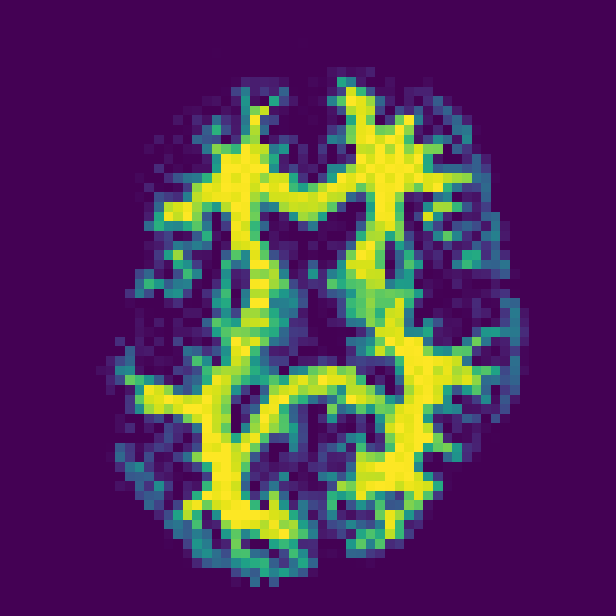}
    \caption{From the left are CSF, GM, and WM. The first and third rows show the raw tissue probability maps for subjects 41 and 43, respectively, of the BrainWeb dataset. The second and fourth rows show the optimized probability maps using the baseline configuration for subjects 41 and 43, respectively.}
    \label{sup_fig:subject_41_43_best_results}
\end{figure}

\begin{figure}
    \textbf{\hspace{\csfoffset}CSF\hspace{\gmoffset}GM\hspace{\wmoffset}WM}\par\medskip
    \centering
    \includegraphics[width=\figsize\textwidth]{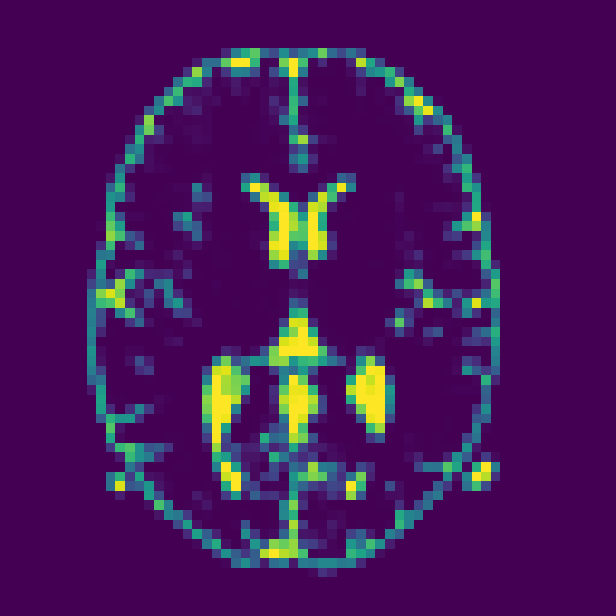}
    \includegraphics[width=\figsize\textwidth]{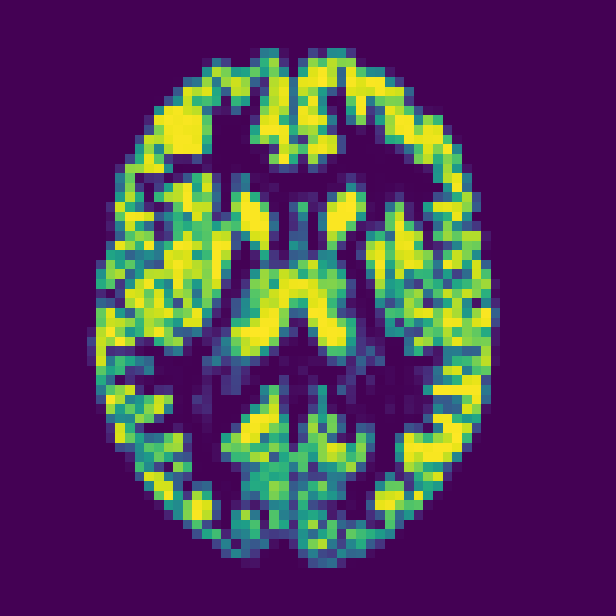}
    \includegraphics[width=\figsize\textwidth]{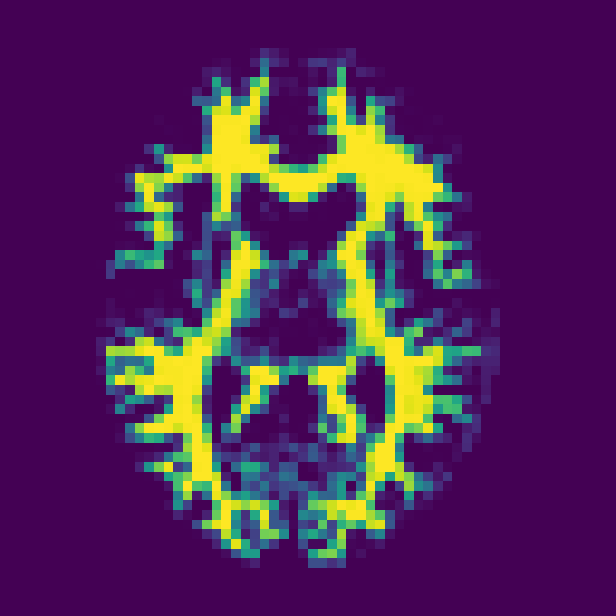}
    \\[\smallskipamount]
    \includegraphics[width=\figsize\textwidth]{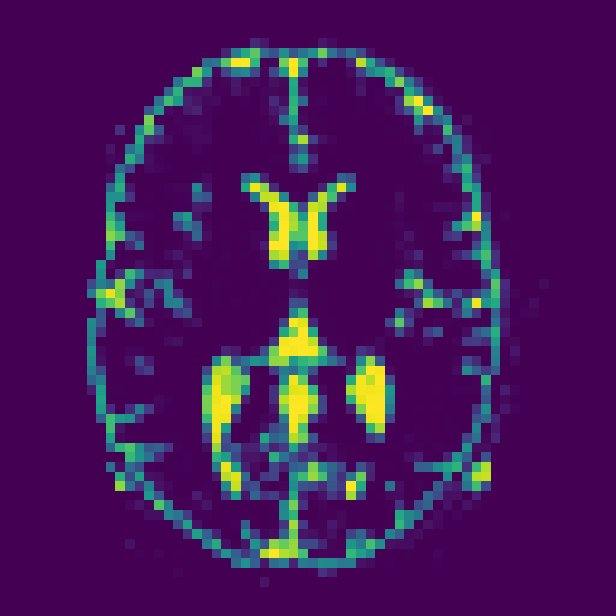}
    \includegraphics[width=\figsize\textwidth]{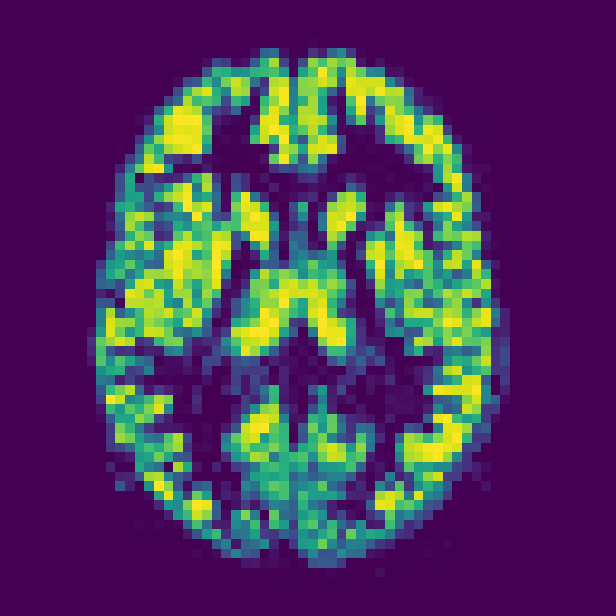}
    \includegraphics[width=\figsize\textwidth]{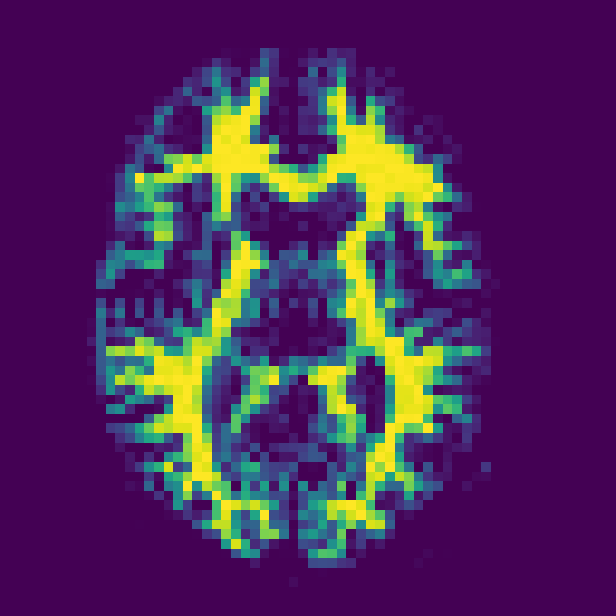}
    \\[\smallskipamount]
    \includegraphics[width=\figsize\textwidth]{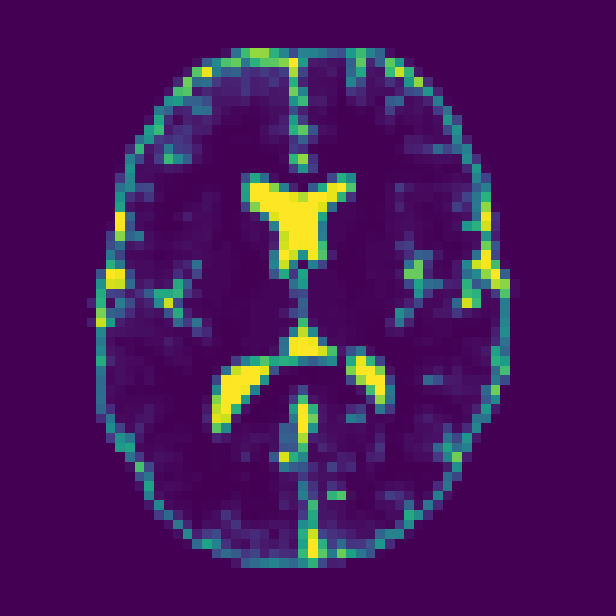}
    \includegraphics[width=\figsize\textwidth]{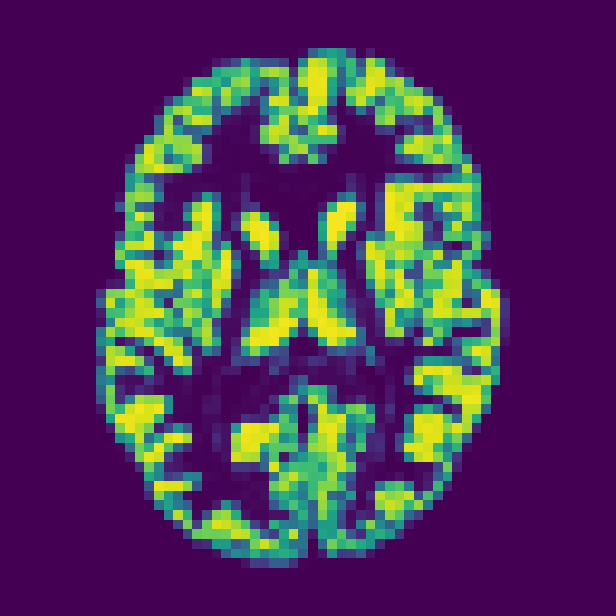}
    \includegraphics[width=\figsize\textwidth]{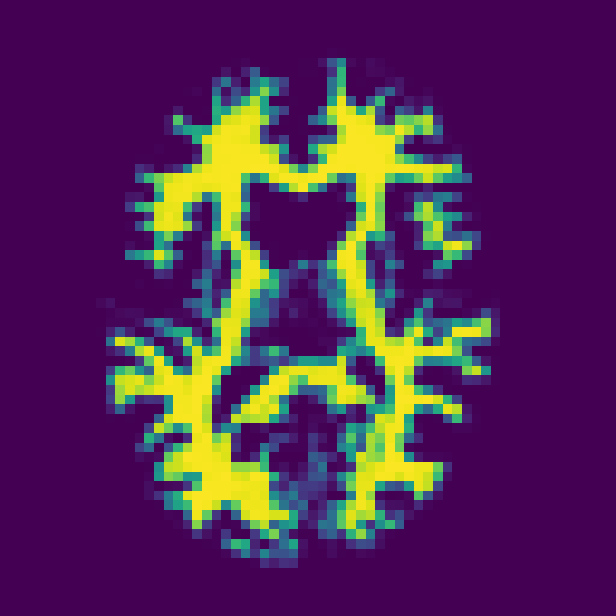}
    \\[\smallskipamount]
    \includegraphics[width=\figsize\textwidth]{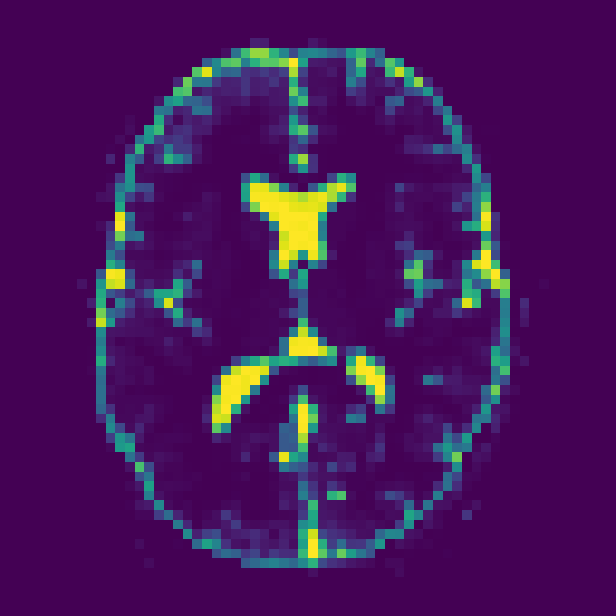}
    \includegraphics[width=\figsize\textwidth]{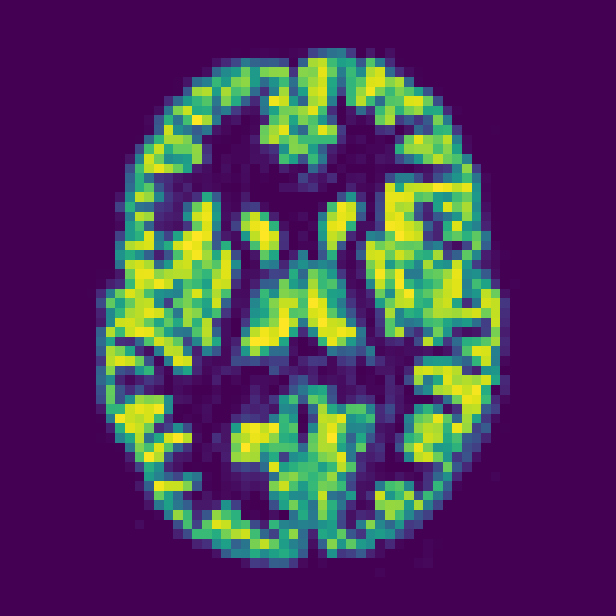}
    \includegraphics[width=\figsize\textwidth]{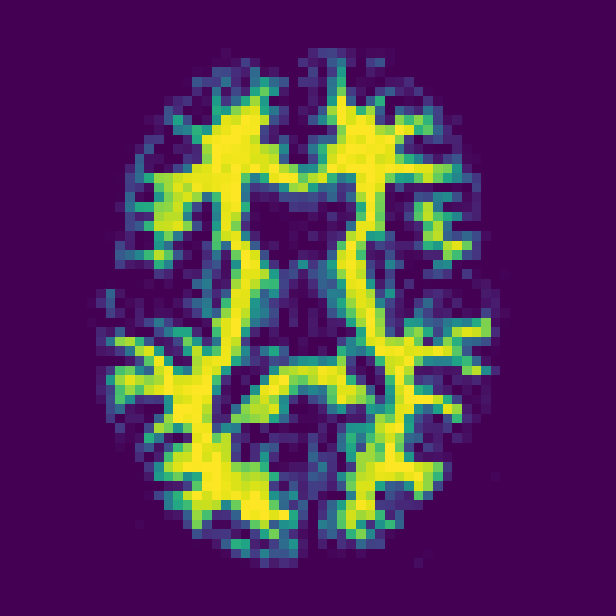}
    \caption{From the left are CSF, GM, and WM. The first and third rows show the raw tissue probability maps for subjects 44 and 45, respectively, of the BrainWeb dataset. The second and fourth rows show the optimized probability maps using the baseline configuration for subjects 44 and 45, respectively.}
    \label{sup_fig:subject_44_45_best_results}
\end{figure}

\begin{figure}
    \textbf{\hspace{\csfoffset}CSF\hspace{\gmoffset}GM\hspace{\wmoffset}WM}\par\medskip
    \centering
    \includegraphics[width=\figsize\textwidth]{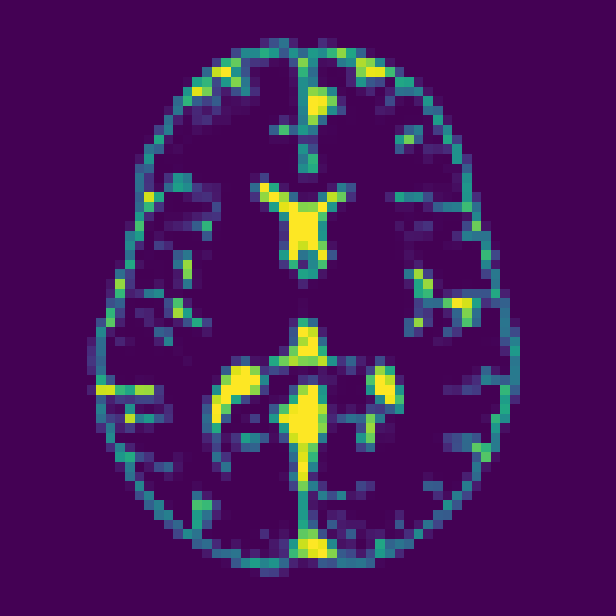}
    \includegraphics[width=\figsize\textwidth]{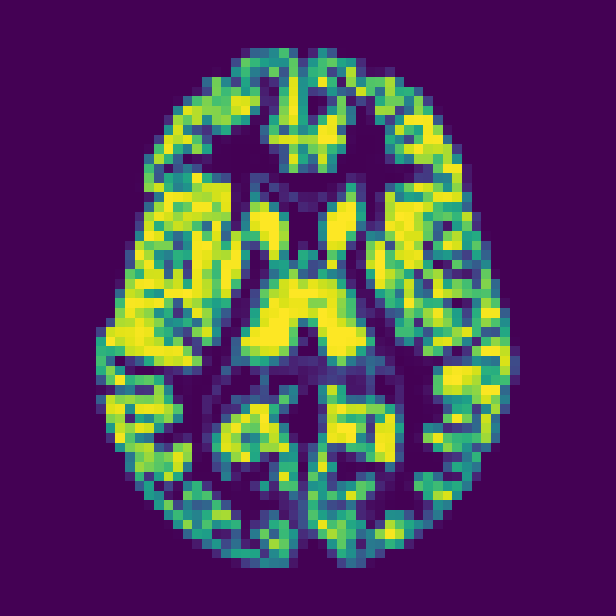}
    \includegraphics[width=\figsize\textwidth]{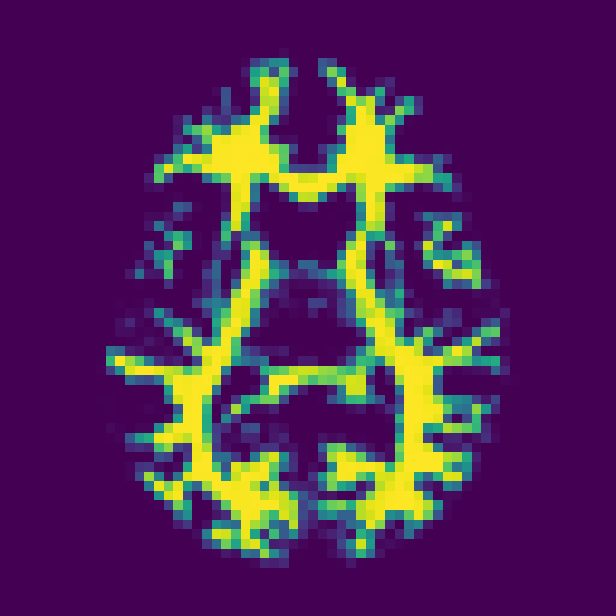}
    \\[\smallskipamount]
    \includegraphics[width=\figsize\textwidth]{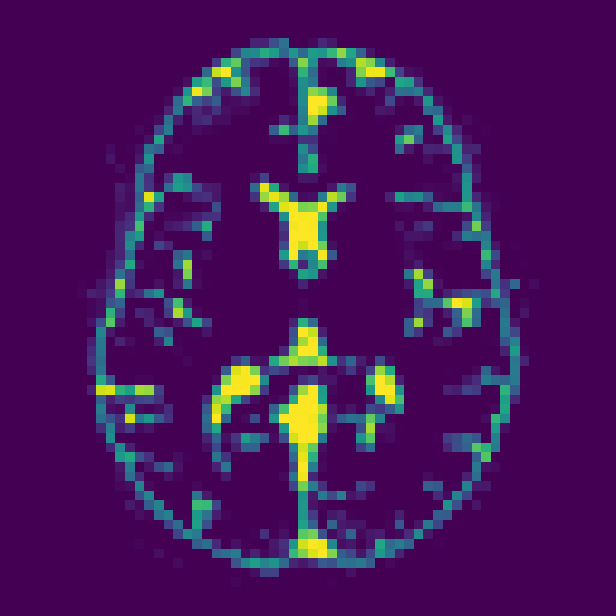}
    \includegraphics[width=\figsize\textwidth]{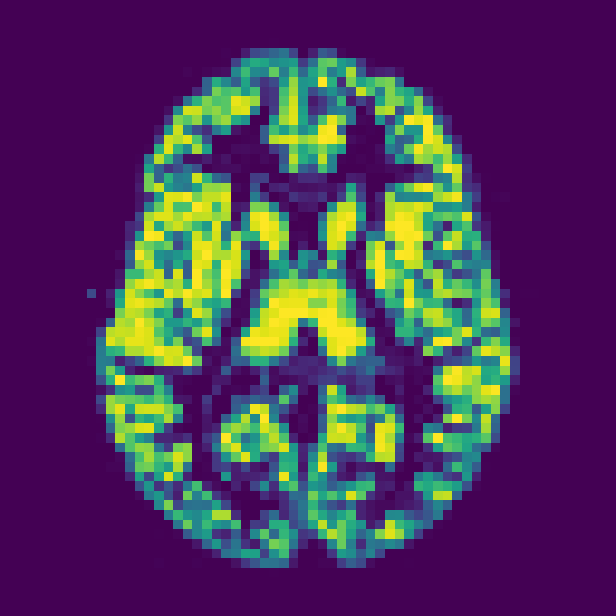}
    \includegraphics[width=\figsize\textwidth]{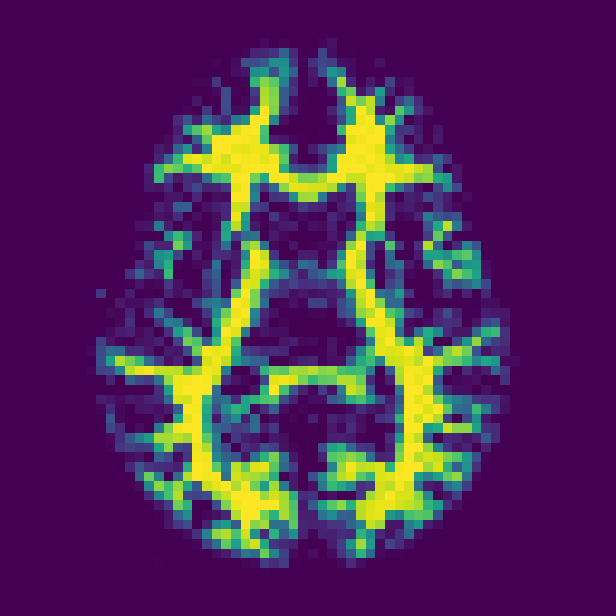}
    \\[\smallskipamount]
    \includegraphics[width=\figsize\textwidth]{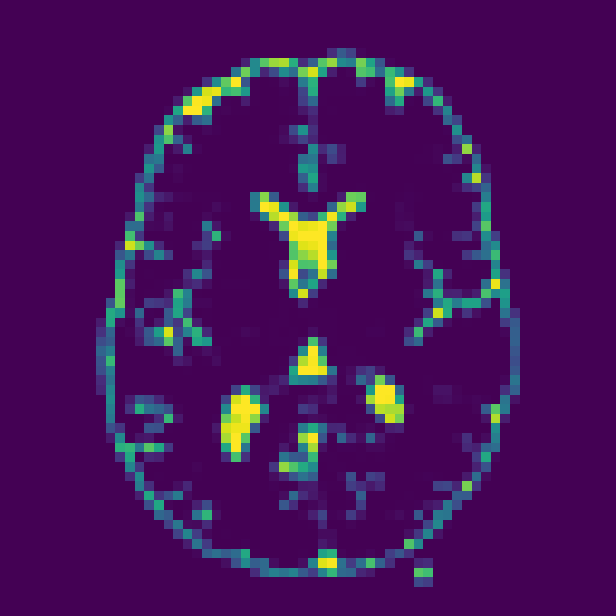}
    \includegraphics[width=\figsize\textwidth]{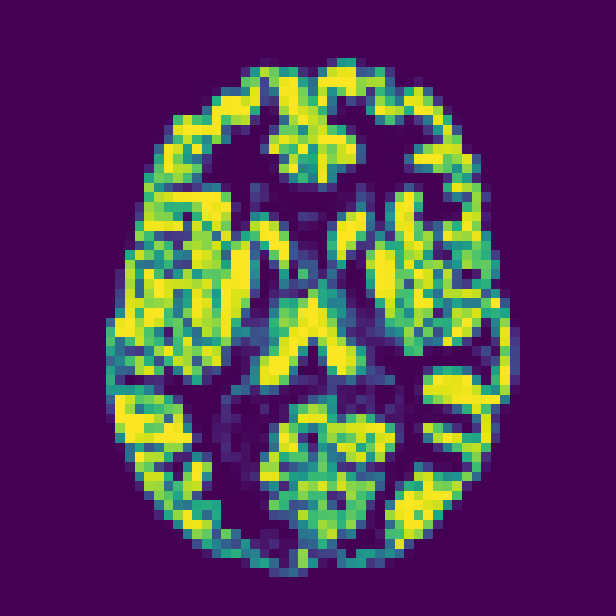}
    \includegraphics[width=\figsize\textwidth]{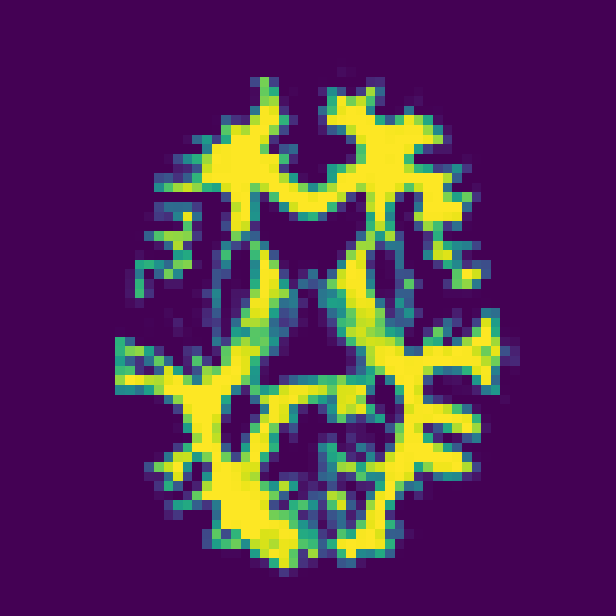}
    \\[\smallskipamount]
    \includegraphics[width=\figsize\textwidth]{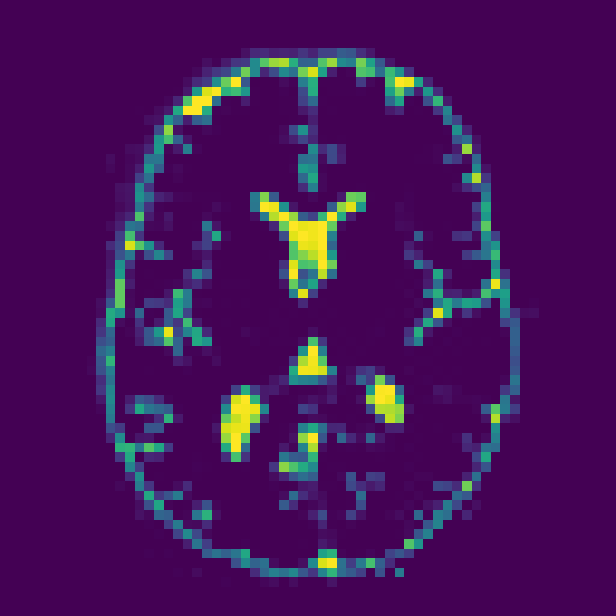}
    \includegraphics[width=\figsize\textwidth]{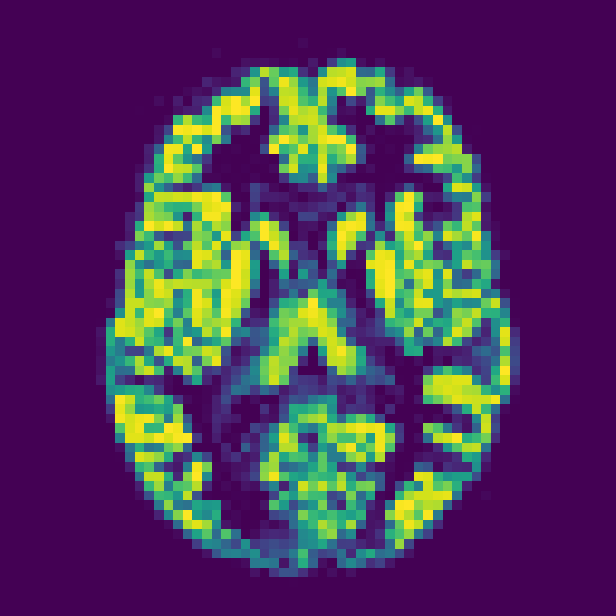}
    \includegraphics[width=\figsize\textwidth]{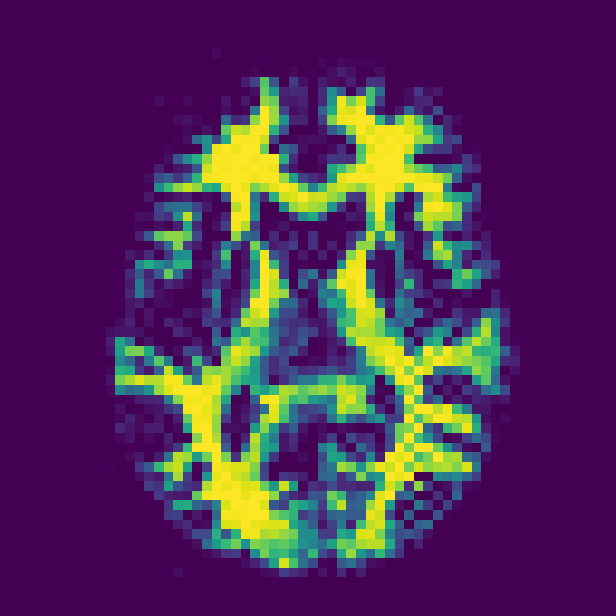}
    \caption{From the left are CSF, GM, and WM. The first and third rows show the raw tissue probability maps for subjects 46 and 47, respectively, of the BrainWeb dataset. The second and fourth rows show the optimized probability maps using the baseline configuration for subjects 46 and 47, respectively.}
    \label{sup_fig:subject_46_47_best_results}
\end{figure}

\begin{figure}
    \textbf{\hspace{\csfoffset}CSF\hspace{\gmoffset}GM\hspace{\wmoffset}WM}\par\medskip
    \centering
    \includegraphics[width=\figsize\textwidth]{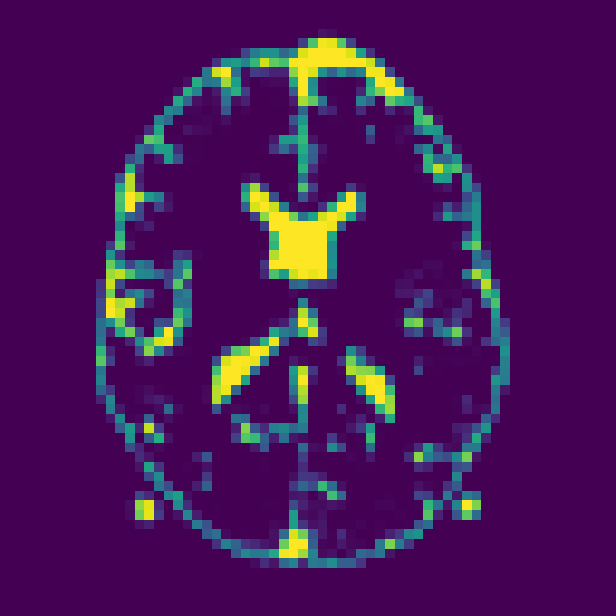}
    \includegraphics[width=\figsize\textwidth]{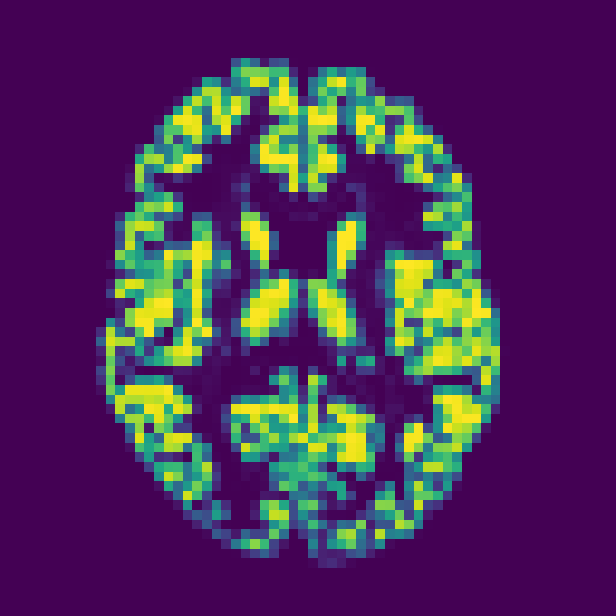}
    \includegraphics[width=\figsize\textwidth]{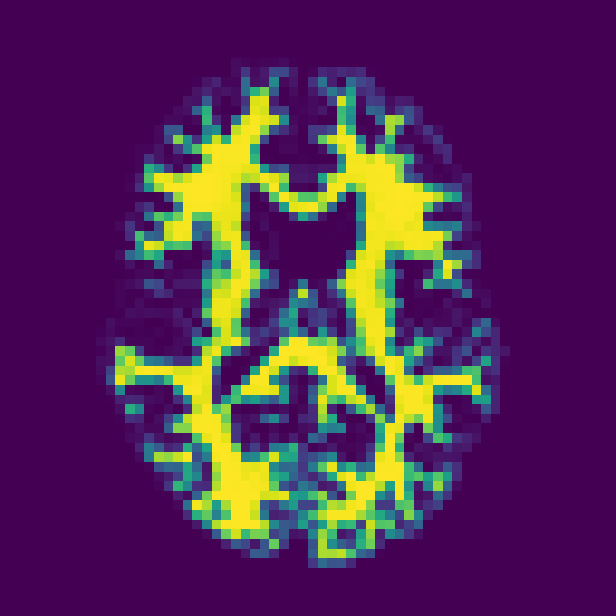}
    \\[\smallskipamount]
    \includegraphics[width=\figsize\textwidth]{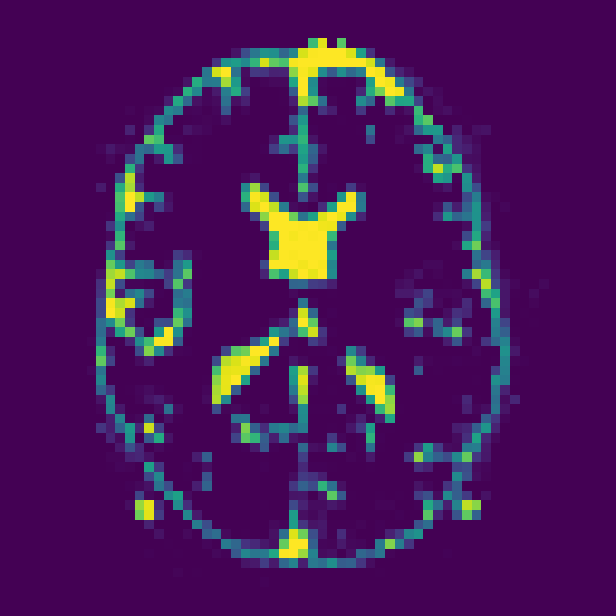}
    \includegraphics[width=\figsize\textwidth]{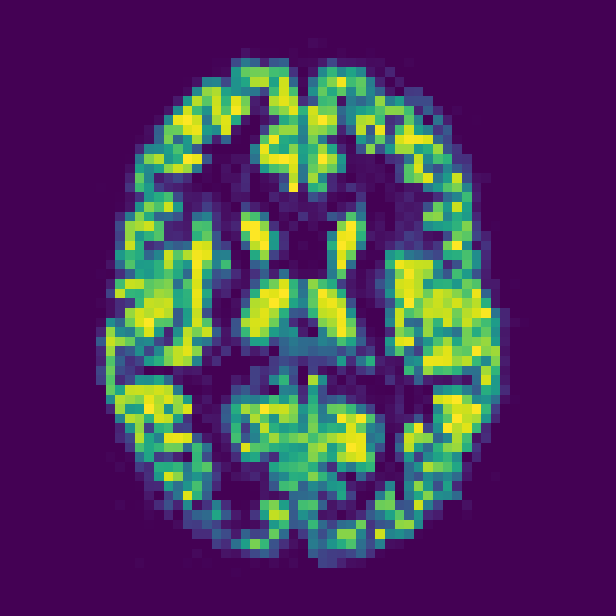}
    \includegraphics[width=\figsize\textwidth]{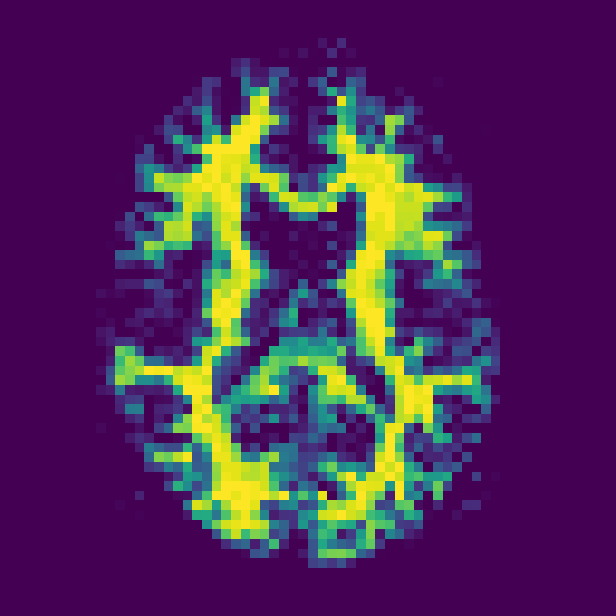}
    \\[\smallskipamount]
    \includegraphics[width=\figsize\textwidth]{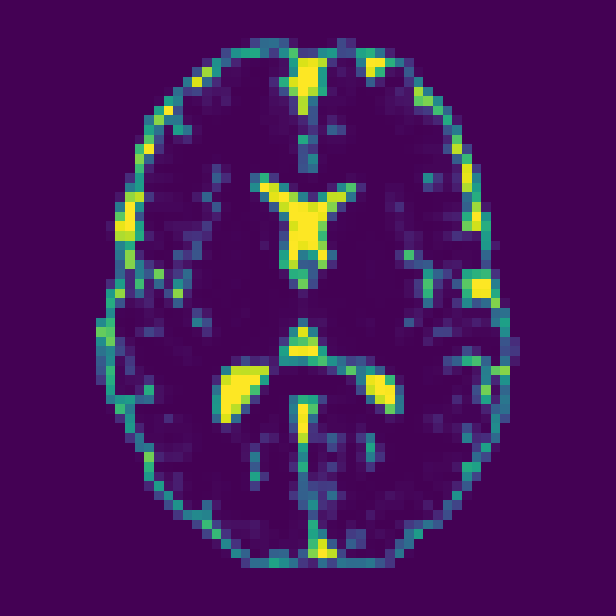}
    \includegraphics[width=\figsize\textwidth]{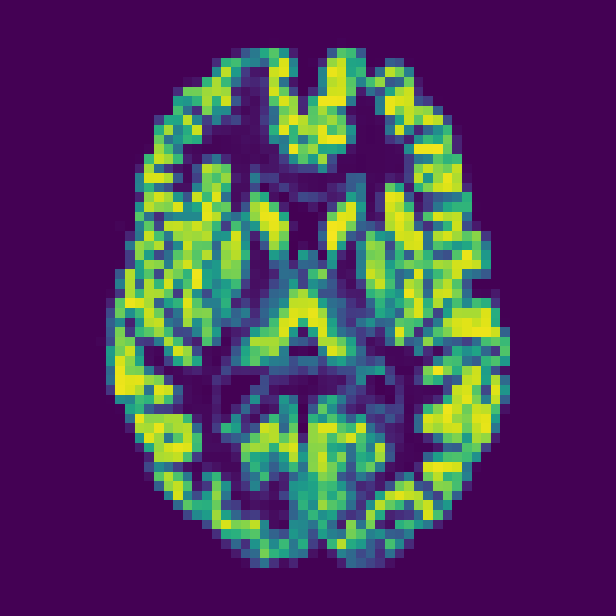}
    \includegraphics[width=\figsize\textwidth]{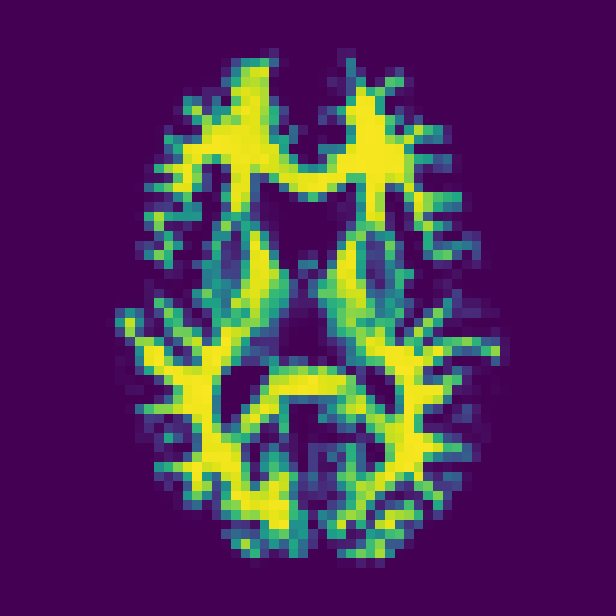}
    \\[\smallskipamount]
    \includegraphics[width=\figsize\textwidth]{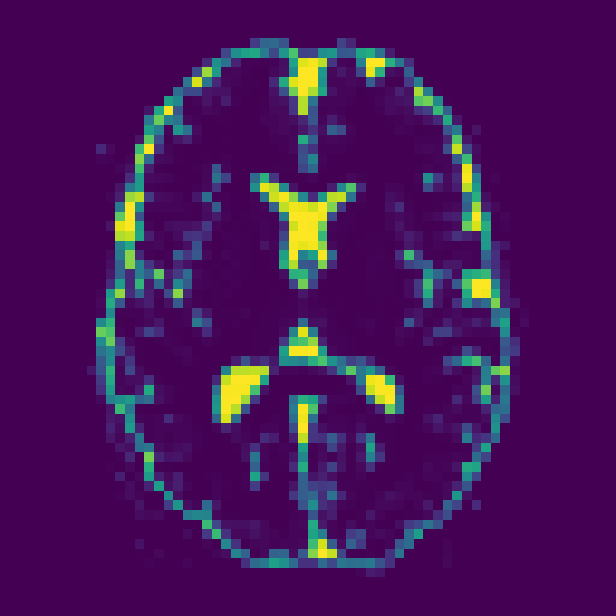}
    \includegraphics[width=\figsize\textwidth]{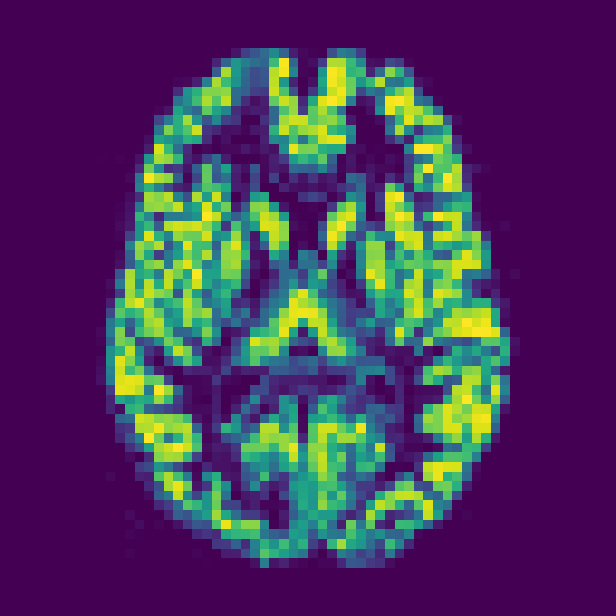}
    \includegraphics[width=\figsize\textwidth]{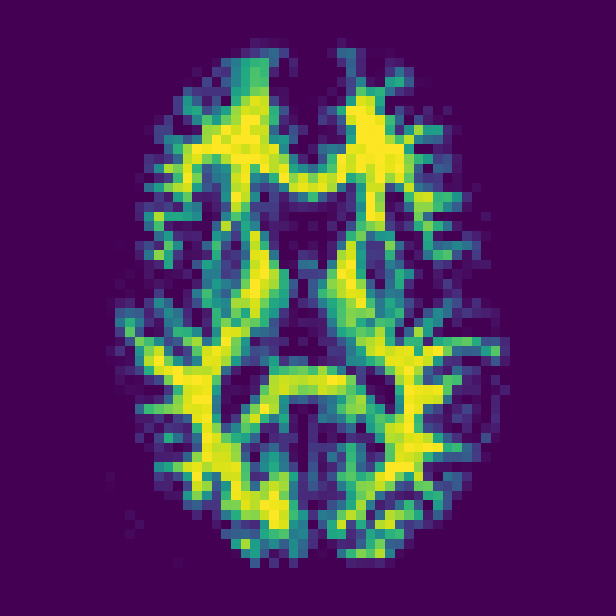}
    \caption{From the left are CSF, GM, and WM. The first and third rows show the raw tissue probability maps for subjects 48 and 49, respectively, of the BrainWeb dataset. The second and fourth rows show the optimized probability maps using the baseline configuration for subjects 48 and 49, respectively.}
    \label{sup_fig:subject_48_49_best_results}
\end{figure}

\begin{figure}
    \textbf{\hspace{\csfoffset}CSF\hspace{\gmoffset}GM\hspace{\wmoffset}WM}\par\medskip
    \centering
    \includegraphics[width=\figsize\textwidth]{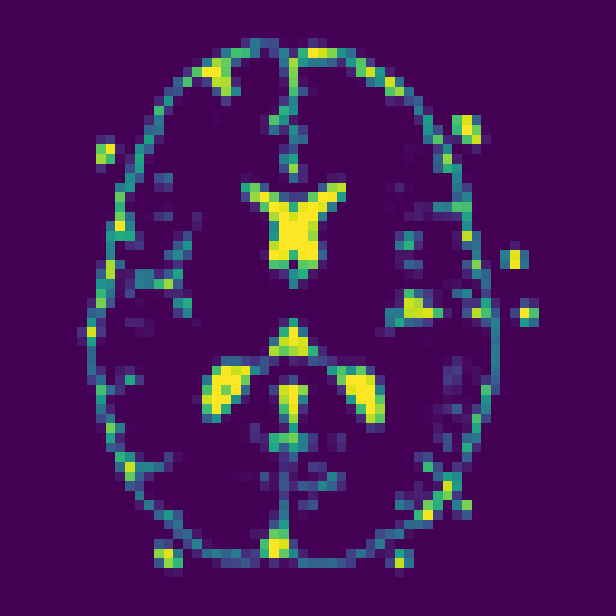}
    \includegraphics[width=\figsize\textwidth]{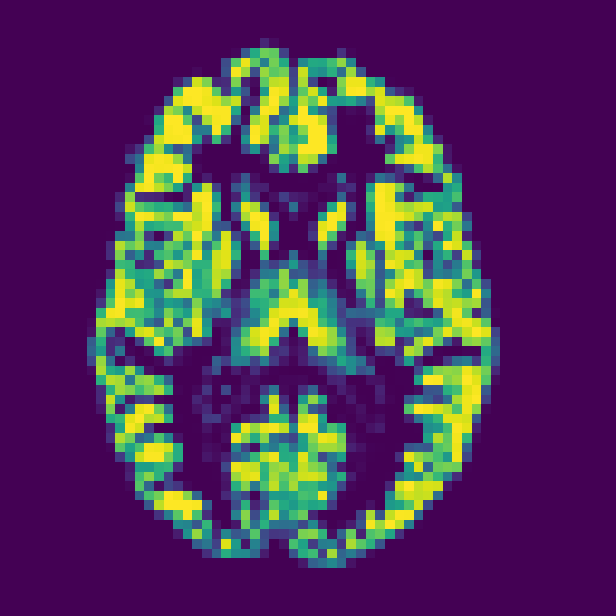}
    \includegraphics[width=\figsize\textwidth]{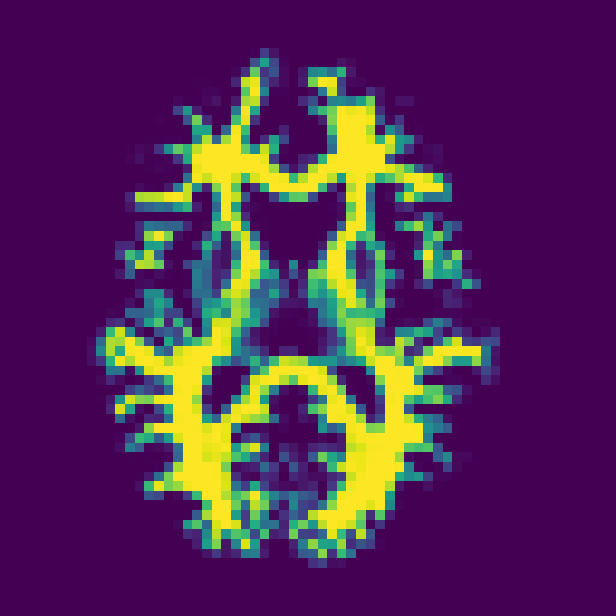}
    \\[\smallskipamount]
    \includegraphics[width=\figsize\textwidth]{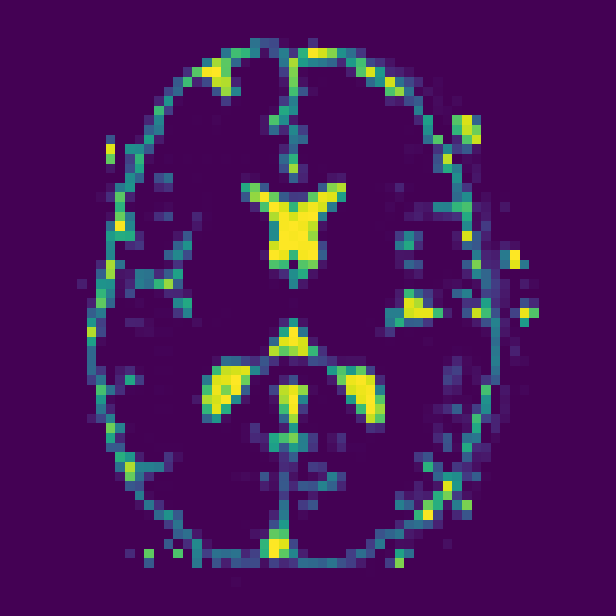}
    \includegraphics[width=\figsize\textwidth]{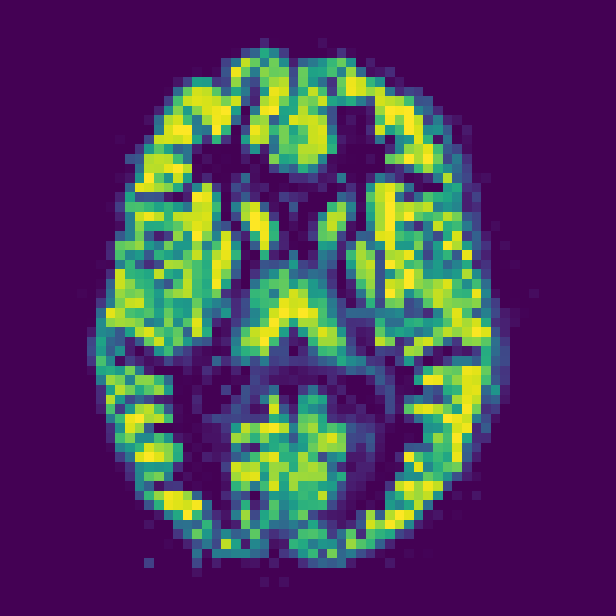}
    \includegraphics[width=\figsize\textwidth]{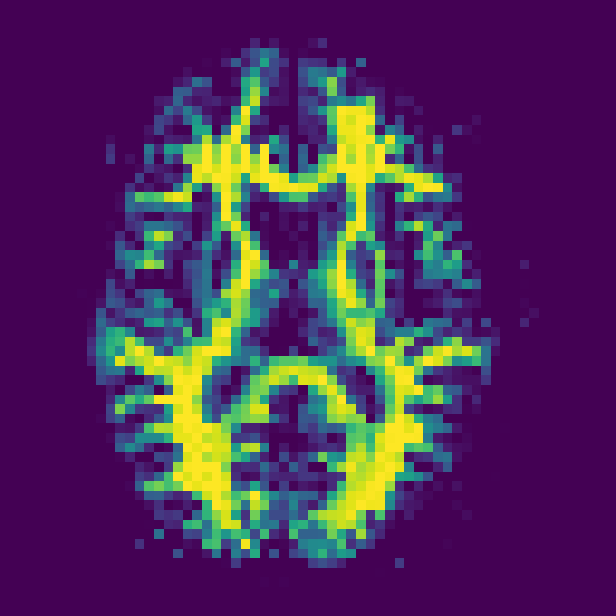}
    \\[\smallskipamount]
    \includegraphics[width=\figsize\textwidth]{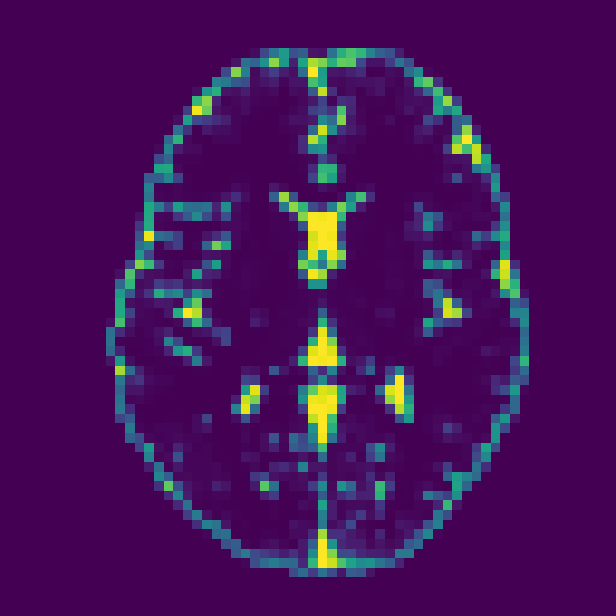}
    \includegraphics[width=\figsize\textwidth]{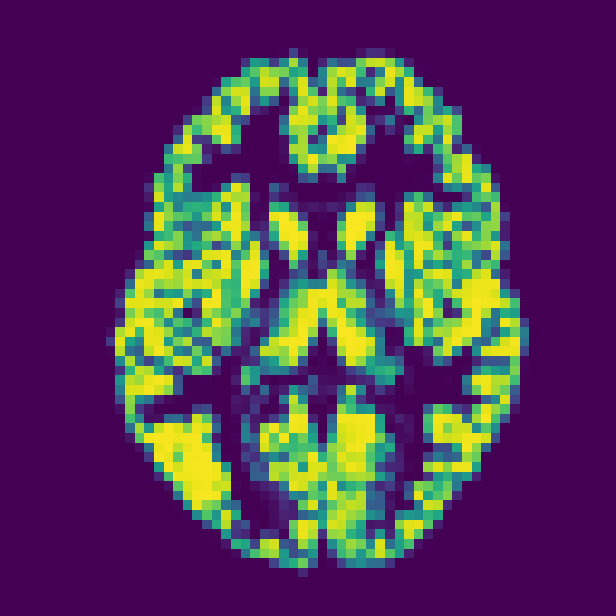}
    \includegraphics[width=\figsize\textwidth]{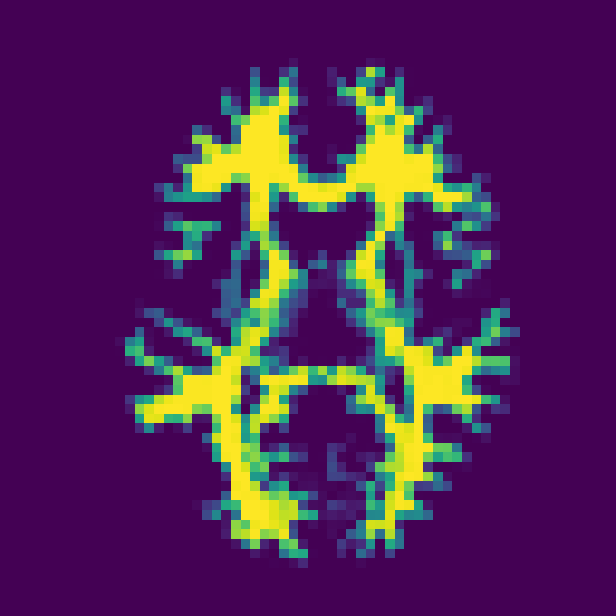}
    \\[\smallskipamount]
    \includegraphics[width=\figsize\textwidth]{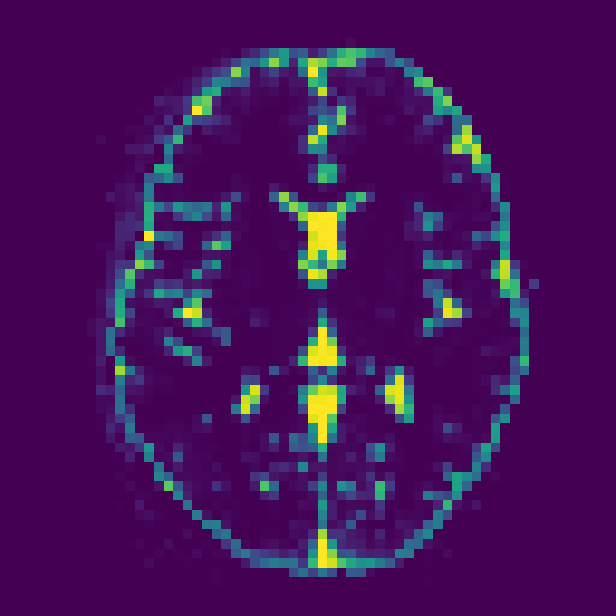}
    \includegraphics[width=\figsize\textwidth]{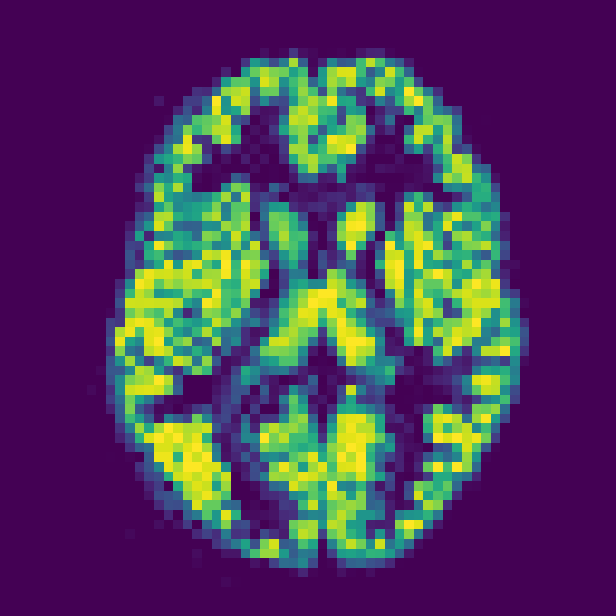}
    \includegraphics[width=\figsize\textwidth]{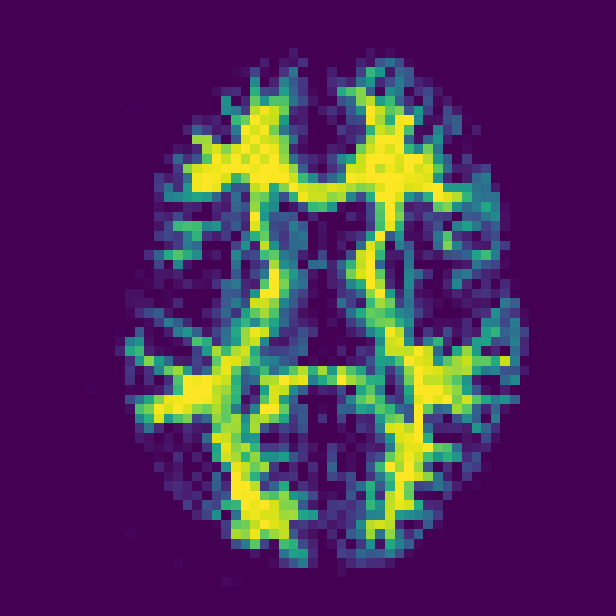}
    \caption{From the left are CSF, GM, and WM. The first and third rows show the raw tissue probability maps for subjects 50 and 51, respectively, of the BrainWeb dataset. The second and fourth rows show the optimized probability maps using the baseline configuration for subjects 50 and 51, respectively.}
    \label{sup_fig:subject_50_51_best_results}
\end{figure}

\begin{figure}
    \textbf{\hspace{\csfoffset}CSF\hspace{\gmoffset}GM\hspace{\wmoffset}WM}\par\medskip
    \centering
    \includegraphics[width=\figsize\textwidth]{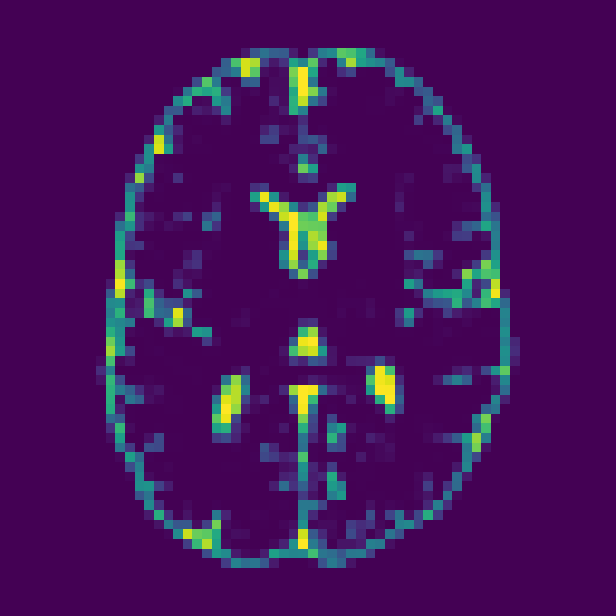}
    \includegraphics[width=\figsize\textwidth]{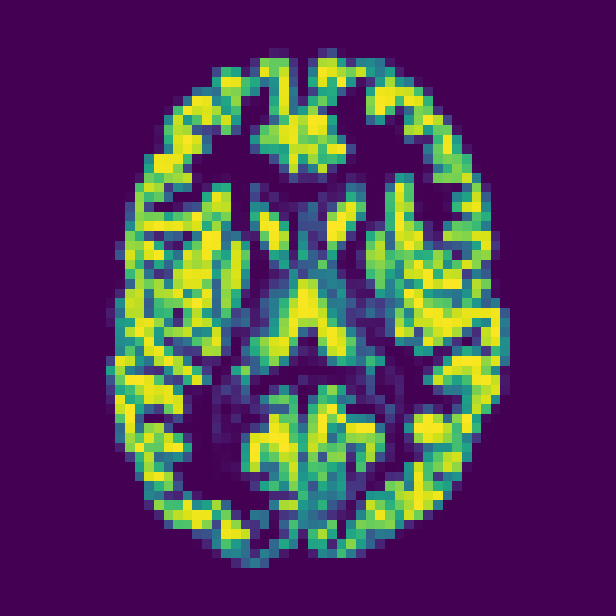}
    \includegraphics[width=\figsize\textwidth]{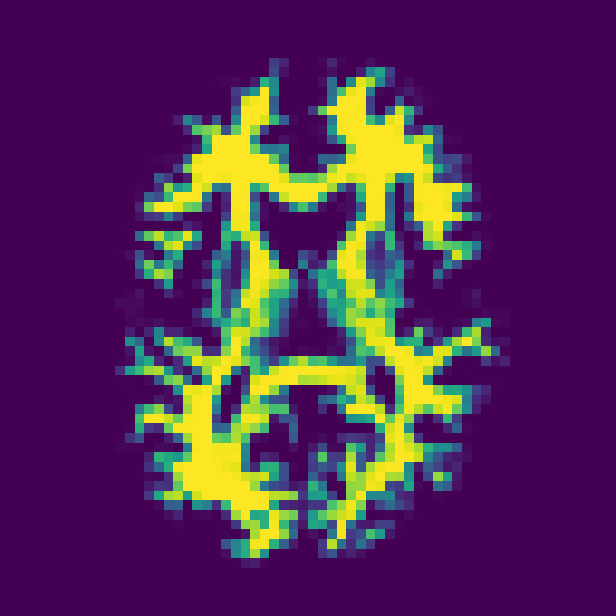}
    \\[\smallskipamount]
    \includegraphics[width=\figsize\textwidth]{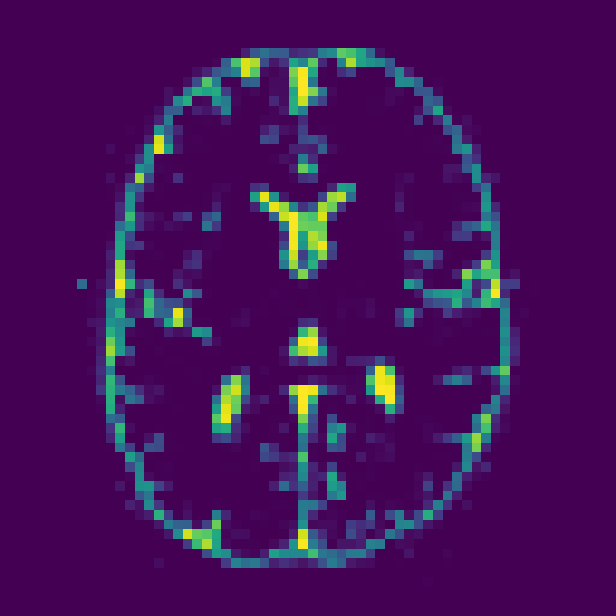}
    \includegraphics[width=\figsize\textwidth]{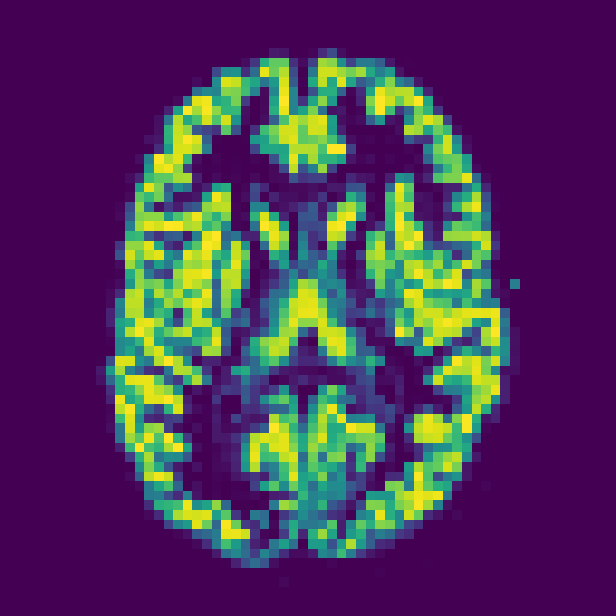}
    \includegraphics[width=\figsize\textwidth]{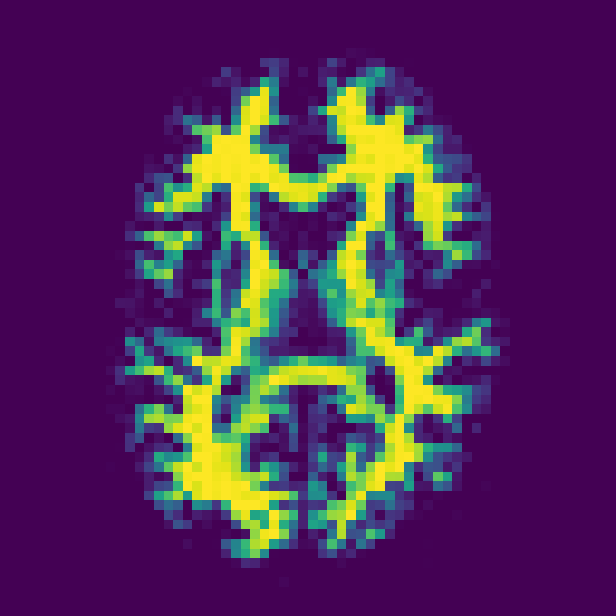}
    \\[\smallskipamount]
    \includegraphics[width=\figsize\textwidth]{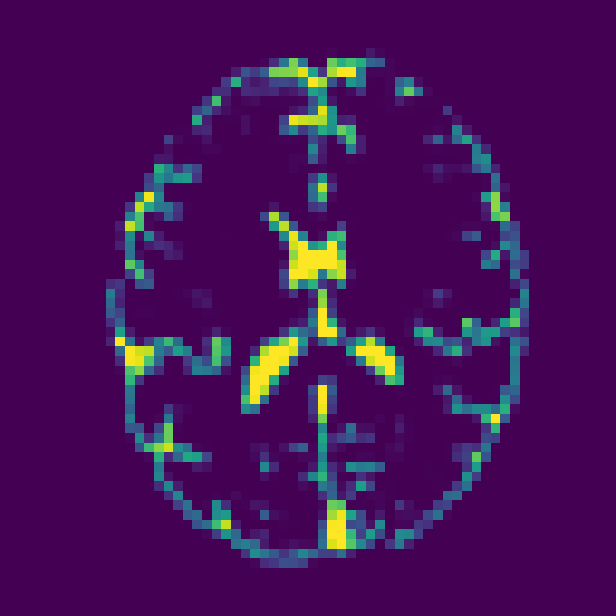}
    \includegraphics[width=\figsize\textwidth]{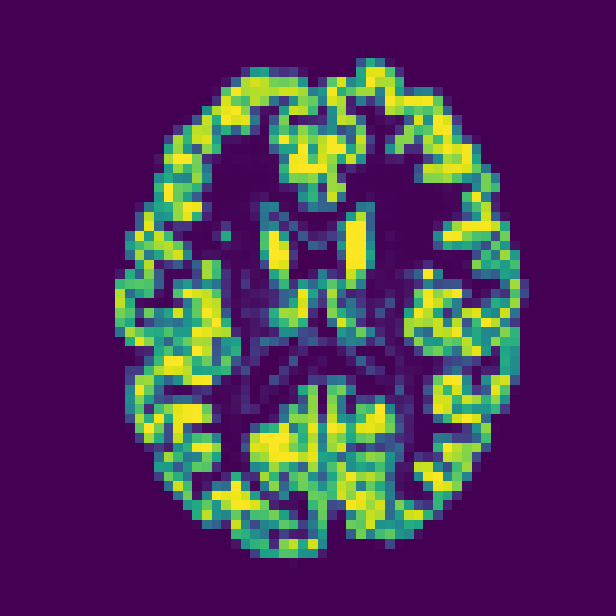}
    \includegraphics[width=\figsize\textwidth]{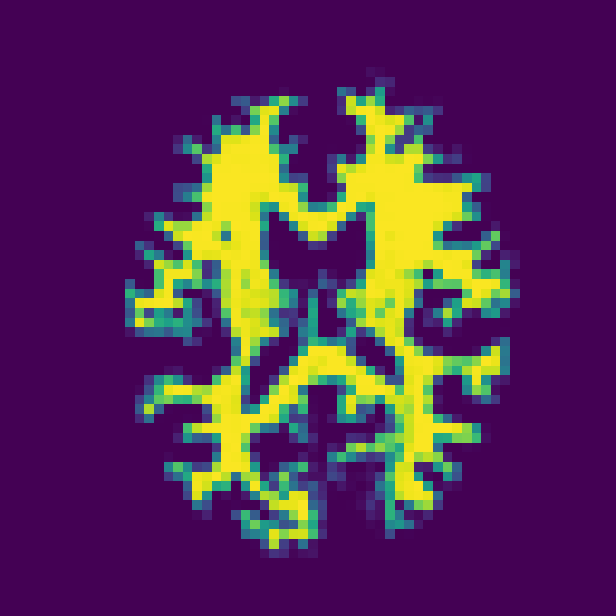}
    \\[\smallskipamount]
    \includegraphics[width=\figsize\textwidth]{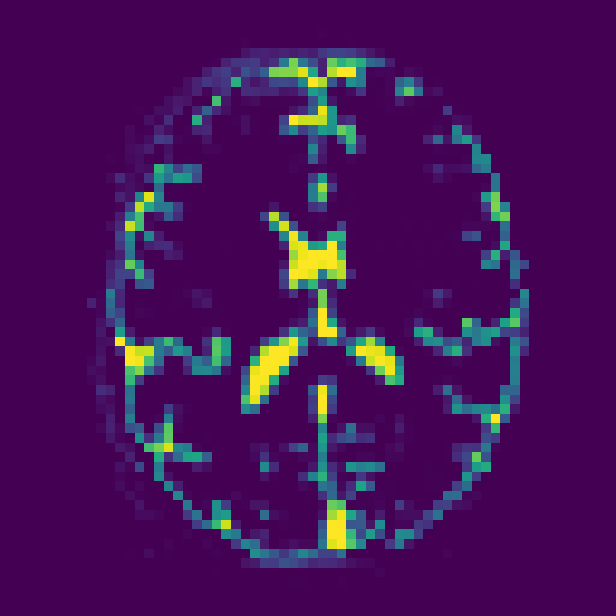}
    \includegraphics[width=\figsize\textwidth]{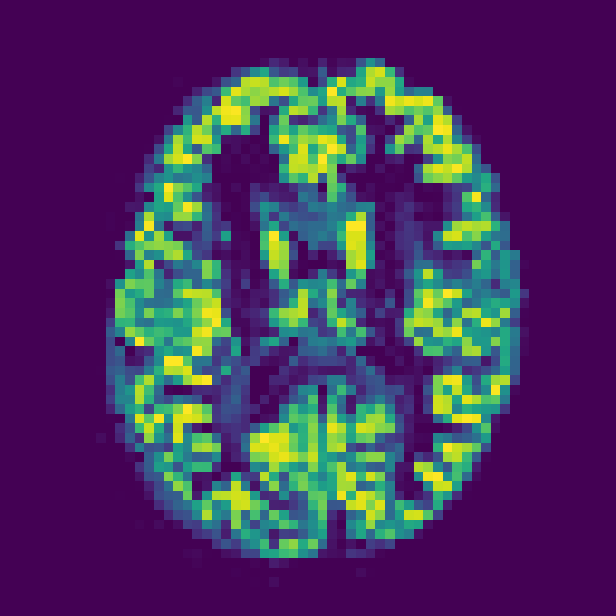}
    \includegraphics[width=\figsize\textwidth]{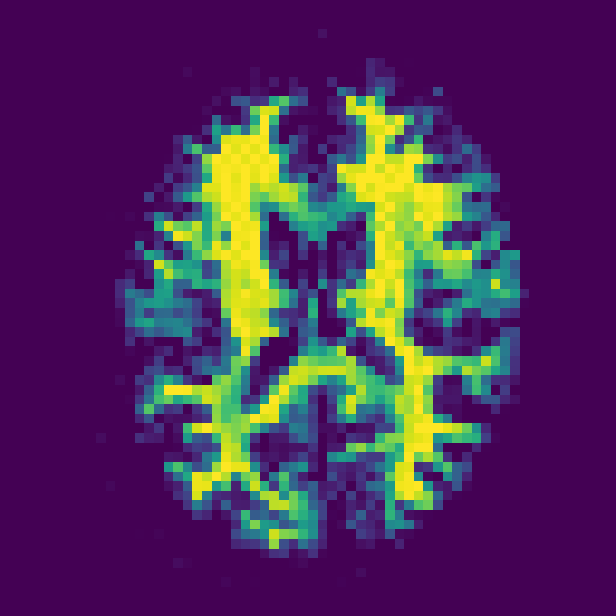}
    \caption{From the left are CSF, GM, and WM. The first and third rows show the raw tissue probability maps for subjects 52 and 53, respectively, of the BrainWeb dataset. The second and fourth rows show the optimized probability maps using the baseline configuration for subjects 52 and 53, respectively.}
    \label{sup_fig:subject_52_53_best_results}
\end{figure}

\begin{figure}
    \textbf{\hspace{\csfoffset}CSF\hspace{\gmoffset}GM\hspace{\wmoffset}WM}\par\medskip
    \centering
    \includegraphics[width=\figsize\textwidth]{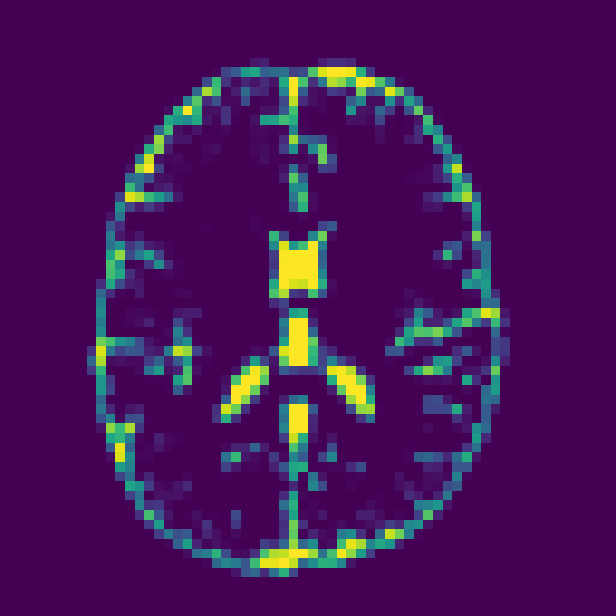}
    \includegraphics[width=\figsize\textwidth]{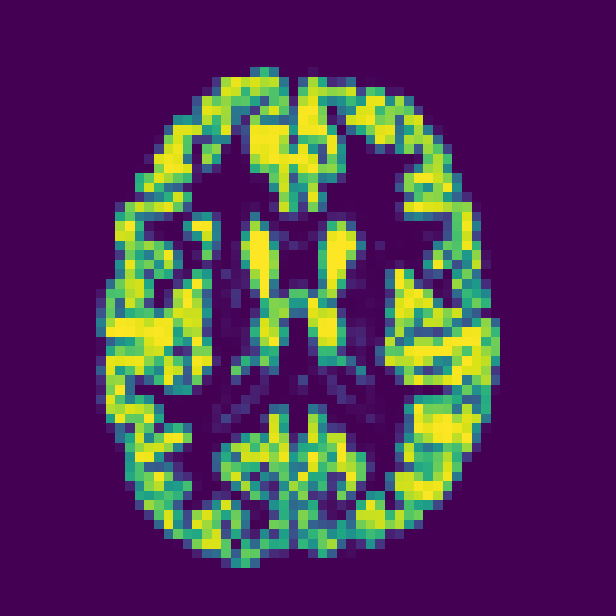}
    \includegraphics[width=\figsize\textwidth]{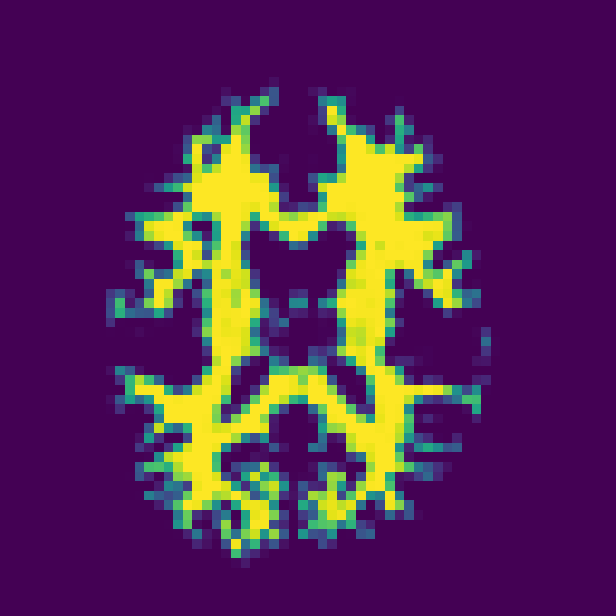}
    \\[\smallskipamount]
    \includegraphics[width=\figsize\textwidth]{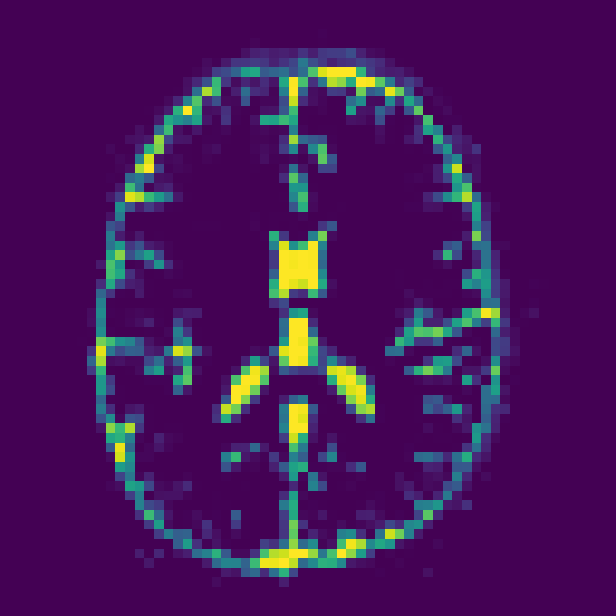}
    \includegraphics[width=\figsize\textwidth]{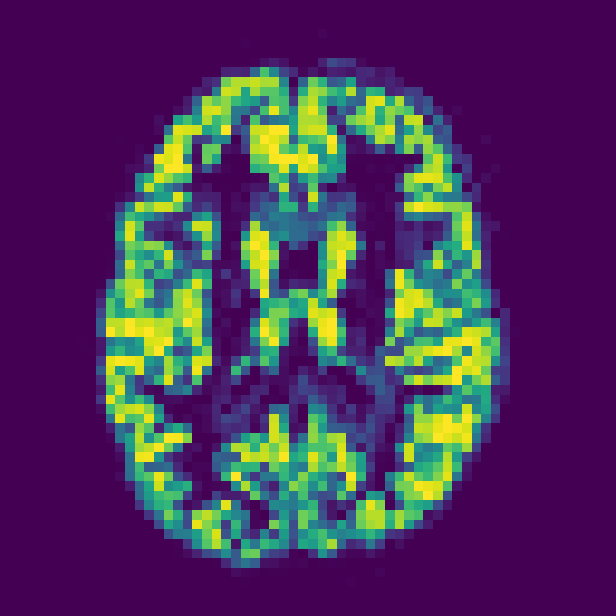}
    \includegraphics[width=\figsize\textwidth]{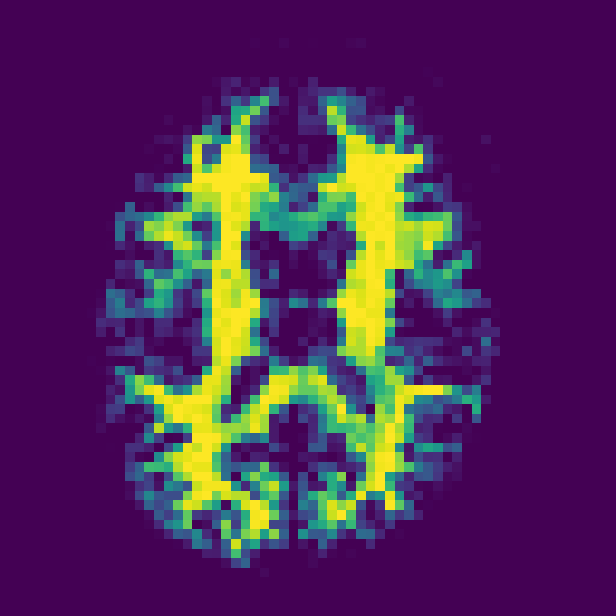}
    \caption{From the left are CSF, GM, and WM. The first row shows the raw tissue probability maps for subject 54 of the BrainWeb dataset. The second row shows the optimized probability maps using the baseline configuration for subject 54.}
    \label{sup_fig:subject_54_best_results}
\end{figure}

\end{document}